\documentclass[12pt,a4paper]{article}
\usepackage{amsfonts}
\usepackage{amssymb}
\usepackage{graphicx}
\usepackage{setspace}
\usepackage{comment}
\usepackage{natbib}
\usepackage{amsmath}
\usepackage{amsthm}
\usepackage{palatino}
\usepackage{paralist}
\usepackage{subcaption}
\usepackage[table]{xcolor}
\usepackage{multirow}
\usepackage{booktabs}
\usepackage{dsfont}
\usepackage{indentfirst}
\usepackage{enumerate}
\usepackage{longtable}
\usepackage{caption}
\usepackage{algpseudocode} 
\usepackage{tikz}
\usetikzlibrary{arrows}
\usetikzlibrary{positioning}
\usetikzlibrary{calc}
\newdimen\nodeDist
\usetikzlibrary{decorations.shapes}
\tikzset{decorate sep/.style 2 args=
	{decorate,decoration={shape backgrounds,shape=circle,shape size=#1,shape sep=#2}}}

\tikzset{
	treenode/.style = {shape=rectangle, 
		draw, align=center,
		top color=white, bottom color=white},
	env/.style      = {treenode}
}
\tikzstyle{level 1}=[level distance=1.5cm, sibling distance=4cm, font=\normalsize]
\tikzstyle{level 2}=[level distance=1.5cm, sibling distance=2cm, font=\normalsize]
\tikzstyle{level 3}=[level distance=1.5cm, sibling distance=1cm, font=\normalsize]
\tikzstyle{level 4}=[level distance=1.5cm, sibling distance=.5cm, font=\normalsize]

\usepackage[hyperindex,breaklinks]{hyperref}
\hypersetup{colorlinks=true,       
	linkcolor=red,       
	citecolor=blue,        
	filecolor=magenta,      
	urlcolor=cyan           
}  

\usepackage{algorithm,algpseudocode}

\makeatletter
\renewcommand{\fnum@algorithm}{\fname@algorithm}
\makeatother

\newcommand{\F}{\mathbf{F}}

\newcommand{\rr}{\mathbf{r}}

\newcommand{\R}{\mathcal{R}}
\newcommand{\RR}{\mathbf{R}}
\newcommand{\T}{\mathcal{T}}

\def\fs{\footnotesize}

\def\argmin{\operatornamewithlimits{arg\,min}}
\def\argmax{\operatornamewithlimits{arg\,max}}

\newcommand{\w}{\mathbf{w}}

\newcommand{\z}{\mathbf{z}}
\newcommand{\cc}{\mathbf{c}}

\usepackage[toc,page]{appendix}
\usepackage{setspace}
\usepackage[left = 1in, right = 1in, top = 1in, bottom = 1in]{geometry}
\usepackage{titlesec}
\titleformat*{\section}{\large\bfseries}
\titleformat*{\subsection}{\normalsize\bfseries}
\titleformat*{\subsubsection}{\normalsize\bfseries}
\titleformat*{\paragraph}{\normalsize\bfseries}

\titlespacing{\section}{0pt}{*1.6}{*1.6}
\titlespacing{\subsection}{0pt}{*1.4}{*1.4}
\titlespacing{\subsubsection}{0pt}{*1.2}{*1.2}
\titlespacing{\paragraph}{0pt}{*1.2}{*1.2}

\usepackage{lscape}
\usepackage{rotating}

\title{\Large \bf  Growing the Efficient Frontier on Panel Trees\vspace{-0.6cm}
\thanks{\scriptsize{We are grateful to Fahiz Baba-Yara (discussant), Svetlana Bryzgalova (discussant), John Campbell, Vic Chien, Andrei Goncalves, Bing Han, Kewei Hou, Bob Jarrow, Ron Kaniel, Chris Malloy, Alberto Rossi (discussant), Gideon Saar, Shrihari Santosh (discussant), Artem Streltsov, Ziying Sun (discussant), Peixuan Yuan (discussant), and Guofu Zhou for detailed comments and suggestions. 
We also thank Doron Avramov, Jie Cao, Zhanhui Chen, Eric Ghysels, P. Richard Hahn, Dashan Huang, Sophia Zhengzi Li, Asaf Manela, Stefan Nagel, Adam Reed, Artem Streltsov, Yinan Su, Fred Sun, Yaki Tsaig, Mao Ye, Dacheng Xiu, Bobby Yu, and seminar and conference participants at 
ArrowStreet Capital, 2024 AsianFA Annual Meeting, Baylor University, BUAA, Ca' Foscari University of Venice, Cambridge University Algorithmic Trading Society Quant Conference, CFTRC 2022, CityUHK, CKGSB, Conference on FinTech, Innovation and Development at Hangzhou (2nd), Cornell University, 2024 EFMA Annual Meeting, Goethe University Frankfurt, GSU-RFS FinTech Conference 2022, HKAIFT-Columbia joint seminar, Hunan University, INFORMS Annual Meeting 2021, 4th International FinTech Research Forum (RUC), 2023 Mid-South DATA Conference (Memphis), KAIST Digital Finance Conference, NFA 2022, University of Oxford SML-Fin Seminar, OSU, PHBS, PKU Guanghua, PKU-NSD, PKU-NUS Annual International Conference on Quantitative Finance and Economics, Qube Research and Technology, Reichman University (IDC Herzliya), Schroders Investments, Shanghai Financial Forefront Symposium (3rd), SHUFE, NUS, SMU, SUSTech, TAMU, Tsinghua SEM, UNC, University of Bath, University of Hawai'i, USC, USTC, University of Macau, World Online Seminars on Machine Learning in Finance, WFA 2022, and 2023 XJTLU AI and Big Data in Accounting and Finance Conference for constructive discussions and feedback. 
Feng's research is partly supported by HK RGC grants (GRF-11502721, GRF-11502023) and an NSFC grant (NSFC-72203190). 
He J.'s research is partly supported by HK RGC grants 
(ECS-21504921, GRF-11507022, GRF-11509224) and an NSFC grant (NSFC-72403214).
Feng and He J. are partly supported by the InnoHK initiative and the Laboratory for AI-Powered Financial Technologies.
The authors also acknowledge research awards from INQUIRE Europe and IQAM Research.
Cong (E-mail: \texttt{will.cong@cornell.edu}) is at Cornell University SC Johnson College of Business, Fudan University FISF, and NBER; 
Feng (E-mail: \texttt{gavin.feng@cityu.edu.hk}) and He (E-mail: \texttt{jingyuhe@cityu.edu.hk}) are at the City University of Hong Kong; He (E-mail: \texttt{xin.he@ustc.edu.cn}) is at Faculty of Business for Science and Technology, School of Management, University of Science and Technology of China.}}
}

\author{
         \normalsize{Lin William Cong} \\
	\and \normalsize{Guanhao Feng} \\
	\and \normalsize{Jingyu He} \\
	\and \normalsize{Xin He} 
}

\date{\vspace{-1cm} \small{First draft: Oct. 2021; this draft: Oct. 2024}}
\begin{document}

\maketitle
\vspace{-1cm}

\begin{abstract}

\setstretch{1.1}{\small 
\noindent 
We introduce a new class of tree-based models, P-Trees, for analyzing (unbalanced) panel of individual asset returns, generalizing high-dimensional sorting with economic guidance and interpretability. Under the mean-variance efficient framework, P-Trees construct test assets that significantly advance the efficient frontier compared to commonly used test assets, with alphas unexplained by benchmark pricing models. P-Tree tangency portfolios also constitute traded factors, recovering the pricing kernel and outperforming popular observable and latent factor models for investments and cross-sectional pricing. Finally, P-Trees capture the complexity of asset returns with sparsity, achieving out-of-sample Sharpe ratios close to those attained only by over-parameterized large models.

}

\medskip


\medskip


\end{abstract}

\newpage
\setstretch{1.6}
\section{Introduction}

Estimating the mean-variance efficient (MVE) frontier is crucial for asset pricing and investment management. Yet, estimating the tangency portfolio \citep{Markowitz1952} using the unbalanced panel of thousands of individual asset returns proves impracticable.
Empirical studies typically consider a "diversified" set of test assets (e.g., ME-BM 25 portfolios) to estimate and evaluate factor models, hoping these test assets or a few common factors can span the same efficient frontier as individual assets.
However, popular factor models hardly explain the cross section of conventional pre-specified test assets \citep[e.g.,][]{kozak2018interpreting,lopez2020common}, not to mention the ad hoc nature of these test assets hampers the effectiveness of model estimations and evaluations \citep{lewellen2010skeptical,Ang2020JFQA}.
For example, characteristics-based test assets are often limited to univariate- and bivariate-sorted portfolios due to the challenges of high-dimensional sorting \citep{cochrane2011presidential}, overlooking nonlinearity and asymmetric interactions (that do not uniformly apply to all assets), even with dependent sorting \citep{daniel1997measuring}.
These problems cannot be fully addressed by evaluating different test assets separately for robustness checks \citep[e.g.,][]{fama1996multifactor}, selecting test assets \citep{daniel2020cross,giglio2023test,bryzgalova2023forest}, or combining test assets \citep[e.g.,][]{feng2020taming}.

The evaluation of factor models and test assets is about mean-variance diversification and, therefore, the performance of the tangency portfolio constructed using these test assets or factors.
\citet{cochrane2011presidential} states that expected returns, variances, and covariances are stable functions of asset characteristics (e.g., size, value), and \cite{kelly2019characteristics} further show return-factor covariances are associated with characteristics.
The key to bridging the gap between the ultimate efficient frontier and tangency portfolios of test assets or factors spanning the SDF therefore lies in systematically utilizing the high-dimensional asset characteristics, which contain rich information on the joint distribution of asset returns dynamics \citep[e.g.,][]{Kozak2022when}.

To achieve this, we propose a new approach that integrates nonlinearity and asymmetric interactions with the high-dimensional characteristics to create test assets that span the efficient frontier of individual asset returns.
The Panel Tree (P-Tree) framework is inspired by modern AI and constitutes a versatile family of models that cluster panel observations to achieve given economic objectives.
Creating test assets requires clustering individual assets into groups to form portfolios.\footnote{The commonly used security sorting in empirical research is one type of unsupervised clustering based on firm characteristics, similar to the decision tree \citep{bryzgalova2023forest}. However, no unified method has been developed for sorting securities and generating test assets.} 
Under a global split criterion for goal-oriented search, P-Tree clusters individual asset returns and creates basis portfolios.
It utilizes high-dimensional characteristics under the MVE framework to jointly generate test assets and a latent factor, which is the MVE portfolio of test assets, and recovers the stochastic discount factor (SDF).
Therefore, P-Tree extends the scope of regression tree models beyond pure prediction tasks, particularly generating diversified test assets that reach the ultimate efficient frontier.\footnote{Tree-based models excel in predicting complex data with high dimensionality, nonlinearity, and variable interactions, even in low signal-to-noise environments and small sample sizes \citep[e.g.,][]{sorensen2000decision,rossi2015modeling,gu2020empirical,bianchi2021bond,capponi2024pricediscoverty}.
}

Specifically, P-Tree employs a "top-down" approach (typical trees are drawn with the root at the top), splitting the cross section of thousands of individual assets and grouping them into a small number of clusters based on characteristic values to form (value-weighted) portfolios. Guided by asset pricing considerations, we grow a P-Tree to iteratively construct test assets and latent factors for the SDF, following the baseline specification of the global split criterion that maximizes the Sharpe ratio of the SDF.
The high Sharpe ratio of the constructed SDF reflects the high MVE frontier spanned by the generated leaf basis portfolios.
The resulting leaf basis portfolios and latent factors provide researchers with a diversified set of test assets and the SDF model.
Furthermore, P-Trees are intuitive and transparent, allowing economic interpretation and identifying a sparse set of useful characteristics that interact to generate test assets and latent factors jointly.

Fundamentally, P-Tree is a greedy search algorithm guided by asset pricing goals for optimal clustering based on similar characteristic values within a large modeling space.
In empirical asset pricing, basis portfolios are usually formed by grouping assets based on classifications such as countries or industries, or through unsupervised clustering (e.g., security sorting) based on return correlations or characteristics.
In contrast, P-Tree generates characteristics-managed leaf basis portfolios to maximize the Sharpe ratio of their MVE portfolios, without being either supervised or unsupervised. The objective can be flexibly specified, making the P-Tree framework applicable for a wide range of panel data analyses.\footnote{We have developed and shared the P-Tree package, \texttt{PTree}, in a public repository in \texttt{R} for other researchers to explore (see \url{https://github.com/Quantactix/PTree}).}

\paragraph{Methodological innovations.} 
Our asset pricing application effectively demonstrates the key methodological innovations. 
First, off-the-shelf ML methods, including the famous Classification and Regression Trees \citep[CART,][]{breiman1984classification}, typically assume that observations are i.i.d., and not designed for analyzing panel data.
Though naively fitting CART or tree ensembles (boosted trees or random forest) on characteristics-return data shows positive predictive performance \citep[e.g.,][]{gu2020empirical}, they ignore the panel data structure.
Alternatively, P-Tree fully embraces the panel data structure and incorporates a time-invariant tree structure for multi-period observations. 
The time-invariant P-Tree allows for economic interpretability when building ML models on panel data. 
For example, similar to security sorting, the time-invariant P-Tree allows the same set of characteristics-managed leaf basis portfolios for all periods. 
	
Second, the standard tree-based models in ML, including CART, focus purely on prediction. Furthermore, CART-based models grow recursively, optimizing the quadratic loss of prediction. These models optimize locally at each node without considering sibling nodes, mainly for computational efficiency. However, this ``myopic" strategy often leads to overfitting, because it operates on fewer observations in each node as the tree grows.
By contrast, P-Trees broaden the applications beyond prediction, encompassing the generation of test assets and latent factors.
The iterative growth of P-Trees is designed to utilize data from the entire cross section to guard against the overfitting that afflicts conventional trees grown using local split criteria.
P-Tree combines economic principles and nonlinear ML algorithms while ensuring the interpretability of the graphical tree diagram, providing a unified approach for splitting the cross section and growing the efficient frontier.

Third, the P-Tree framework can integrate the boosting or bagging strategy of ML, enabling multiple P-Trees to form a multi-factor model. 
On the one hand, the Boosted P-Tree grows additional P-Trees based on the previous ones, providing a unified framework for a multi-factor model, such that additional factors and test assets must provide an incremental contribution. The ensemble approach makes P-Tree a versatile tool that can enhance the performance of any predetermined factor model by exploiting the unspanned efficient frontier under the MVE framework.
When boosted to generate multiple factors, P-Trees offer an alternative to principal component analysis (PCA) and deep neural networks, with greater interpretability and sparsity, while capturing (asymmetric) interactions.\footnote{Other recent studies explore latent factors on PCA \citep{kelly2019characteristics, lettau2020factors,kim2021arbitrage,he2023shrinking} and deep learning \citep[e.g.,][]{gu2021autoencoder,chen2024deep,feng2024deep}.}
On the other hand, the random P-Forest generates additional P-Trees on random bootstrap samples, offering a unified framework to create multiple sets of uncorrelated factors and test assets, which can be used to assess characteristic importance.
A large random P-Forest is also connected to the recent literature of large and over-parameterized models \citep[e.g.,][]{didisheim2024complexity}.

\paragraph{Empirical findings.} 
First, we study monthly U.S. equity returns from 1981 to 2020 using 61 firm characteristics. 
The P-Tree tangency portfolio is constructed as traded factor models, significantly advancing the efficient frontier with high annualized Sharpe ratios, ranging from 6.37 for a single P-Tree (with 10 portfolios) to 15.63 for 20 boosted P-Trees (with 200 portfolios).
These numbers are substantially higher than those constructed by conventional basis portfolios (univariate- or bivariate-sorted portfolios) or commonly used factor models. 
These findings provide strong evidence of a significant gap between the current empirical literature and the potential limits of the efficient frontier. 
Moreover, generated under the unified MVE framework, boosted P-Trees produce multi-factor models that effectively price the cross-sectional returns. Also, P-Tree factors offer annualized Sharpe ratios over 3 and significantly positive alphas in OOS analyses for past-predicting-future and future-predicting-past tests. \footnote{There is a gap between the in- and out-of-sample Sharpe ratios. The paragraph "limits to learning" in Appendix I, "Simulation," provides more discussion on this gap.}

Second, these diversified P-Tree test assets span the efficient frontier and pose a significant challenge for alternative factor models, highlighting the importance of test assets.
We identify many economically and statistically unexplained test assets, indicated by most monthly alphas larger than 1\%, and an extremely high GRS test statistic of 141.27 (aggregated weighted pricing error) for the first P-Tree against the Fama-French five-factor (FF5) model.
These asymmetric, interactive, and nonlinear characteristics-managed leaf basis portfolios are difficult to explain using other well-known observable factor models \citep[e.g.,][]{hou2021augmented} and ML-based latent factor models \citep[e.g.,][]{kelly2019characteristics, lettau2020factors}.
Given the insufficient MVE spanning and low hurdles of univariate- or bivariate-sorted portfolios for model testing, we recommend including these multi-characteristic, systematically sorted P-Tree test assets in future model evaluations.

Finally, we design a random P-Forest with bagging to evaluate characteristic importance, account for variable selection risk, and validate transparent and interpretable P-Tree models.
We confirm that the same small set of characteristics (e.g., \texttt{SUE}, \texttt{DOLVOL}, and \texttt{BM\_IA}) P-Tree selects is likely proxies for the true fundamental risk inherent in the SDF, which may be overlooked in a linear factor model with ad hoc selected factors. 
In addition, the random P-Forest SDF as a large regularized model has an excellent OOS Sharpe ratio, which shows similar patterns to the random split SDF but is significantly more efficient in computation. In a sense, goal-oriented search improves upon brute-force large models with statistical and economic regularizations, effectively combining economic objectives with ML models.
Finally, P-Trees display all split rules, aiding researchers in understanding feature interactions.\footnote{P-Trees allow the long and short legs of a long-short portfolio to interact with different characteristics, thereby loading the portfolio on different leaf basis portfolios. This asymmetric interactive sorting contrasts with the traditional treatment of a long-short portfolio as a single asset and complements the pioneering work of \cite{jarrow2023low} in modeling the two legs of anomaly portfolios separately.}

\paragraph{Literature.} 

P-Trees constitute the first goal-oriented, systematic clustering of individual securities and generation of leaf basis portfolios under the MVE framework.
\cite{hoberg2009optimized} propose constructing factors and test assets by optimizing objective functions instead of sorting.
For unsupervised clustering, \cite{ahn2009basis} model within- and between-group return correlations to form portfolio groups, \cite{chen2017hedge} evaluate the performance by grouping hedge fund alpha values, and \cite{patton2019risk} group assets based on their heterogeneous risk premia on risk factors.
We join \citet{cong2023uncommon} as the earliest studies that iteratively cluster panel data by maximizing specific economic objectives. 

Our latent factor model is related to but different from the recent regularized portfolio (or SDF) literature \citep[e.g.,][]{ao2019approaching,kozak2020shrinking,bryzgalova2023forest}, which typically estimate their regularized portfolio (or SDF) on a large number of prespecified test assets. 
In contrast, the SDF and test assets are generated iteratively in our unified P-Tree framework by maximizing the Sharpe ratio.
Notably, \cite{bryzgalova2023forest} shrink useless assets when estimating the optimal portfolio and describe this process as pruning or regularizing a decision tree (instead of growing it) working from potential leaf portfolios. They highlight the resemblance between the decision tree and security sorting but do not specify a split criterion for tree growth. Their paper manually specifies a small set of split candidates and a shallow depth for initial trees.
P-Trees differ by growing the tree from the root to provide a goal-oriented clustering approach and efficiently scanning the large space of generalized sequential sorting on high-dimensional characteristics.

Moreover, our study contributes to the growing literature on latent factor models in asset pricing.
For recent developments of PCA, in addition to the projected PCA \citep{kim2021arbitrage} and reduced-rank approach \citep{he2023shrinking},
the instrumental PCA \citep[IPCA,][]{kelly2019characteristics} 
uses characteristics to model the time-varying factor loadings and estimate principal components, and the risk premia PCA \citep[RPPCA,][]{lettau2020factors} adds asset pricing regularization in the objective.
Recent studies, such as the auto-encoder \citep{gu2021autoencoder}, generative adversarial network \citep{chen2024deep}, characteristics-sorted factor approximation \citep{feng2024deep, feng2023DTP}, and structural neural network \citep{fan2022structural}, have also developed customized deep learning models for nonlinear latent factor modeling. 
Following these developments, P-Trees provide a graphical representation for variable nonlinearity and asymmetric interactions, which PCA or deep learning methods do not offer.

While ML and tree-based methods are widely adopted in finance due to their powerful nonlinear modeling capabilities using high-dimensional characteristics \citep[e.g.,][]{gu2020empirical,bianchi2021bond,bali2023option}, most studies either apply off-the-shelf ML models for prediction without considering the panel data structure or incorporating economic guidance, or focus on supervised or unsupervised learning and prediction tasks. 
We add to several recent exceptions involving generative models, including \cite{cong2023uncommon}, which combines Bayesian spike-and-slab priors and Panel Trees for asset pricing with uncommon factors, \cite{cong2020alphaportfolio}, which develops the first "large" model in finance for portfolio management using reinforcement learning, \cite{creal2021bart}, which adapts Bayesian trees to currency returns, and \cite{cong2024mosaics}, which studies heterogeneity in return predictability and links that to trading profitability of a predictability-based anomaly.

The P-Tree framework has a notable extension, the random P-Forest, which connects to recent discussions on ``benign overfitting'' and ``high-dimensional interpolation'' \citep[e.g.,][]{belkin2019reconciling,hastie2022surprises} in statistics , as well as the corresponding ``virtue of complexity'' in financial contexts \citep[e.g.,][]{kelly2022virtueEverywhere,kelly2024virtue,didisheim2024complexity}.
    We corroborate these studies by showing that large tree-based models for which the number of parameters exceeds the number of observations perform better OOS, provided that appropriate statistical regularization is applied.
Moreover, we contribute to recent research that supports incorporating economic restrictions when estimating and evaluating machine learning or factor models \citep[e.g.,][]{gagliardini2016time, avramov2023machine, jensen2024machine}. 

Finally, our tree-based greedy algorithm demonstrates human-like intelligence through a ``divide-and-conquer'' strategy, offering a sparse, interpretable, and computationally efficient modeling alternative to modern AI, distinct from deep reinforcement learning.

The remainder of the article is as follows: 
Section \ref{sec:model} introduces the P-Tree models.
Section \ref{sec:empirical_example} illustrates the empirical applications of a single P-Tree to split the cross-section and generate test assets.
Section \ref{sec:empirical_boost} demonstrates the empirical results of the boosted and multi-factor P-Trees.
Section \ref{sec:chars_and_macro} discusses the random P-Forest and P-Tree's robustness to macro regimes.
Section \ref{sec:conclusion} concludes, and the internet appendix includes simulation and additional empirical discussions.

\section{Panel Tree for Asset Pricing } \label{sec:model}

Section \ref{sec:cart} describes how P-Tree innovates over standard tree-based models. Section \ref{sec:tree_growth} delves into the growth of P-Trees, and Section \ref{sec:boosting} introduces boosted P-Trees for multi-factor models. 
Section \ref{sec:main_text_simulation} presents simulation results.

\subsection{CART and P-Tree Innovations} \label{sec:cart}

Designed for predictions, the standard CART \citep{breiman1984classification} model and its variants partition the predictor space into distinct, non-overlapping regions of leaf nodes and assign a constant leaf parameter to each region.\footnote{CART is a binary decision tree model and serves as the foundation for ensemble methods such as random forest \citep{breiman2001random} and boosting trees \citep{freund1997decision}. Other notable Bayesian tree models include BART \citep{chipman2010bart} and XBART \citep{he2019xbart,he2021stochastic}.} 
It searches and records the partition as follows: Suppose the data constitute $K$ predictors, and the $i$-th observation is denoted by $\mathbf{z}_i = (z_{i,1}, \cdots, z_{i,K})$. The $j$-th split rule of the decision tree is denoted by $\tilde{\mathbf{c}}^{(j)} = (z_{\cdot,k}, c_j)$, which partitions data by checking whether the value of the $k$-th predictor $z_{\cdot,k}$ is greater or smaller than cutoff $c_j$. 
CART considers only binary splits, since any multiway-split tree can be represented as one with multiple binary splits. The optimal split rule is chosen to minimize the prediction error.

After $J$ splits, CART partitions the predictor space into $J+1$ leaf nodes denoted by $\{\mathcal{R}_n\}_{n=1}^{J+1}$ and assigns a constant leaf parameter $\mu_n$ to each node.
The regression tree $\T$, with parameters $\Theta_J = \{ \{\tilde{\cc}^{(j)}\}^{J}_{j=1}, \{\mu_n\}^{J+1}_{n=1} \}$, constitutes a high-dimensional step function:
\begin{equation}
    \T(\z_i \mid \Theta_J) = \sum^{J+1}_{n=1}\mu_n \mathbb{I}\left\{\z_i \in \R_j\right\}.\label{eqn:tree}
\end{equation} 
The indicator $\mathbb{I}\{\mathbf{z}_i \in \mathcal{R}_j\}$ takes $1$ for one leaf node and $0$ for others. The leaf parameters in the step function are estimated by averaging the training data within each leaf node. This nonparametric approach adapts CART to low signal-to-noise environments and small sample sizes. The tree model predicts new observations by locating the corresponding leaf and using its parameter as the prediction after training.

\begin{figure}[!h]
	\caption{\bf Example of a Decision Tree}\label{fig:tree_example}
	\footnotesize{
		Left: A decision tree has two splits, three leaf nodes, and three leaf parameters. Right: The corresponding partition plot is generated for the sample of predictor space on value and size. 
	}
	\bigskip
	        
		\begin{subfigure}{0.45\textwidth}
			\begin{center}
				\begin{tikzpicture}[
						scale=0.8,
						node/.style={%
							draw,
							rectangle,
						},
						node2/.style={%
							draw,
							circle,
						},
					]
					
					\node [node] (A) {$\texttt{size}<0.5$};
					\path (A) ++(-135:\nodeDist) node [node2] (B) {$\mu_{1}$};
					\path (A) ++(-45:\nodeDist) node [node] (C) {$\texttt{value}<0.7$};
					\path (C) ++(-135:\nodeDist) node [node2] (D) {$\mu_{2}$};
					\path (C) ++(-45:\nodeDist) node [node2] (E) {$\mu_{3}$};
					
					\draw (A) -- (B) node [left,pos=0.5] {Yes}(A);
					\draw (A) -- (C) node [right,pos=0.5] {No}(A);
					\draw (C) -- (D) node [left,pos=0.5] {Yes}(A);
					\draw (C) -- (E) node [right,pos=0.5] {No}(A);
				\end{tikzpicture}
			\end{center}
		\end{subfigure}
		\hfill
		\begin{subfigure}{0.45\textwidth}
			\begin{center}
				\begin{tikzpicture}[scale=3]
					\draw [thick, -] (0,1) -- (0,0) -- (1,0) -- (1,1)--(0,1);
					\draw[decorate sep={1mm}{40mm},fill] (0.97, 0.61);
					\draw[decorate sep={1mm}{40mm},fill] (0.24, 0.24);
					\draw[decorate sep={1mm}{40mm},fill] (0.78, 0.77);
					\draw[decorate sep={1mm}{40mm},fill] (0.92, 0.43);
					\draw[decorate sep={1mm}{40mm},fill] (0.61, 0.06);
					\draw[decorate sep={1mm}{40mm},fill] (0.65, 0.53);
					\draw[decorate sep={1mm}{40mm},fill] (0.05, 0.45);
					\draw[decorate sep={1mm}{40mm},fill] (0.56, 0.21);
					\draw[decorate sep={1mm}{40mm},fill] (0.06, 0.11);
					\draw[decorate sep={1mm}{40mm},fill] (0.85, 0.73);
					\node at (0.5,-0.2) {$\texttt{size}$};
					\node at (-0.3,0.5) {$\texttt{value}$};
					\draw [thick, -] (0.5, 0) -- (0.5, 1);
					\draw [thick, -] (0.5, 0.7) -- (1, 0.7);
				\end{tikzpicture}
			\end{center}
		\end{subfigure}
\vspace{-0.4cm}
\end{figure}

Figure \ref{fig:tree_example} illustrates how conventional tree-based models (e.g., CART) are applied to predict stock returns with firm characteristics, through recursively partitioning the characteristic space into clusters (small-cap, large-cap growth, and large-cap value). 
The bottom nodes, often called leaf nodes, are associated with constant leaf parameters, meaning assets within the same node share the same return forecast.

However, due to multiple limitations, these methods are unsuitable for economic or financial panel data and for generating test assets or pricing kernels. 
First, they treat the panel of asset returns as i.i.d observations. 
Second, they focus on predictions using statistical local split criteria, such as minimizing the sum of squared error objectives in the child nodes of a particular split. Thus, they do not incorporate panel structure or economic restrictions. 
As the tree grows, the number of observations in each node decreases, and idiosyncratic noise causes overfitting. While tree ensembles help mitigate overfitting, they are less interpretable than a single decision tree.

P-Tree addresses both issues by (i) utilizing a time-invariant tree structure and (ii) employing economic objectives that consider all observations, rather than just those in a parent node, to guide the tree growth.
This (i) enables thorough extraction of panel data to construct leaf basis portfolios and 
(ii) iteratively builds P-Trees using global split criteria to prevent overfitting, extending the scope of trees from pure prediction to goal-oriented clustering for test asset and factor generation.

\subsection{Growing a P-Tree} \label{sec:tree_growth}
P-Tree partitions the universe of individual assets into non-overlapping leaf nodes based on the values of ranked characteristics with a time-invariant structure over $T$ periods.
The leaf basis portfolios are (value-weighted) portfolios of stocks within each leaf node from the time-invariant tree structure. 
P-Tree splits sequentially from the root node and generates one extra leaf basis portfolio after each additional split. Specifically, it produces $j+1$ leaf basis portfolios after the $j$-th split, reducing the dimension from thousands of individual assets to $j+1$ portfolios.
These characteristics-managed leaf basis portfolios expand on characteristics-sorted portfolios to accommodate asymmetric interactions of multiple characteristics.

Let $\RR^{(j)}_t$ denote all excess return vectors of leaf basis portfolios after the $j$-th split of the tree, and let $f_t^{(j)}$ denote the factor spanned by $\RR^{(j)}_t$. The tree has $j+1$ basis portfolios in all leaf nodes, denoted by $\RR^{(j)}_t = [R_{1,t}^{(j)}, \cdots, R_{n,t}^{(j)}, \cdots, R_{j+1,t}^{(j)}]$, where $R_{n,t}^{(j)}$ represents a length-$T$ vector of returns over $T$ periods for the $n$-th leaf node.
To find the first split, we begin with a portfolio of all assets, denoted as $\mathbf{R}_t^{(0)}=\left[R_t^{(0)}\right]$, which serves as the root node for P-Tree. The tree is expanded through iterative updates of $\left\{\RR_t^{(j)}, f_t^{(j)}\right\}$.
First, the leaf basis portfolios $\RR_t^{(j)}$ are expanded as the tree grows.
Second, using the expanded leaf basis portfolios, we re-estimate the P-Tree factor $f_t^{(j)}$. Being a greedy algorithm, the sequential growth of the tree ensures computational feasibility compared to a full enumeration of possible sortings.


\begin{figure}[h!]
	\caption{\bf  Illustration of the First Split}
 	\label{fig:firstsplit}
	\footnotesize{
		To determine the optimal characteristic and cutpoint value, we evaluate a list of candidate splits (i.e., standardized unexpected quarterly earnings (\texttt{SUE}) $\leq -0.2$; 40\% percentile) for calculating the split criterion value.
	}
	\bigskip
		
	\centering
	\begin{subfigure}{0.66\textwidth}
		\begin{center}
			\begin{tikzpicture}[
					scale=1,
					node/.style={%
						draw,
						rectangle,
					},
					node2/.style={%
						draw,
						circle,
					},
				]
				
				\node [node] (A) {\footnotesize{$R_t^{(0)}: \texttt{SUE} \leq -0.2$}};
				\path (A) ++(-135:\nodeDist) node [node2] (B) {\footnotesize{$R_{1,t}^{(1)}$}};
				\path (A) ++(-45:\nodeDist) node [node2] (C) {$R_{2,t}^{(1)}$};
				
				\draw (A) -- (B) node [left,pos=0.5] {Yes}(A);
				\draw (A) -- (C) node [right,pos=0.5] {No}(A);
			\end{tikzpicture}
		\end{center}
	\end{subfigure}
\vspace{-0.4cm}
\end{figure}

\paragraph{First split.}  
The root node contains all assets corresponding to the value-weighted market factor. Firm characteristics are uniformly normalized cross-sectionally to the range of $[-1,1]$ within each period. We assess different split threshold options $c_m$ for characteristics $z_{\cdot,k}$, such as quintile splits $c_m \in [-0.6, -0.2, 0.2, 0.6]$.\footnote{Quintile splitting efficiently reduces the search space with only about four thousand stocks and many highly correlated characteristics for growing a P-Tree with 10 leaves. Decile or denser splitting may lead to overfitting and non-diversified test assets due to some portfolios having few stocks.
} 
Consider a candidate split $\tilde{\cc}_{k,m} = (z_{\cdot,k}, c_m)$ as in Figure \ref{fig:firstsplit}. The candidate partitions the root node into left and right child nodes based on whether $z_{\cdot,k}$ is less than or equal to $c_m$.

The observations of stock returns within each potential child leaf form a leaf basis portfolio, with vectors of return $R_{1,t}^{(1)}$ and $R_{2,t}^{(1)}$, respectively. Note they are vectors of returns, thus maintaining the panel structure of asset returns. The P-Tree factor is estimated as the MVE portfolio of all leaf basis portfolios,
\vspace{-0.2cm}
\begin{equation}\label{eqn:SDF}
	f_t^{(1)} = \w^{(1)\prime} \RR_{t}^{(1)}, \quad \w^{(1)} \propto 
    \hat E[\RR_t^{(1)} \RR_t^{(1)\prime}]^{-1} 
    \hat E[\RR_t^{(1)}],
\end{equation}
where $\RR_t^{(1)}=\left[R_{1,t}^{(1)}, R_{2,t}^{(1)}\right]$ is the matrix of returns of two leaf basis portfolios after the first split, $\hat E[\RR^{(1)}]$ 
and
$\hat E[\RR_t^{(1)} \RR_t^{(1)\prime}]$ 
are the sample mean and second moment matrix of leaf basis portfolios.\footnote{
Portfolio weights $\w^{(j)}$ are normalized such that the sum of absolute weights equals one.
Following the regularization approaches in \citep{kozak2020shrinking, bryzgalova2023forest, didisheim2024complexity}, we utilize small shrinkage parameters, $\gamma = 10^{-4}$, for robustly estimating efficient portfolio weights:
\vspace{-0.2cm}
\begin{equation} \label{eqn:robust_weight}
		\w^{(j)} = 
		\left(\hat E[\RR_{t}^{(j)} \RR_{t}^{(j)\prime}] + \gamma \mathbf{I}_{k+1} \right)^{-1} 
		\hat E[\RR_{t}^{(j)}],
\end{equation}
where $\mathbf{I}_{k+1}$ is the identity matrix. Shrinkage parameters stabilize portfolio weight estimation and prevent over-leveraging, which can be adjusted to control the degree of regularization.}
Generating latent factors aligns with the P-Tree split criteria, as the P-Tree factor represents the MVE portfolio of all leaf basis portfolios.

In the baseline specification, our split criterion therefore aims to maximize the Sharpe ratio of the MVE portfolio:
\begin{equation} \label{eqn:criteria}
	\mathcal{L}(\tilde{\cc}_{k,m}) = \sqrt{ \boldsymbol{\hat \mu}_{\F}'\boldsymbol{\hat \Sigma}_\F^{-1}\boldsymbol{\hat \mu}_\F}, 
\end{equation}
where $\F = f_t^{(1)}$ is the generated latent factor, with the sample mean $\boldsymbol{\hat \mu}_F$ and covariance matrix $\boldsymbol{\hat \Sigma}_\mathbf{F}$.
The criterion allows P-Tree to construct and estimate basis portfolios, latent factors, and efficient portfolio weights simultaneously, for the global objective of constructing the efficient frontier.

Each candidate split $\tilde{\cc}_{k,m}$ generates a different partition of the data, resulting in unique leaf basis portfolios, corresponding P-Tree factors, and ultimately, varying valuations of the split criteria in (\ref{eqn:criteria}). Consequently, we loop over all candidate splits and select the one that maximizes the split criteria as our initial split rule.


\paragraph{Second split.} 
The second split can occur at the root's left or right child node. We assess the split criteria for all candidate splits for \textit{both} leaf nodes and choose the split that maximize the criteria as in (\ref{eqn:criteria}). Figure \ref{fig:secondsplit} depicts the tree of the candidates for the second split. In either scenario, one leaf node splits, becoming an internal node and generating two new leaf nodes. The P-Tree factor is then constructed based on \textit{all three} leaf basis portfolios (thus global):
\vspace{-0.2cm}
\begin{equation}
	f_t^{(2)} = \w^{(2)\prime} \RR_{t}^{(2)}, \quad \w^{(2)} \propto 
    \hat E[\RR_t^{(2)} \RR_t^{(2)\prime}]^{-1} 
    \hat E[\RR_t^{(2)}],
\end{equation}
where $\hat E[\RR^{(2)}]$ and $\hat E[\RR_t^{(2)} \RR_t^{(2)\prime}]$ are the sample mean and second moment matrix of leaf basis portfolios $\RR_{t}^{(2)} = \left[R_{1,t}^{(2)}, R_{2,t}^{(2)}, R_{3,t}^{(2)}\right]$ after the second split. The construction of the three basis portfolios depends on which node the candidate splits, as shown in Figure \ref{fig:secondsplit}. The updated P-Tree factor is plugged in to maximize the Sharpe ratio in \eqref{eqn:criteria}, where $\F = f_t^{(2)}$ is the updated P-Tree factor.

Notably, our proposed objective is a \textit{global} split criterion, because it considers all leaf basis portfolios when constructing $\mathbf{F}$ (the global MVE portfolio). Unlike CART, which focuses on a specific leaf node, our model explores all candidate splits in all leaf nodes to find the one with the largest investment improvement.

\begin{figure}[!ht]
	\caption{\bf Illustration of the Second Split}
 	\label{fig:secondsplit}
	\footnotesize{
            There are two potential candidates for the second split, each splitting one of the original leaf nodes. After the second split, three leaf basis portfolios are produced to update the latent factor.
	}
	\bigskip
		
	\centering
	\begin{subfigure}{0.48\textwidth}
		\begin{center}
			\begin{tikzpicture}[
					scale=1,
					node/.style={%
						draw,
						rectangle,
					},
					node2/.style={%
						draw,
						circle,
					},
				]
				
				\node [node] (A) {\footnotesize{$R_t^{(0)}: \texttt{SUE} \leq -0.2$}};
				\path (A) ++(-135:\nodeDist) node [node] (B) {\footnotesize{$R_{1,t}^{(1)}: \texttt{DOLVOL} \leq -0.6$}};
				\path (A) ++(-45:\nodeDist) node [node2] (C) {$R_{3,t}^{(2)}$};
				\path (B) ++(-135:\nodeDist) node [node2] (D) {$R_{1,t}^{(2)}$};
				\path (B) ++(-45:\nodeDist) node [node2] (E) {$R_{2,t}^{(2)}$};
				
				\draw (A) -- (B) node [left,pos=0.5] {Yes}(A);
				\draw (A) -- (C) node [right,pos=0.5] {No}(A);
				\draw (B) -- (D) node [left,pos=0.5] {Yes}(A);
				\draw (B) -- (E) node [right,pos=0.5] {No}(A);
			\end{tikzpicture}
		\end{center}
		\caption{Splitting node \footnotesize{$R_{1,t}^{(1)}$ at $\texttt{DOLVOL}$}.}
	\end{subfigure}
	\begin{subfigure}{0.48\textwidth}
		\begin{center}
			\begin{tikzpicture}[
					scale=1,
					node/.style={%
						draw,
						rectangle,
					},
					node2/.style={%
						draw,
						circle,
					},
				]
				
				\node [node] (A) {\footnotesize{$R_t^{(0)}: \texttt{SUE} \leq -0.2$}};
				\path (A) ++(-135:\nodeDist) node [node2] (B) {$R_{1,t}^{(2)}$};
				\path (A) ++(-45:\nodeDist) node [node] (C) {\footnotesize{$R_{2,t}^{(1)}: \texttt{DOLVOL} \leq -0.6$}};
				\path (C) ++(-135:\nodeDist) node [node2] (D) {$R_{2,t}^{(2)}$};
				\path (C) ++(-45:\nodeDist) node [node2] (E) {$R_{3,t}^{(2)}$};
				
				\draw (A) -- (B) node [left,pos=0.5] {Yes}(A);
				\draw (A) -- (C) node [right,pos=0.5] {No}(A);
				\draw (C) -- (D) node [left,pos=0.5] {Yes}(A);
				\draw (C) -- (E) node [right,pos=0.5] {No}(A);
			\end{tikzpicture}
		\end{center}
		\caption{Splitting node \footnotesize{$R_{2,t}^{(1)}$ at $\texttt{DOLVOL}$}.}
	\end{subfigure}
\vspace{-0.4cm}
\end{figure}

\paragraph{Growth termination.}
All subsequent splits proceed similarly. The tree-growing procedure is outlined in Algorithm \ref{alg:GT}.
Determining the number of leaves is a natural turning point in tree growth, and it is also the only tuning parameter needed for P-Tree. 
We consider P-Trees with $J+1=10$ leaf nodes in the baseline specification. 
Furthermore, we mandate a minimum leaf size of 20 for growing the tree because the leaf needs to serve as a basis portfolio, and any leaves that do not meet this criterion are not subjected to further splitting.
Once the tree growing process terminates, it outputs the P-Tree split sequence, leaf basis portfolios $\mathbf{R}_t^{(J)}=[R_{1,t}^{(J)},\cdots,R_{J+1,t}^{(J)}]$, and the P-Tree factor $f_t^{(J)}$. Note each leaf portfolio can be interpreted through economic fundamentals revealed by the sequential splits.

\paragraph{Possible extensions.} 
The P-Tree framework is flexible to accommodate alternative objectives to the global split criteria in \eqref{eqn:criteria} for estimating the MVE portfolio using individual asset returns.
For example, one might aim to create the minimum-variance portfolio or develop latent factors explaining test assets, such as individual or basis portfolio returns. The crucial aspect is defining a clear economic goal and utilizing the greedy growth algorithm within trees to explore the extensive observation clustering space and optimize the objective.
P-Trees also allow for flexible greediness, enabling the consideration of multiple splits in each iteration instead of the conventional single split, or simultaneous splits along multiple characteristics by partitioning along some linear combinations of the characteristics. 
We leave these for future exploration.

\subsection{Boosted P-Trees} 
\label{sec:boosting}
Boosting is an ML technique that combines weak learners to form a strong learner \citep[e.g.,][]{freund1997decision,rossi2015modeling,capponi2024pricediscoverty}. 
Boosted P-Trees sequentially grow a list of additive trees that augment previous trees under the MVE framework, which helps further span the efficient frontier with additional basis portfolios and generate multi-factor models. Considering existing factors, the boosting procedure seamlessly integrates into the P-Tree framework by creating additional P-Tree factors to maximize the collective Sharpe ratio of the tangency portfolio. The boosting procedure is outlined below.
\begin{enumerate}
	\item The initial factor, denoted as $f_{1,t}$, can be either generated by the first single P-Tree or benchmark factors selected by researchers (e.g., the market factor).

	\item The second factor $f_{2,t}$ is generated by the second P-Tree to complement the first factor. The second tree growth follows the split criterion in \eqref{eqn:criteria}, but $\F=[f_{1,t}, f_{2,t}]$. The MVE portfolio is generated directly from all the factors. \footnote{We use a small shrinkage parameter, $\gamma_f = 10^{-5}$, for estimating MVE portfolio weights on factors.}
	      		
	\item Repeat Step 2 to sequentially generate $P$ factors $\F_t = [f_{1,t}, \cdots, f_{P,t}]$ until the stopping criteria are triggered. 
 
\end{enumerate}

\paragraph{Block structure in boosted P-Trees.}
The boosted P-Tree utilizes a two-step process to produce either the MVE portfolio or the SDF by incorporating multiple sets of leaf basis portfolios.
Initially, the tree-specific MVE portfolio of all leaf basis portfolios within each P-Tree is computed to determine the current P-Tree factor. 
Subsequently, this factor is merged with all previous P-Tree factors or benchmarks to create an all-tree MVE portfolio for the multi-factor P-Tree model.
Boosted P-Trees indirectly impose regularization on the all-tree MVE portfolio weights through the sequential tree block structure, even though the resulting Sharpe ratio may not surpass the one generated directly from all basis portfolios in all P-Trees.
This avoids high-dimensional estimation issues and, in turn, leads to robust estimation and OOS performance.


\subsection{Simulation} \label{sec:main_text_simulation}
Simulation exercises in this section highlight the importance of using out-of-sample tests and of including most available observable features when applying P-Tree, as well as demonstrate the ability of P-Tree to capture true underlying characteristics generating the data and their non-linear interactions. For more details, please refer to Internet Appendix \ref{sec:simulation}.

We introduce three true underlying characteristics, their interactions, and nonlinear terms (detailed in Internet Appendix \ref{sec:simulation}). We calibrate the return-generating process on historical data (introduced in the next section) and include a large set of redundant characteristics correlated with the true characteristics generating the data, but do not drive returns. We then conduct the following four sets of exercises.

First, we compare the Sharpe ratios and alphas of P-Tree test assets with conventional characteristics-sorted portfolios, which do not incorporate characteristic interactions. 
The single P-Tree and boosted P-Trees consistently produce leaf basis portfolios with higher Sharpe ratios and larger unexplained pricing errors under various signal-to-noise-ratio levels, both in- and out-of-sample. 
Even when sorting is performed on the true characteristics directly, univariate- and bivariate-sorted portfolios still underperform P-Tree test assets generated on high-dimensional characteristics.
This result is corroborated later in our empirical findings that P-Tree test assets achieve better mean-variance efficiency than sorted portfolios (Section \ref{sec:p_tree_test_asset}). P-Tree indeed captures important interactions in the data-generating process.

Second, the interpretability of one P-Tree or a few P-Trees relies on their ability to select variables that matter. P-Tree helps distinguish sparse characteristics underlying the true data generation from a large number of redundant or useless ones. With repeated simulations, P-Tree selects the true characteristics in the first few splits with high probabilities.
These improved selection rates are particularly evident in scenarios with a high signal-to-noise ratio. This discovery can only be discerned through simulations, as real-world data's sparse characteristics are still unknown. The consistent selection behavior relates to assessing characteristic importance in Section \ref{sec:random_p_forest}.

Third, we break down the difference between the in- and out-of-sample performance of P-Tree models into two components: \textit{overfitting} and \textit{limits to learning}, as discussed by \cite{didisheim2024complexity}. 
Estimating the \textit{true predictability} is challenging in real data due to the unknown market data-generating process. In our simulation, however, the true predictability is defined as the OOS Sharpe ratio achieved by P-Trees with oracle characteristics.
We find positive values for overfitting and limits to learning, which increase with more boosted P-Trees, supporting the conclusion in  \cite{didisheim2024complexity} that overfitting and limits to learning become more problematic with additional model parameters and a limited number of observations. Later in empirical analyses, we observe again the large gap between the in- and out-of-sample performance of P-Tree models, justifying our focus on OOS metrics.

Fourth, we evaluate the efficiency loss when a P-Tree misses some true characteristics among input features (but redundant or useless characteristics abound).
This exercise evaluates the model's performance with incomplete or useless information, which may occur for various reasons.
Under our calibrated return-generating process, the Sharpe ratios and CAPM alphas decline substantially, and P-Tree selects redundant or useless characteristics more frequently. Therefore, we should include all available predictors. 
In our later analyses of historical data, we indeed use all firm characteristics available in the dataset.

\section{Splitting the Cross Section of U.S. Equities}
\label{sec:empirical_example}

The P-Tree framework easily applies to public equities or corporate bonds. We focus on U.S. equities for an illustrative application. Notably, against the backdrop of emergent large models, P-Tree offers an interpretable alternative that does not require excessive computation resources.\footnote{The baseline single P-Tree model trained on U.S. data runs about 20 minutes on a server with an Intel Xeon Gold 6230 CPU, for a training dataset with 61 characteristics and 2.2 million observations.}

\subsection{Data on U.S. Public Equities}\label{sec:emp_data}

\paragraph{Equity data and characteristics.} 
The standard filters (e.g., same as in Fama-French factor construction) are applied to the universe of U.S. equities. This universe includes only stocks listed on NYSE, AMEX, or NASDAQ for more than one year and uses those observations for firms with a CRSP share code of 10 or 11. We exclude stocks with negative book equity or lag market equity. We use 61 firm characteristics with monthly observations for each stock, covering six major categories: momentum, value, investment, profitability, frictions (or size), and intangibles. Characteristics are standardized cross-sectionally to the range $[-1,1]$. Table \ref{tab:chars} lists these input variables.\footnote{For example, market equity values in Dec. 2020 are uniformly standardized into $[0, 1]$. The firm with the lowest value is 0, and the highest is 1. All others are distributed uniformly in between. 
Missing values of characteristics are imputed as 0.5, which implies the firm is neutral in the security sorting.}

The monthly data ranges from 1981 to 2020. The average and median monthly stock observations are 5,265 and 4,925 for the first 20 years, and 4,110 and 3,837 for the latter 20 years. We apply cross-sectional winsorization at 1\% and 99\% to mitigate the impact of outliers on individual stock returns. The entire 40-year sample is used for benchmark analysis. The sample is split into two parts—the first 20 years from 1981 to 2000 and the recent 20 years from 2001 to 2020—for a subsample robustness check and as training and test samples.

\paragraph{Macroeconomic variables.} 
In addition, we use 10 macro variables to capture potential regime switches. Table \ref{tab:macro} summarizes the macro variables, which include market timing macro predictors, bond market predictors, and aggregate characteristics for S\&P 500. We standardize these macro predictor data by the historical percentile numbers for the past 10 years.\footnote{For example, inflation greater than 0.7 implies the current inflation level is higher than 70\% of observations during the past decade.} This rolling-window data standardization is useful when comparing the predictor level to detect different macroeconomic regimes.

\subsection{Visualizing a Single P-Tree}  \label{sec:visualizing}
\begin{figure}[!h]
    \caption{\bf Panel Tree from 1981 to 2020}
    {\footnotesize
    We provide splitting characteristics, cutpoint values for each parent node, and their respective node and splitting indexes. For example, node N1 is split by \texttt{SUE} at -0.2 (40\% perentile) as the first split S1, and the second split S2 is on node N3 by \texttt{DOLVOL} at -0.6 (20\% percentile). The median monthly number of assets in the terminal leaf basis portfolios is also included. For example, node N8 has 134 stocks by monthly median. Table \ref{tab:chars} describes equity characteristics. Figure \ref{fig:tree_a_T1_T2} reports the P-Tree diagrams for subsamples.
    }
    \label{fig:tree_a}
    \begin{center}
        \includegraphics[width=0.95\textwidth]{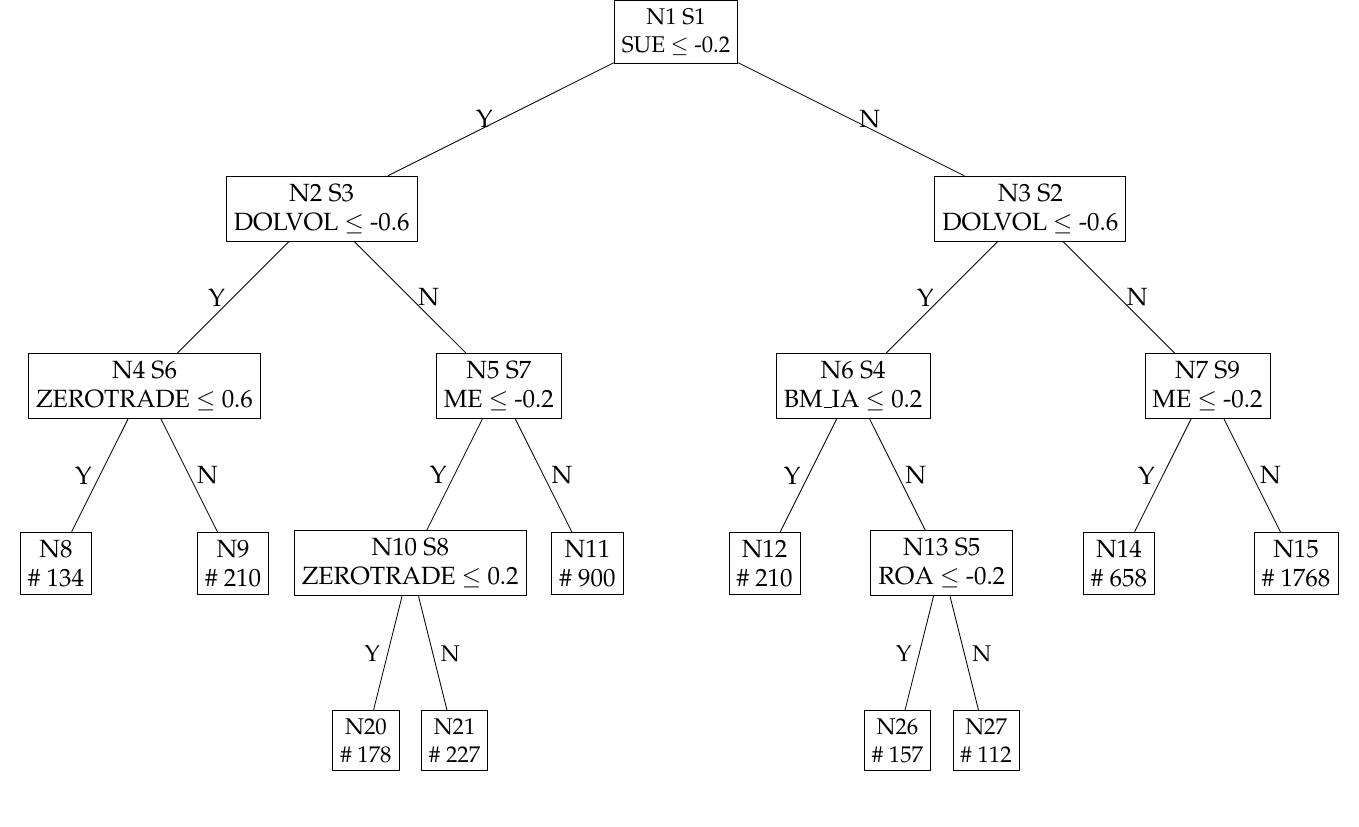}    
    \end{center}
\vspace{-0.6cm}  
\end{figure}

Figure \ref{fig:tree_a} plots the P-Tree diagram. In each leaf node, S\# represents the order of sequential splits, and N\# is the node index.
We provide the numbers in the terminal leaves for the portfolio size: the monthly median number of stock observations. 
The split rules are the selected splitting characteristics and cross-sectional quintile cutpoints $[-0.6, -0.2, 0.2, 0.6]$.
Before the first split, the tree grows from the root node (N1), whose leaf basis portfolio represents the value-weighted market portfolio.

The data-driven P-Tree first splits along the standardized unexpected quarterly earnings 
\citep[\texttt{SUE},][]{rendleman1982empirical}
at -0.2 (40\% percentile).
After this split, 40\% of the stocks go to the left leaf (labeled N2), and 60\% go to the right (N3).
Then, the second split is on the dollar trading volume 
\citep[\texttt{DOLVOL},][]{chordia2001trading}
at -0.6 of the right leaf (N3), and the third split is also on \texttt{DOLVOL} at -0.6 of the left leaf (N2).
Furthermore, subsequent splits include the industry-adjusted book-to-market ratio (\texttt{BM\_IA}), return-on-equity (\texttt{ROA}), zero trade (\texttt{ZEROTRADE}), and market equity (\texttt{ME}).
After nine splits, we stop the P-Tree growth and obtain 10 leaf basis portfolios.

\paragraph{Clustering patterns and asymmetric interactions.} 
P-Tree clusters similar assets based on underlying characteristics, revealing sources of mean-variance diversification. Figure \ref{fig:tree_a} shows the (asymmetric) interactions of characteristics for splitting the cross section.
By jointly defining the partition corresponding to the leaf node, P-Tree learns the interaction of characteristics appearing in the same path. 
For instance, \texttt{ZEROTRADE} of liquidity \citep{liu2006liquidity} is a valuable indicator for further splitting low-\texttt{SUE} low-\texttt{DOLVOL} stocks. 
However, for low-\texttt{SUE} non-low-\texttt{DOLVOL} stocks, \texttt{ME} of size \citep{banz1981relationship} might be a better indicator for a further split under the MVE framework.

\begin{figure}[h!]
	\caption{\bf Visualizing Nonlinear Interactions with Partition Plots
     }
	\label{fig:partition_a}
	\footnotesize{
		This diagram illustrates the partitions for the first few splits of the tree structure in Figure \ref{fig:tree_a}. The first split (S1) occurs at 40\% of \texttt{SUE} on the entire stock universe, and the second split occurs at 80\% of \texttt{DOLVOL} on the high \texttt{SUE} partition. The portfolio results for each partition are provided (monthly average excess returns (AVG) and annualized Sharpe ratios (SR)). The arrows indicate the next split is implemented on the partitioned area from the previous one.
	}
\vspace{-0.4cm}
	\begin{center}
		\includegraphics[width=0.95\textwidth]{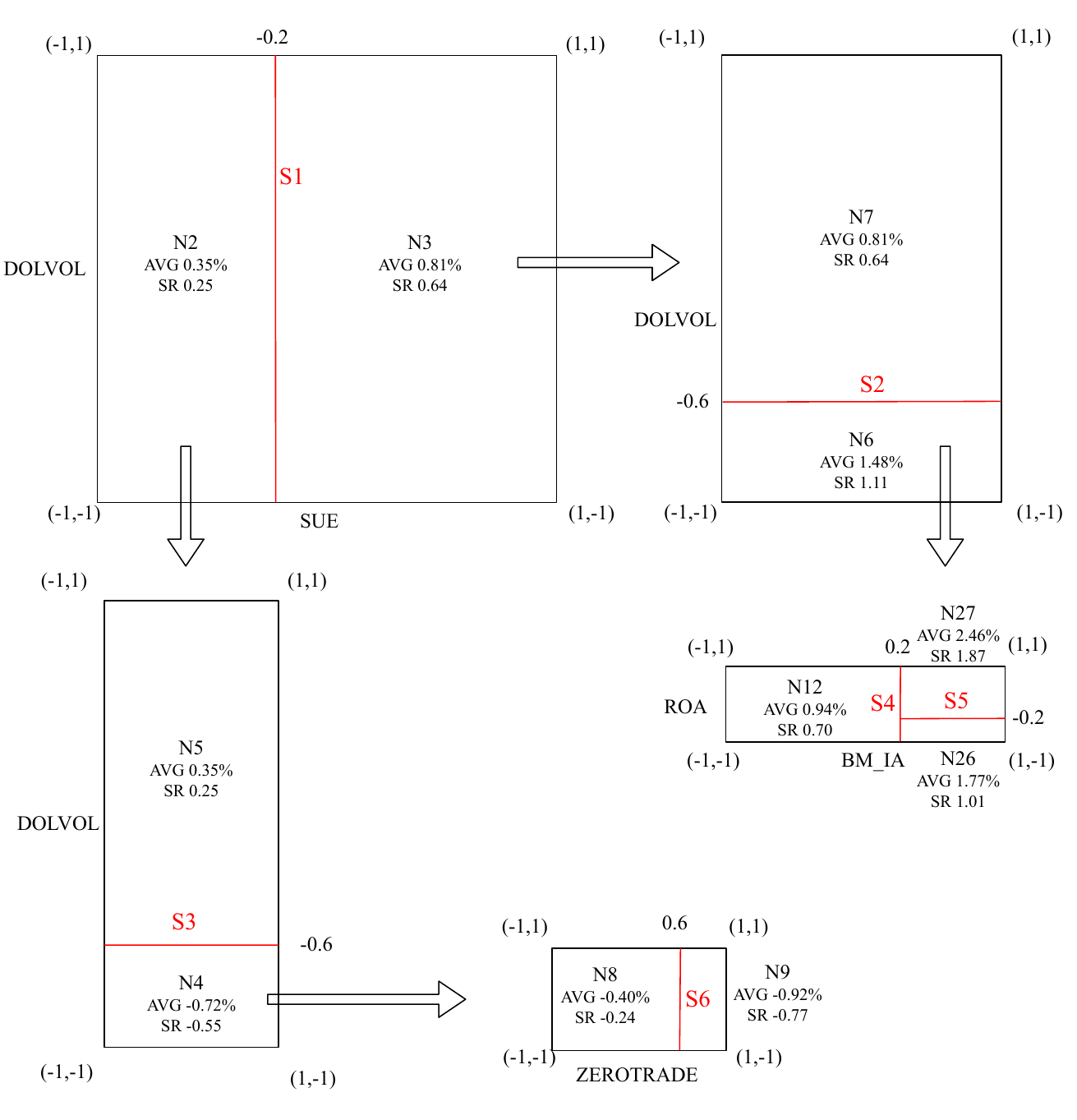}
	\end{center}
\vspace{-0.4cm}
\end{figure}

The current literature has recognized interactions between two characteristics, but a more systematic investigation is needed. 
For instance, \cite{lee2000price} enhance momentum strategies by interacting with trading volume, and \cite{da2014frog} find that the momentum effect is stronger for firms with continuous information than firms with discrete information.
Our framework allows for exploring interactions among multiple characteristics, going beyond the typical bivariate relations. This is demonstrated in Figure \ref{fig:tree_a}, where interaction paths involving at least four characteristics, such as \texttt{SUE}, \texttt{DOLVOL}, \texttt{ME}, and \texttt{ZEROTRADE}, can be identified.

The partition plot in Figure \ref{fig:partition_a} is an alternative way to visualize the clustering and asymmetric patterns. We report each leaf's monthly average excess returns and annualized Sharpe ratios. First, P-Tree splits on \texttt{SUE} at -0.2, which yields a low-\texttt{SUE} portfolio including 40\% of the stocks and a high-\texttt{SUE} portfolio constituted by 60\% of the stocks in the cross section. The spread of monthly expected returns between N2 and N3 is 0.46\%, and the Sharpe ratios range from 0.25 to 0.64.

Second, we split on \texttt{DOLVOL} at -0.6 on N3 to harvest N6 and N7, where the low-\texttt{DOLVOL} portfolio N6 has \textit{higher} expected returns and Sharpe ratio than N7.
Third, we split on \texttt{DOLVOL} at -0.6 on N2 to harvest N4 and N5, where the low-\texttt{DOLVOL} portfolio N4 has \textit{lower} expected returns and Sharpe ratio than N5.
We find \texttt{DOLVOL} has different impacts on N2 and N3. On the high-\texttt{SUE} side, \texttt{DOLVOL} positively correlates to asset returns. However, on the low-\texttt{SUE} side, \texttt{DOLVOL} negatively correlates with asset returns. 
This is an example of asymmetric interaction between \texttt{SUE} and \texttt{DOLVOL}. A simple trading strategy that shorts N4 (Low-\texttt{SUE} -Low-\texttt{DOLVOL} portfolio) and longs N6 (High-\texttt{SUE} - Low-\texttt{DOLVOL} portfolio) makes over 2\% monthly expected return. The return gaps among the leaves show the usefulness of splitting the cross section via the asymmetric interaction of characteristics.

\subsection{P-Tree Leaf Basis Portfolios}

\paragraph{Asset clustering.} 
P-Tree generates leaves, a.k.a., leaf basis portfolios. Unlike the scalar output in CART for pure prediction, a leaf of P-Tree represents a time series of portfolio returns from a time-invariant P-Tree structure. These leaf basis portfolios are nonlinear, interactive, and high-dimensional characteristics-managed portfolios.
Table \ref{tab:leaves_a}, panel A, summarizes leaf basis portfolios generated by the first P-Tree in Figure \ref{fig:tree_a}: the index of nodes, median number of stocks in the leaf basis portfolio, average returns, CAPM $\alpha$ (\%), $ \beta$, time series regression $R^2$, and alphas with respect to FF5, Q5, RP-PCA, and IPCA five-factor models \citep{fama2015five, hou2021augmented, kelly2019characteristics, lettau2020factors}.

\begin{table}[!h]
    \caption{\bf Evaluation for Leaf Basis Portfolios}
    \label{tab:leaves_a}
    \footnotesize{
        This table reports the performance of 10 leaf basis portfolios generated by the first P-Tree. Panel A corresponds to Figure \ref{fig:tree_a}, whereas Panels B and C correspond respectively to (a) and (b) in Figure \ref{fig:tree_a_T1_T2}.
        We report the median number of stocks (leaf size), average returns, CAPM $\alpha$ (\%), $\beta$, $R^2$, and the alphas (\%) with respect to FF5, Q5, RP-PCA, and IPCA five-factor models. 
        *, **, and *** represent significance levels of 10\%, 5\%, and 1\%, respectively.
    
    \vspace{-0.3cm} 
    \begin{center}

    \resizebox{\textwidth}{!}{
    
        \begin{tabular}{l llllll c llll }
        \toprule

        ID  & \# Median & AVG & STD  & $\alpha_{CAPM}$ & $\beta_{CAPM}$ & $R^2_{CAPM}$ &       & $\alpha_{FF5}$ & $\alpha_{Q5}$ & $\alpha_{RP5}$ & $\alpha_{IP5}$  \\
        \hline
        
        \\
        \multicolumn{12}{c}{\underline{Panel A: 40 Years (1981-2020)}} \\
        \\
        
        N8  & 134   & -0.40  & 5.76 & -0.97*** & 0.84  & 0.42  &       & -0.91*** & -0.66*** & -1.51*** & -0.58* \\
        N9  & 210 & -0.92*** & 4.14  & -1.35*** & 0.63  & 0.47  &       & -1.46*** & -1.25*** & -1.84*** & -0.36* \\
        N20  & 178   & -0.34 & 8.89 & -1.34*** & 1.45  & 0.53  &       & -1.10 & -0.49* & -1.55*** & -1.78*** \\
        N21  & 227   & -1.14*** & 5.90 & -1.83*** & 1.00     & 0.58  &       & -1.83*** & -1.56*** & -2.34*** & -1.19*** \\
        N11  & 900   & 0.36*  & 4.81  & -0.35*** & 1.04  & 0.93  &       & -0.30*** & -0.15** & -0.29*** & 0.16  \\
        N12  & 210   & 0.94*** & 4.63  & 0.42**  & 0.76  & 0.54  &       & 0.29**  & 0.48***  & -0.22 & 1.17***  \\
        N26  & 157   & 1.77*** & 6.04   & 1.16***  & 0.88  & 0.43  &       & 1.08***  & 1.37***  & 0.40**   & 1.40*** \\
        N27  & 112   & 2.46*** & 4.55  & 1.97***  & 0.71  & 0.49  &       & 1.80***  & 2.00*** & 1.18***  & 2.18*** \\
        N14  & 658   & 1.03*** & 6.88   & 0.21  & 1.18  & 0.59  &       & 0.32*  & 0.64***  & -0.33** & 0.06 \\
        N15 & 1768  & 0.81*** & 4.41  & 0.14***  & 0.98  & 0.98  &       & 0.10***  & 0.03  & -0.26*** & 0.29  \\ 
        \hline
        \\
        \multicolumn{12}{c}{\underline{Panel B: 20 Years (1981-2000)}} \\
        \\
                
        N16 & 97    & 0.25  & 5.99  & -0.34 & 0.81  & 0.36  &             & -0.27  & -0.08 & -0.27 & -0.02   \\
        N17 & 256    & -0.74*** & 4.40 & -1.16*** & 0.57  & 0.33  &       & -1.36*** & -1.17*** & -1.56*** & -1.47*** \\
        N18 & 95    & 1.72*** & 5.72  & 1.17***  & 0.75  & 0.34  &        & 1.07***  & 1.27***  & 0.84**  & 0.03  \\
        N19 & 48    & 2.91*** & 4.97 & 2.44***  & 0.64  & 0.33  &         & 2.13***  & 2.50***  & 2.36***  & 1.87*** \\
        N5 & 3443    & 0.76*** & 4.39 & 0.03*  & 0.99  & 1.00     &       & -0.02* & -0.04*** & -0.49*** & 1.34***  \\
        N24 & 160    & -3.53*** & 7.75 & -4.36*** & 1.13  & 0.42  &       & -3.90*** & -3.29*** & -3.58*** & 0.79 \\
        N25 & 250    & -2.82*** & 6.85 & -3.58*** & 1.04  & 0.45  &       & -3.20*** & -2.54*** & -1.89*** & 1.55* \\
        N13 & 286    & -1.75*** & 9.94 & -2.86*** & 1.52  & 0.46  &       & -1.58*** & -1.19*** & -1.29*** & 2.22** \\
        N14 & 217    & -1.21*   & 9.60 & -2.21*** & 1.37  & 0.40  &       & -1.16***     & -0.63 & -0.09 & 4.52**  \\
        N15 & 116   & 1.23*  & 10.07 & 0.16  & 1.46  & 0.41  &            & 1.12***  & 1.52***  & 1.39**  & 3.58  \\
        \hline
        \\
        \multicolumn{12}{c}{\underline{Panel C: 20 Years (2001-2020)}} \\
        \\
        
        N8 & 153    & -0.20 & 4.44  & -0.63** & 0.67  & 0.45  &       & -0.63** & -0.46** & -0.93*** & -0.10  \\
        N18 & 58    & -0.31 & 6.82  & -0.96*** & 1.01  & 0.44  &       & -0.93*** & -0.53* & -1.36*** & -0.88**  \\
        N19 & 112    & -0.90*** & 4.72  & -1.39*** & 0.76  & 0.53  &       & -1.39*** & -1.31*** & -1.89*** & -0.93***  \\
        N5 & 1206    & 0.35 & 5.23  & -0.38*** & 1.13  & 0.95  &       & -0.27*** & -0.25*** & -0.21*** & 0.37*   \\
        N24 & 171    & 0.94** & 6.47 & 0.26  & 1.05  & 0.53  &       & 0.31  & 0.57**  & -0.13 & 0.59**   \\
        N25 & 104    & 2.44*** & 6.80 & 1.79***  & 1.02  & 0.45  &       & 1.88***  & 2.17***  & 1.16***  & 1.43***   \\
        N52 & 47    & 1.68*** & 5.28 & 1.17***  & 0.79  & 0.46  &       & 1.17***  & 1.45***  & 0.52**  & 1.56***   \\
        N53 & 76    & 3.53*** & 5.27  & 2.97***  & 0.86  & 0.54  &       & 2.88***  & 3.10***   & 2.24***  & 3.19*** \\
        N27 & 76    & 1.79*** & 4.05 & 1.36***  & 0.66  & 0.53  &       & 1.31***   & 1.37***  & 0.92***  & 1.6*** \\
        N7 & 1896   & 0.76*** & 4.27 & 0.16***  & 0.94  & 0.98  &       & 0.10**  & 0.09**  & -0.05 & 0.63***  \\

        \bottomrule
        \end{tabular}%
        
        } 

    \end{center}
    
    }

    \vspace{-0.3cm} 

\end{table}

We observe two large leaves: N11 containing 900 stocks and N15 containing over 1,700 stocks by monthly median. Their CAPM $R^2$ are over 90\%, and the $\beta$ are close to one, meaning their return time series are highly correlated with the market returns.
No further splitting results in higher Sharpe ratios on these two leaves, yet other leaves offer higher investment improvement under the MVE framework.
For instance, N21 represents the low-\texttt{SUE}, high-\texttt{DOLVOL}, low-\texttt{ME}, and high-\texttt{ZEROTRADE} leaf, which has -1.14\% average return and -1.83\% CAPM $\alpha$, and N27 is the high-\texttt{SUE}, low-\texttt{DOLVOL}, high-\texttt{BM\_IA}, and high-\texttt{ROA} leaf, which has 2.46\% average returns and 1.97\% CAPM $\alpha$.
By employing a simple monthly rebalanced strategy of buying N27 and selling short N21, one can expect an average return of 3.6\%.

These leaf basis portfolios are generated via interactive and nonlinear splits on 61 characteristics. We expect them to be very hard to price using the prevalent linear factor models.
Nine leaves have significant CAPM alphas among the 10 leaf basis portfolios. 
At a confidence level of 10\%, the number of statistically significant non-zero alphas for FF5, Q5, RP-PCA 5, and IPCA 5 factor models are 9, 9, 9, and 7, respectively.

P-Tree is a goal-oriented clustering algorithm that maximizes the Sharpe ratio in the baseline specification. Therefore, we anticipate these test assets having economically large and statistically significant alphas against those ``inadequate" benchmark factor models.
With such distinct basis portfolios, we expect them to achieve higher mean-variance efficiency, spanning an efficient frontier that benchmark factor models fail to span. 
Panels B and C of Table \ref{tab:test_asset_a} provide subsample analysis for the first and the latter 20 years. Although the tree structure and leaf identity change because of refitting, we still find similar patterns. For example, pricing these portfolios based on leaf factors remains difficult using benchmark factor models.

\begin{figure}[h!]
    \caption{\bf Diversified P-Tree Test Assets
    }
    \label{fig:leaf-basis-portfolio-scatter-plot}
    
    {
    \footnotesize
    This figure reports the performance of P-Tree test assets (presented in Table \ref{tab:leaves_a}) and the $5 \times 5$ ME-BM portfolios. The 10 black circles represent P-Tree test assets, and the 25 light-red triangles represent ME-BM portfolios. Panel A shows the scatter plots of the mean and standard deviation for portfolio returns in percentage. Panel B shows the scatter plots of CAPM alpha and beta, with alphas in percentage.
    }
    
    \vspace{-0.2cm}
    
	\begin{center}
 
    { \small Panel A: P-Tree Test Asset: Mean v.s. Standard Deviation }
    \\
		\begin{subfigure}[b]{.32\textwidth}
			
            \centering
			\includegraphics[width=\textwidth]{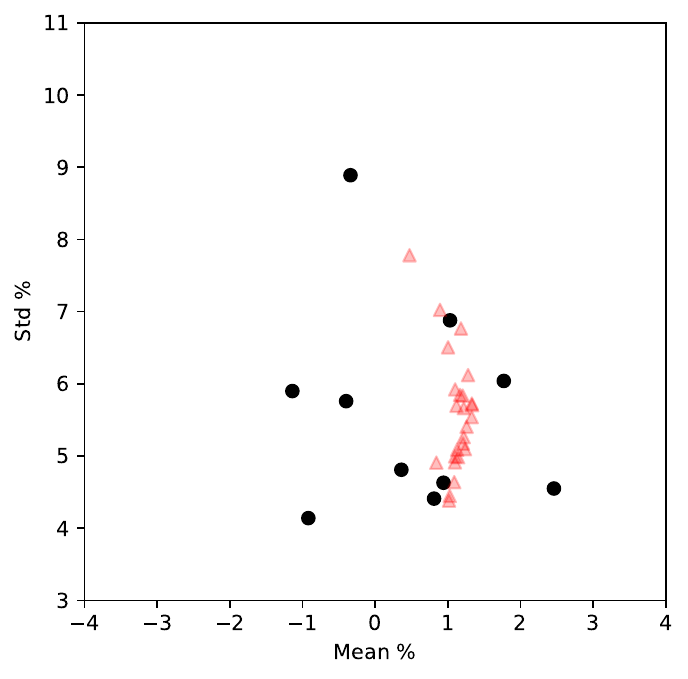}
			\caption*{(A1) 1981-2020}
		\end{subfigure}
		\begin{subfigure}[b]{.32\textwidth}
			
			\centering
			\includegraphics[width=\textwidth]{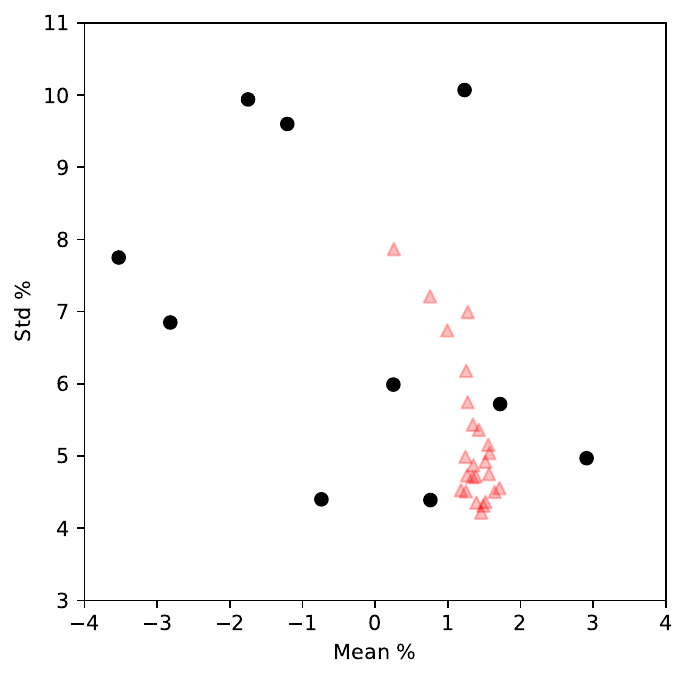}
			\caption*{(A2) 1981-2000}
		\end{subfigure}
		\begin{subfigure}[b]{.32\textwidth}
  
			\centering
			\includegraphics[width=\textwidth]{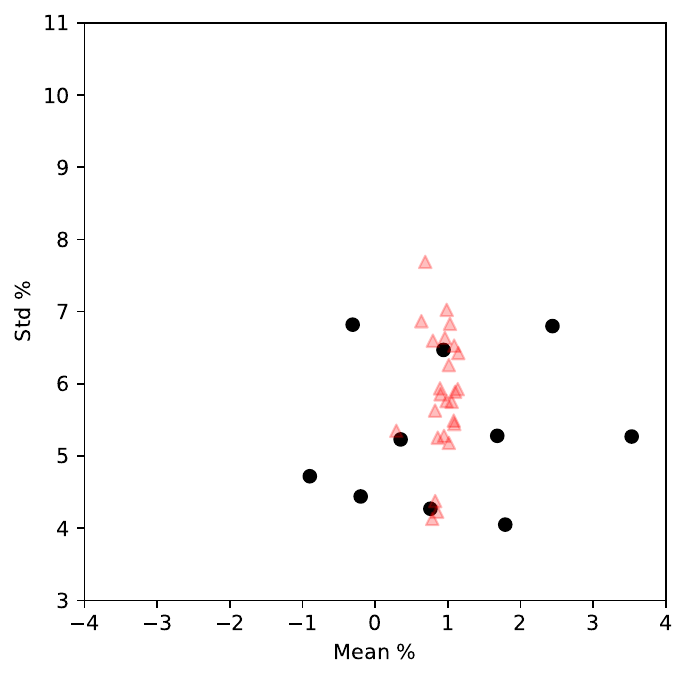}
			\caption*{(A3) 2001-2020}
		\end{subfigure}
        \\
    { \small Panel B: P-Tree Test Asset: CAPM Alpha-Beta }
    \\
		\begin{subfigure}[b]{.32\textwidth}

            \centering
			\includegraphics[width=\textwidth]{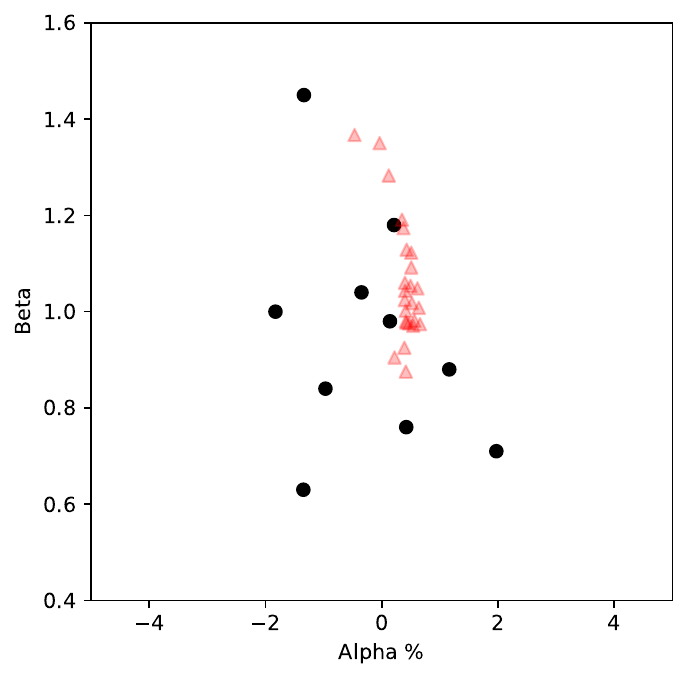}
			\caption*{(B1) 1981-2020}
		\end{subfigure}
		\begin{subfigure}[b]{.32\textwidth}
  
			\centering
			\includegraphics[width=\textwidth]{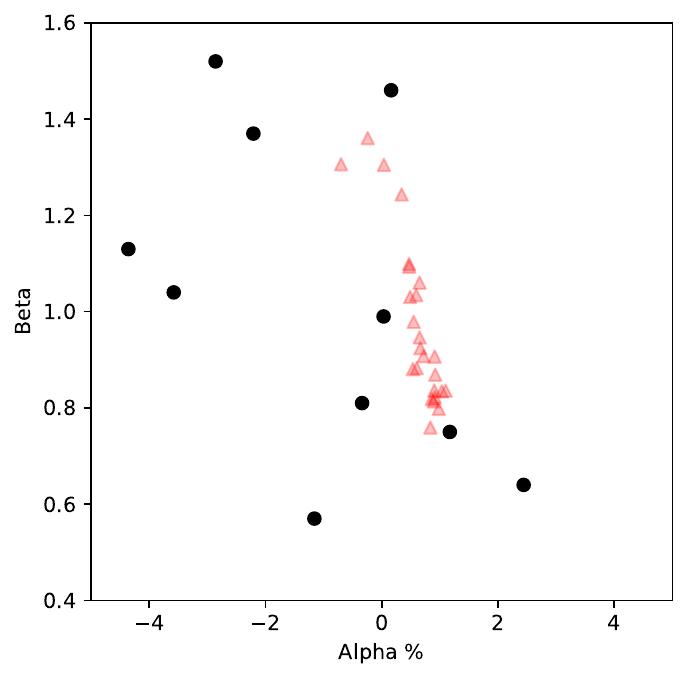}
			\caption*{(B2) 1981-2000}
		\end{subfigure}
		\begin{subfigure}[b]{.32\textwidth}
  
			\centering
			\includegraphics[width=\textwidth]{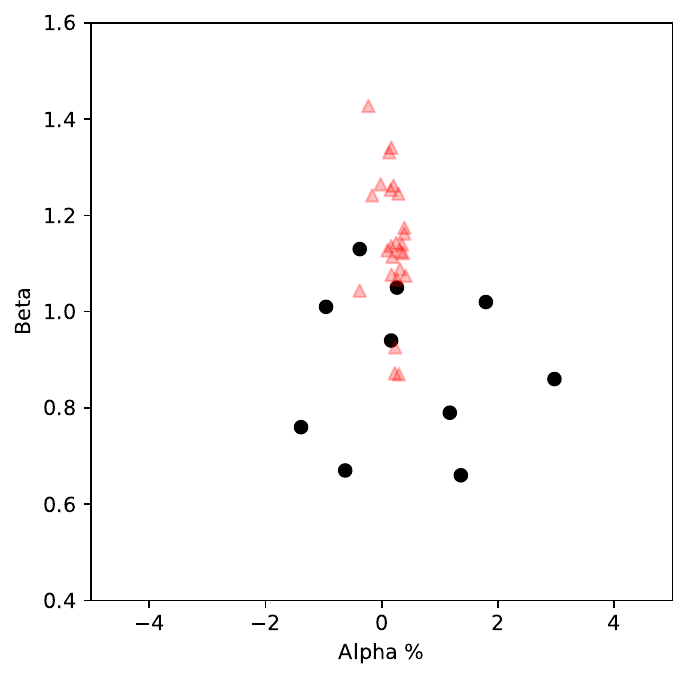}
			\caption*{(B3) 2001-2020}
		\end{subfigure}
    \\
	\end{center}
\vspace{-0.4cm} 
\end{figure}

\paragraph{Diversified test assets.}
Figure \ref{fig:leaf-basis-portfolio-scatter-plot}, Panel A, depicts the performance statistics of P-Tree test assets. Subfigure A1 shows the mean-standard-deviation scatter plot for leaf basis portfolios of the first P-Tree on the 40-year sample from 1981 to 2020 in black circles. The expected returns are in the range of -2\% to 3\%, and the standard deviations are in the range of 4\% to 9\%. We observe large variations in these portfolios' expected returns and risks, which means they are diversified under the MVE framework. 
Significant variations exist among the portfolios in Panels A2 and A3 for subsample analysis, with the earlier sample showing more diversification and the latter sample being more concentrated. By contrast, the $5\times5$ ME-BM portfolios in light-red triangles cluster around the same level of average returns, which are much less diversified than those of P-Tree test assets. This finding indicates that P-Tree test assets are more diversified under the MVE framework than the ME-BM portfolios.

Panel B of Figure \ref{fig:leaf-basis-portfolio-scatter-plot} shows the CAPM model-implied alpha-beta plots. In subfigure B1's 40-year sample, we find CAPM $\beta$s of P-Tree test assets are scattered around one in the range of 0.6 to 1.6. CAPM alphas are around zero, ranging from -2\% to 2\%. These basis portfolios have a lot of variations in CAPM alphas and $\beta$s. The alpha and beta ranges were larger in the first 20 years in B1, whereas the variation was smaller in the second 20 years in B2. 
The $5\times5$ ME-BM portfolios are more easily explained for the CAPM model than P-Tree test assets because they align closely with the vertical line of zero alpha with more negligible diversification in all figures. 
In summary, this is highly positive evidence for P-Tree test assets' better-diversified patterns than conventional ones, even when priced by CAPM.

\section{Boosted P-Trees for Multiple Factors and Test Assets} \label{sec:empirical_boost}

\subsection{Growing the Efficient Frontier} \label{sec:frontier}

Our investigation focuses on the efficient frontiers spanned by test assets generated using (boosted) P-Tree.
Many P-Tree test assets have significant alphas over multiple well-known benchmark factor models, indicating those factor models cannot effectively span the efficient frontier.
The boosting technique enhances our ability to generate more effective test assets that expand the efficient frontier better, that is, ``growing the efficient frontiers on P-Trees."

\begin{figure}[h!]
    \caption{\bf Characterizing the Efficient Frontier with P-Trees
    }
    \label{fig:ef_a}

    {\footnotesize
    This figure shows the MVE frontiers of the first P-Tree, the sequence of boosted P-Tree factor models, benchmark factor models, and benchmark test assets. 
    The dots on the frontiers represent the tangency portfolios.
    The mean and standard deviation are on a monthly scale. 
    The sample period is from 1981 to 2020.
    }
\vspace{-0.2cm}    
    \begin{center}
        \includegraphics[width=0.95\textwidth]{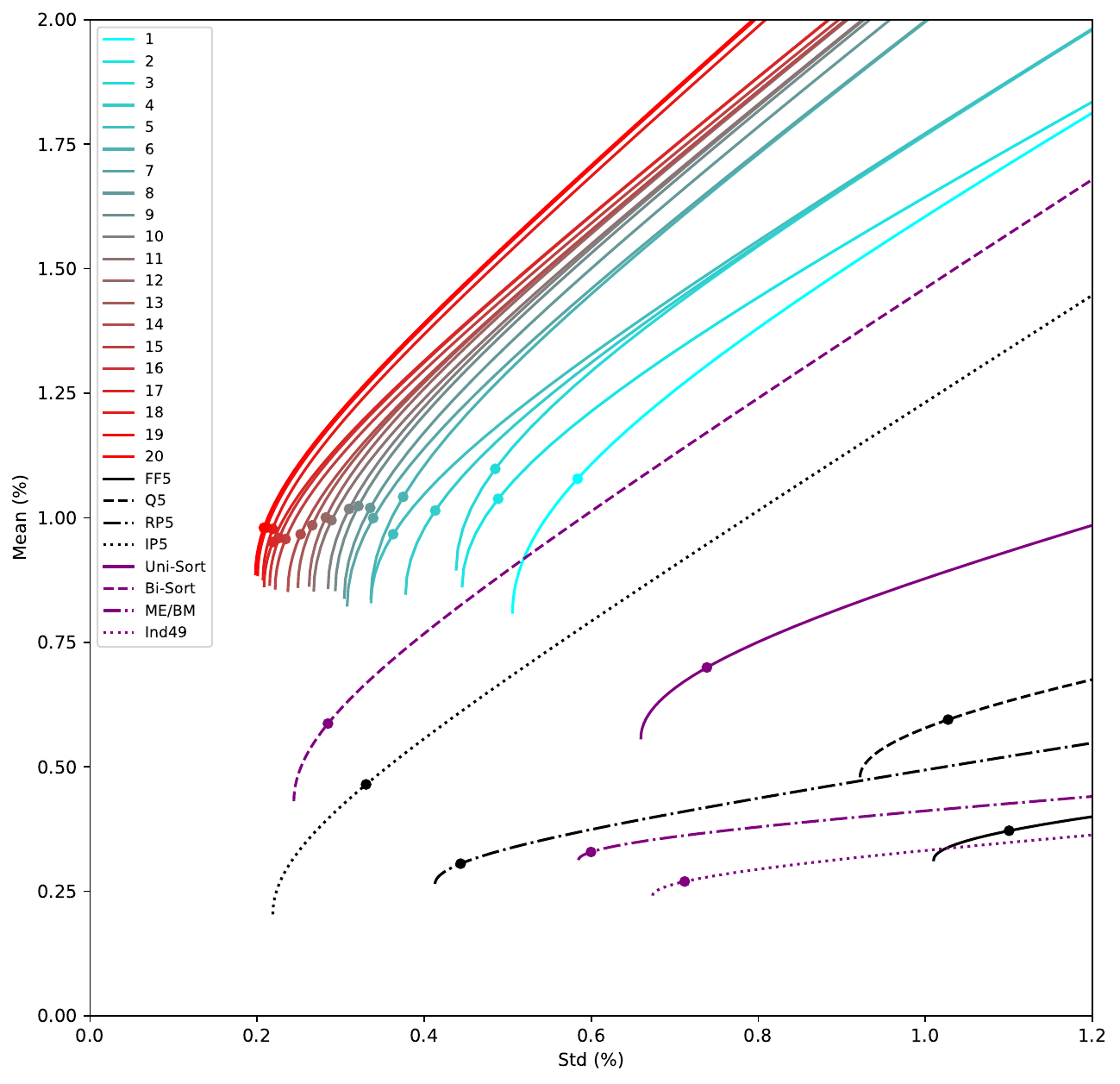}
    \end{center}
\vspace{-0.6cm} 
\end{figure}

\paragraph{Unspanned efficient frontier.}

Figure \ref{fig:ef_a} shows the MVE frontiers of P-Trees and boosted P-Trees in gradient color from blue to red for the sample period from 1981 to 2020. Benchmark factor models are printed in black, and benchmark test assets in purple. Notably, the P-Tree1 frontier (generated by the first P-Tree test assets) is already more efficient than any frontier spanned by the benchmark factor models and test assets. Among the benchmark test assets, 285 bivariate-sorted portfolios span a frontier very close to our P-Tree1 frontier. However, P-Tree1 contains only 10 portfolios, much smaller than 285. Among the factor models, the IPCA five-factor model is a strong candidate but still less efficient than the P-Tree1 frontier. Overall, these 10 leaf basis portfolios consist of promising test assets to span the efficient frontier, which is more efficient than the benchmarks under the MVE framework. In other words, the listed benchmark factor models and test assets are insufficient to span the real efficient frontier.

\paragraph{Boosting the empirical frontier.}
Beyond the first P-Tree, boosted P-Trees improve portfolio efficiency from one factor to 20 factors. In Figure \ref{fig:ef_a}, the spanned frontiers move toward the top left corner of the mean-variance diagram. Our approach consistently advances the efficient frontier throughout the 20 P-Trees. Combining 200 assets in one MVE portfolio is challenging because of the curse of dimensionality. 
Unlike \cite{ait2017using}, who use exogenous industry classifications to construct block-diagonal patterns, our boosted P-Trees entertain an endogenous block-diagonal structure in the return covariance. 
We combine the 10 leaf basis portfolios of each P-Tree into a single P-Tree factor and then estimate the tangency portfolio of the boosted P-Tree factors.
As such, we can generate multiple factors and combine multiple sets of test assets in one frontier under the MVE framework.

Moving from a single P-Tree to 20 P-Trees results in more efficient frontiers, allowing investors to achieve higher expected returns while bearing the same level of risk. 
Although a significant difference exists between the frontiers of P-Tree1 and P-Tree10, the frontier lines increase slowly after P-Tree10, indicating a more negligible improvement in mean-variance efficiency with each subsequent boosting.

We employ asset pricing tests to demonstrate the incremental efficiency of boosted P-Trees. 
First, we show the Sharpe ratios of each boosted P-Tree factor and the multi-factor model. 
Second, we evaluate each boosted factor with CAPM and FF5 by checking for a significant alpha. 
As we know from Table \ref{tab:leaves_a}, P-Tree test assets are hard to price, and we expect the P-Tree factors to pass the CAPM test and FF5 test. 
Third, we regress each P-Tree factor on all the previous boosted P-Tree factors generated for the expanding factor tests. 
The efficiency of the frontier increases if the previous P-Tree factors do not span the additional P-Tree factor.

\begin{table}[h!]
  
  \caption{\bf Testing the Boosted P-Tree Growth}
  \label{tab:factor_dim_a}%
  
  {\footnotesize
  This table shows the performance of each sequentially generated boosted P-Tree factor, including the Sharpe ratio, CAPM test, FF5 test, expanding factors test, and the \cite{barillas2017alpha} test.
  The "Single" column displays the Sharpe ratio of the single factor, and the "Cumu." (cumulative) column shows the Sharpe ratio of the MVE spanned by multiple factors from the first to the current P-Tree factors.
  The columns report the $\alpha$ (\%) and $t$-statistic for CAPM and FF5 spanning regression tests. For the expanding factor tests, we regress each P-Tree factor on all previously boosted P-Tree factors and report the $\alpha$ (\%), $t$-statistic, and $R^2$.
  Table \ref{tab:app_factor_dim_a} shows the results for subsample analysis.
\vspace{-0.3cm} 
  \begin{center}
    \resizebox{\textwidth}{!}{
      \begin{tabular}{l ccccccccccccccc}
      
      \toprule
        
        &
        & \multicolumn{2}{c}{Sharpe Ratio} &
        & \multicolumn{2}{c}{CAPM Test} &
        & \multicolumn{2}{c}{FF5 Test}   &
        & \multicolumn{3}{c}{Expanding Factors Test}   &
        & \multicolumn{1}{c}{BS Test} \\
        
        \cline{3-4} \cline{6-7} \cline{9-10} \cline{12-14} \cline{16-16}

        &
        & Single    & Cumu. &
        & $\alpha$ (\%) & $t$-stat & 
        & $\alpha$ (\%) & $t$-stat & 
        & $\alpha$ (\%) & $t$-stat & $R^2$ & 
        & $p$-value \\
        \hline
        \\

        1     &       & 6.37  & 6.37  &       & 1.39  & 35.36 &       & 1.37  & 35.81 &       & -     & -     & -     &       &  \\
        2     &       & 3.20  & 7.35  &       & 0.52  & 17.65 &       & 0.48  & 15.27 &       & 0.62  & 9.55  & 0.01  &       & 0.00 \\
        3     &       & 1.18  & 7.80  &       & 0.34  & 5.25  &       & 0.18  & 3.28  &       & -0.86 & -5.43 & 0.26  &       & 0.00 \\
        4     &       & 2.06  & 8.46  &       & 0.44  & 11.22 &       & 0.38  & 9.56  &       & 0.69  & 7.63  & 0.16  &       & 0.00 \\
        5     &       & 1.99  & 9.18  &       & 0.48  & 10.25 &       & 0.41  & 9.25  &       & 0.84  & 5.61  & 0.21  &       & 0.00 \\
        6     &       & 1.01  & 9.57  &       & 0.18  & 4.24  &       & 0.08  & 2.94  &       & -0.50 & -5.47 & 0.45  &       & 0.00 \\
        7     &       & 1.42  & 10.11 &       & 0.28  & 7.62  &       & 0.22  & 6.67  &       & 0.63  & 7.03  & 0.36  &       & 0.00 \\
        8     &       & 1.32  & 10.40 &       & 0.28  & 7.14  &       & 0.20  & 4.95  &       & -0.50 & -5.33 & 0.41  &       & 0.00 \\
        9     &       & 1.83  & 10.88 &       & 0.53  & 10.11 &       & 0.43  & 9.38  &       & 0.85  & 6.88  & 0.34  &       & 0.00 \\
        10    &       & 1.48  & 11.20 &       & 0.44  & 7.35  &       & 0.30  & 7.30  &       & -0.68 & -5.49 & 0.46  &       & 0.00 \\
        11    &       & 1.78  & 11.72 &       & 0.38  & 10.33 &       & 0.31  & 8.62  &       & 0.72  & 6.70  & 0.30  &       & 0.00 \\
        12    &       & 1.02  & 12.06 &       & 0.20  & 4.68  &       & 0.10  & 2.99  &       & -0.55 & -5.76 & 0.55  &       & 0.00 \\
        13    &       & 1.37  & 12.57 &       & 0.29  & 8.38  &       & 0.22  & 6.48  &       & 0.76  & 6.01  & 0.33  &       & 0.00 \\
        14    &       & 1.37  & 13.01 &       & 0.48  & 5.93  &       & 0.32  & 5.86  &       & -0.91 & -5.68 & 0.60  &       & 0.00 \\
        15    &       & 1.37  & 13.81 &       & 0.31  & 6.58  &       & 0.21  & 5.66  &       & 0.97  & 7.11  & 0.48  &       & 0.00 \\
        16    &       & 1.24  & 14.28 &       & 0.28  & 6.23  &       & 0.17  & 4.11  &       & -0.74 & -6.24 & 0.54  &       & 0.00 \\
        17    &       & 1.54  & 14.60 &       & 0.46  & 8.16  &       & 0.40  & 7.52  &       & -0.89 & -4.80 & 0.40  &       & 0.00 \\
        18    &       & 1.64  & 14.92 &       & 0.32  & 8.43  &       & 0.27  & 7.78  &       & -0.65 & -5.43 & 0.34  &       & 0.00 \\
        19    &       & 1.48  & 15.43 &       & 0.43  & 8.63  &       & 0.36  & 7.47  &       & 1.18  & 5.86  & 0.34  &       & 0.00 \\
        20    &       & 1.35  & 15.63 &       & 0.33  & 7.38  &       & 0.23  & 6.19  &       & -0.59 & -4.19 & 0.44  &       & 0.00 \\
        
      \bottomrule
      
      \end{tabular}%
    
    } 

    \end{center}

    } 
    
  \vspace{-0.3cm} 
\end{table}%

The test results are presented in Table \ref{tab:factor_dim_a}. The Sharpe ratios of each boosted factor are above 1, even for the 20-th factor. Meanwhile, the cumulative multi-factor Sharpe ratio increases monotonically from 6 to 15, which means that the test assets of the boosted P-Tree add incremental pricing information to the existing test assets. Furthermore, all CAPM and FF5 alphas are positive and highly significant, indicating that all the boosted P-Tree factors cannot be explained by CAPM or FF5.
In the expanding factor test, we find significant alphas, large $t$-statistics, and high time series $R^2$.
After applying the nested asset pricing test \citep{barillas2017alpha}, we do not find any evidence that supports a sufficient factor model to explain the assets in the boosted P-Tree test.\footnote{The first row contains "-" signs, which indicate that the expanding factor and BS tests are not applicable for the single factor case.} 
In summary, Table \ref{tab:factor_dim_a} shows that boosted P-Trees generate unexplained alphas, and their leaf portfolios thus better span the efficient frontier. By contrast, conventional univariate- or bivariate-sorted test assets used for testing asset pricing models do not fully cover the efficient frontier.

\subsection{Generating Test Assets via Boosted P-Trees} \label{sec:p_tree_test_asset}

First, we show the pricing performance of leaf basis portfolios, implying they serve as diversified test assets and are challenging against the FF5 model.
Then, we utilize boosted P-Trees that generate multiple sets of basis portfolios. Empirically, we grow up to 20 P-Trees, yielding 200 leaf basis portfolios.

\paragraph{P-Tree test assets.}
Table \ref{tab:test_asset_a}, Panel A, lists the number of portfolios, the GRS test statistic and its $p$-value, the $p$-value of PY test,\footnote{\cite{pesaran2023testing} adapts to cases where the number of assets being tested is larger than the number of periods, which \cite{gibbons1989test} cannot address.} the average absolute $\alpha$, the root mean squared $\alpha$, the average time series regression $R^2$, and the percentage of significant alphas with respect to 10\%, 5\%, and 1\% confidence levels.
In the rows, P-Tree1 represents a set of test assets comprising 10 leaf basis portfolios in the first P-Tree. P-Tree1-5 combines all the leaf basis portfolios generated by the first five P-Trees, and P-Tree6-10 contains the 6-th to the 10-th P-Trees. Similarly, we define P-Tree11-15, P-Tree16-20, and P-Tree1-20.

\begin{table}[htbp]
  
  \caption{\bf Comparing Test Assets}
  \label{tab:test_asset_a}%

  {\footnotesize
  
    The table displays performance statistics of P-Tree test assets and others for comparison, including P-Tree1, which consists of 10 portfolios in the first P-Tree.
    P-Tree1-5, P-Tree6-10, P-Tree11-15, P-Tree16-20, and P-Tree1-20 are sets of test assets that combine leaf basis portfolios generated by the first 5, 6-10, 11-15, 16-20, and 1-20 P-Trees, respectively.
    ``Uni-Sort" has 150 univariate-sorted portfolios, ``Bi-Sort" has 285 bivariate-sorted portfolios, ME/BE has 25 portfolios, and ``Ind49" includes 49 industry portfolios. 
    The reported statistics include the number of test assets, GRS test statistics and p-values \citep{gibbons1989test}, $p$-values of PY test \citep{pesaran2023testing}, average absolute $\alpha$ (\%), root mean squared $\alpha$ (\%), average $R^2$ (\%) of regressing the portfolios on Fama-French five factors, and the proportion of test assets with unexplained significant alphas under 10\%, 5\%, and 1\% significance levels. 
    The three panels report on the re-trained models for different sample periods. 
    
    \begin{center}
    
    \begin{tabular}{l cccc cccccc}

    \toprule
    
    & N & \multicolumn{1}{l}{GRS} & \multicolumn{1}{l}{$p$-GRS} 
    & \multicolumn{1}{l}{$p$-PY}
    & \multicolumn{1}{l}{ $\overline{|\alpha|}$} 
    & \multicolumn{1}{l}{ $\sqrt{\overline{\alpha^2}}$}
    & \multicolumn{1}{l}{$\overline{R^2}$}
    & $\%\alpha_{10\%}$ & $\%\alpha_{5\%}$ & $\%\alpha_{1\%}$ \\

    \hline
    
    \\
    \multicolumn{11}{c}{\underline{Panel A: 40 Years (1981-2020)}} \\
    \\
              
    P-Tree1 & 10    & 141.27 & 0.00  & 0.00  & 0.92  & 1.11  & 75    & 100   & 90    & 80 \\
    P-Tree1-5 & 50    & 60.32 & 0.00  & 0.00  & 0.44  & 0.61  & 80    & 70    & 62    & 44 \\
    P-Tree6-10 & 50    & 4.60  & 0.00  & 0.00  & 0.29  & 0.37  & 79    & 56    & 50    & 34 \\
    P-Tree11-15 & 50    & 4.74  & 0.00  & 0.00  & 0.20  & 0.26  & 80    & 38    & 36    & 24 \\
    P-Tree16-20 & 50    & 4.21  & 0.00  & 0.00  & 0.31  & 0.42  & 77    & 52    & 44    & 30 \\
    P-Tree1-20 & 200   & 41.31 & 0.00  & 0.00  & 0.31  & 0.43  & 79    & 54    & 48    & 33 \\
    \\
    Uni-Sort & 150   & 1.62  & 0.00  & 0.00  & 0.10  & 0.14  & 88    & 25    & 18    & 7 \\
    Bi-Sort & 285   & 2.50  & 0.00  & 0.00  & 0.12  & 0.17  & 89    & 30    & 23    & 15 \\
    ME-BM & 25    & 5.01  & 0.00  & 0.00  & 0.12  & 0.16  & 92    & 36    & 28    & 20 \\
    Ind49 & 49    & 1.99  & 0.00  & 0.00  & 0.28  & 0.35  & 60    & 39    & 31    & 18 \\
    
    \hline
    
    \\
    \multicolumn{11}{c}{\underline{Panel B: 20 Years (1981-2000)}} \\
    \\

    P-Tree1 & 10    & 84.36 & 0.00  & 0.00  & 1.58  & 1.95  & 70    & 90    & 80    & 80 \\
    P-Tree1-5 & 50    & 50.84 & 0.00  & 0.00  & 0.79  & 1.26  & 76    & 62    & 60    & 52 \\
    P-Tree6-10 & 50    & 7.27  & 0.00  & 0.00  & 0.58  & 0.87  & 75    & 56    & 44    & 38 \\
    P-Tree11-15 & 50    & 6.39  & 0.00  & 0.00  & 0.55  & 0.82  & 76    & 66    & 60    & 42 \\
    P-Tree16-20 & 50    & 8.42  & 0.00  & 0.00  & 0.52  & 0.74  & 76    & 62    & 54    & 50 \\
    P-Tree1-20 & 200   & 112.90 & 0.00  & 0.00  & 0.61  & 0.95  & 76    & 62    & 55    & 46 \\
    \\
    Uni-Sort & 150   & 1.94  & 0.00  & 0.00  & 0.17  & 0.23  & 88    & 35    & 27    & 19 \\
    Bi-Sort & 285   & - & - & 0.00  & 0.21  & 0.30  & 89    & 42    & 33    & 21 \\
    ME-BM & 25    & 4.75  & 0.00  & 0.00  & 0.21  & 0.25  & 91    & 56    & 48    & 32 \\
    Ind49 & 49    & 2.44  & 0.00  & 0.00  & 0.52  & 0.61  & 61    & 59    & 49    & 27 \\

    \hline
    
    \\
    \multicolumn{11}{c}{\underline{Panel C: 20 Years (2001-2020)}} \\
    \\

    P-Tree1 & 10    & 56.76 & 0.00  & 0.00  & 1.09  & 1.35  & 68    & 90    & 90    & 90 \\
    P-Tree1-5 & 50    & 30.35 & 0.00  & 0.00  & 0.43  & 0.68  & 76    & 52    & 38    & 24 \\
    P-Tree6-10 & 50    & 5.17  & 0.00  & 0.00  & 0.29  & 0.37  & 75    & 34    & 28    & 14 \\
    P-Tree11-15 & 50    & 2.20  & 0.00  & 0.00  & 0.27  & 0.35  & 75    & 30    & 22    & 10 \\
    P-Tree16-20 & 50    & 2.52  & 0.00  & 0.00  & 0.31  & 0.40  & 76    & 42    & 28    & 10 \\
    P-Tree1-20 & 200   & 83.91 & 0.00  & 0.00  & 0.33  & 0.47  & 76    & 40    & 29    & 14 \\
    \\
    Uni-Sort & 150   & 1.46  & 0.03  & 0.94  & 0.09  & 0.12  & 89    & 12    & 7     & 0 \\
    Bi-Sort & 285   & - & - & 0.01  & 0.11  & 0.15  & 91    & 21    & 15    & 6 \\
    ME-BM & 25    & 2.58  & 0.00  & 0.10  & 0.11  & 0.14  & 93    & 24    & 8     & 8 \\
    Ind49 & 49    & 1.29  & 0.11  & 0.36  & 0.25  & 0.32  & 62    & 18    & 8     & 2 \\

    \bottomrule
    
    \end{tabular}%

    \end{center}

    }

\vspace{-0.3cm}  
\end{table}%

We find the GRS test and PY test always reject the null hypothesis that the alphas of test assets are jointly zero for all specifications of P-Tree test assets, which means the expected returns of P-Tree test assets can hardly be explained by the FF5 model.
Furthermore, the last three columns show that many assets have statistically significant alphas.
For comparison, we include four sets of test assets that are commonly used in literature: 150 $10\times 1$ univariate-sorted portfolios, 285 bivariate-sorted portfolios, $5\times5$ ME-BM portfolios, and 49 industry portfolios, all of which can be downloaded from Ken French’s website.
P-Tree test assets have larger GRS test statistics and alphas than these benchmark test assets.

Furthermore, we observe the decline of alphas from the first P-Tree to the subsequently boosted P-Trees as anticipated, indicating diminishing marginal convergence to the limit of the efficient frontier. 
The test assets of the first P-Tree have larger GRS statistics, higher average alphas, and higher proportions of significant alphas than the test assets generated by the follow-up P-Trees.
In addition, we see an evident decline in GRS statistics and alphas from P-Tree1-5 to P-Tree6-10, P-Tree11-15, and P-Tree16-20. P-Tree generates the most informative test assets in the first P-Tree, whereas the incremental information in the boosted P-Trees is complementary and declining.
Although the alphas decline along boosting, the test assets generated by P-Tree16-20 cannot be explained by FF5, with a 4.21 GRS test statistic, 0.31\% mean absolute alpha, and 52\% test assets having significant alphas at the 10\% level. The test assets of P-Tree16-20 are still harder to price than univariate- and bivariate-sorted portfolios.

Overall, these 200 test assets created by 20 P-Trees on high-dimensional characteristics challenge the FF5 model. They are more difficult to price than benchmark test assets, setting a high standard for testing factor models, which respond to the concerns in empirical studies \citep{lewellen2010skeptical, Ang2020JFQA}.

For robustness, Table \ref{tab:test_asset_a}, Panels B and C, report the first and the latter 20-year subsamples. 
In the recent 20 years, FF5 has performed well in pricing the benchmark test asset; that is, the GRS and PY tests cannot be rejected at the 10\% level, and less significant alphas testing on 49 industry portfolios and univariate-sorted portfolios. 
However, the P-Tree test assets always reject the null hypothesis for GRS and PY tests and are consistently challenging to price in both subsamples.\footnote{The GRS test is not applicable in the presence of more test assets than observations, as is the case for Bi-Sort in Panels B and C, marked with ``-".}

\paragraph{Advantages of P-Tree clustering.} 
P-Tree has two advantages that make these test assets efficient. 
First, P-Tree is a goal-oriented clustering algorithm tailored for spanning the efficient frontier iteratively. Each leaf clusters individual asset returns based on similar characteristic values, aggregating as a time series of portfolio returns. The goal is to maximize the collective Sharpe ratio of the MVE portfolio spanned by the leaf basis portfolios. 
The economic objective of P-Tree is vastly different from the CART and other off-the-shelf ML models, which focus on statistical return prediction without economic utilities and cannot ever tackle the task of growing an efficient frontier.

Second, P-Tree exploits the complex asymmetric interactions and nonlinearities on high-dimensional characteristics. 
The commonly used test assets are characteristics-sorted portfolios, such as univariate- and bivariate-sorted portfolios, which only consider decile or quintile sorting, along with up to three characteristics chosen ad hoc or based on researchers' domain expertise. 
In contrast, P-Tree is more flexible, allowing for general sequential and simultaneous sorting with more characteristics and asymmetric interactions in a unified framework. 
P-Trees enable the inclusion of large leaves, like N15, as shown in Figure \ref{fig:tree_a}, where further splitting does not result in a higher Sharpe ratio, as well as small leaves, such as N8. 
The asymmetric splitting allows P-Trees to focus on the finer clusters of informative assets while paying less attention to non-informative assets under the MVE framework.

\subsection{P-Tree Factors and Asset Pricing Performance}\label{sec:ap-performance}

Using the findings from the previous analysis, we have established that P-Tree test assets are distinct from conventional test assets. Moving forward, we proceed with cross-examining P-Tree factors and various benchmark factor models on different sets of test assets, including P-Tree leaf basis portfolios.
We adopt the cross-sectional $R^2$ to evaluate the asset pricing performance:
\vspace{-0.3cm}
\begin{equation}\label{eqn:csR2}
	      	\text{Cross-sectional }R^2 = 1 - \frac{\sum_{i=1}^N\left( \bar R_{i} - \widehat{\bar R_{i}}\right)^2}{\sum_{i=1}^N \left(\bar R_{i} \right)^2},	
\end{equation}
where $\bar R_i$ denotes the average return and $\widehat{\bar R_{i}} = \widehat{\boldsymbol{\beta}_i}^\prime \widetilde{\boldsymbol{\lambda}}$ is the factor model implied average return of asset $i$. The risk premium estimation adopts the cross-sectional regression estimates of factors, and factor loadings are estimated from time series regressions.
The cross-sectional $R^2$ represents the faction of assets' average returns explained by the model-implied expected returns. 
Various test assets, including P-Tree test assets, conventional sorted portfolios, and industry portfolios, are used to test the factor models. 
In Table \ref{tab:csr2_a}, we display the factor models in the columns to explain the expected return of test assets in the rows.
Panel A provides the asset pricing performances for the 40-year sample, and Panels B and C report for the 20-year subsamples.\footnote{The first row contains two ``--,", which indicate the number of assets $N=10$ is equal to or smaller than the number of factors (10 or 20). Thus, the cross-sectional regression is not applicable.}

\begin{table}[!h]
    \caption{\bf  Asset Pricing Performance: Cross-Sectional $R^2$}
    \label{tab:csr2_a}

    {\footnotesize
    This table displays the cross-sectional $R^2$ (\%) of pricing different sets of test assets by the P-Tree factor models and benchmark factor models. 
    The factor models are listed in the columns, and the different specifications of test assets are listed in the rows.
    Specifically, the table includes 1, 5, 10, and 20 P-Tree factor models and FF5, Q5, RP-PCA, and IPCA five-factor models in the columns. In the rows, the test assets are from the top 1, 5, 10, and 20 P-Trees, univariate-sorted portfolios, bivariate-sorted portfolios, ME-BM $5\times 5$ portfolios, and 49 industry portfolios. 

    \begin{center}

    \begin{tabular}{l l cccc cccc}
    
        \toprule
        
         & & P-Tree1F & P-Tree5F & P-Tree10F & P-Tree20F & FF5    & Q5    & RP5 & IP5 \\
        \hline
        \\

        \multicolumn{10}{c}{\underline{Panel A: 40 Years (1981-2020)}} \\
        \\
        
        P-Tree1 & & 98.3 & 99.2 & -- & -- & 36.8 & 33.9 & 51.1 & 73.5 \\
        P-Tree1-5 & & 65.0  & 82.6  & 90.2  & 95.9  & 54.3  & 55.9  & 56.8  & 49.6 \\
        P-Tree1-10 & & 68.5  & 74.0  & 87.9  & 92.1  & 66.9  & 70.0  & 69.1  & 62.4 \\
        P-Tree1-20 & & 72.1  & 77.9  & 84.3  & 88.9  & 73.8  & 76.8  & 77.8  & 68.6 \\
        \\
        Uni-Sort & & 92.7  & 97.3  & 98.5  & 98.7  & 97.0  & 97.9  & 98.0  & 95.3 \\
        Bi-Sort & & 88.7  & 97.2  & 98.0  & 98.5  & 96.1  & 97.5  & 97.3  & 92.9 \\
        ME-BM & & 88.4  & 98.6  & 98.8  & 99.4  & 96.8  & 97.4  & 97.0  & 96.6 \\
        Ind49 & & 82.0  & 92.9  & 95.3  & 97.8  & 96.1  & 95.9  & 95.7  & 91.1 \\
        \hline
        \\
        \multicolumn{10}{c}{\underline{Panel B: 20 Years (1981-2000)}} \\
        \\

        P-Tree1 & & 99.3 & 99.3 & -- & -- & 54.9 & 59.7 & 37.6 & 79.5 \\
        P-Tree1-5 & & 49.2  & 88.5  & 92.6  & 96.0  & 25.8  & 31.8  & 41.1  & 48.6 \\
        P-Tree1-10 & & 47.7  & 59.7  & 77.8  & 82.3  & 37.7  & 40.5  & 47.8  & 46.8 \\
        P-Tree1-20 & & 42.8  & 54.0  & 70.2  & 74.5  & 42.2  & 45.6  & 52.5  & 50.2 \\
        \\
        Uni-Sort & & 80.4  & 96.5  & 96.9  & 97.6  & 94.6  & 95.9  & 97.0  & 94.5 \\
        Bi-Sort & & 66.4  & 92.1  & 93.9  & 94.9  & 89.1  & 91.5  & 92.7  & 91.1 \\
        ME-BM & & 66.6  & 94.0  & 96.2  & 99.0  & 93.4  & 95.3  & 96.4  & 94.5 \\
        Ind49 & & 60.9  & 86.5  & 87.9  & 95.0  & 91.0  & 89.7  & 92.1  & 84.1 \\
        \hline
        \\
        \multicolumn{10}{c}{\underline{Panel C: 20 Year (2001-2020)}} \\
        \\
        P-Tree1 & & 99.2 & 99.7 & -- & -- & 52.2 & 43.6 & 69.7 & 53.9 \\
        P-Tree1-5 & & 71.4  & 87.6  & 92.3  & 94.3  & 60.0  & 60.7  & 64.9  & 61.3 \\
        P-Tree1-10 & & 70.4  & 78.1  & 86.4  & 88.4  & 66.4  & 65.7  & 69.8  & 65.4 \\
        P-Tree1-20 & & 70.5  & 76.4  & 79.8  & 82.8  & 71.7  & 69.9  & 75.4  & 65.2 \\
        Uni-Sort & & 87.5  & 96.6  & 97.6  & 97.9  & 97.3  & 97.2  & 97.4  & 95.7 \\
        Bi-Sort & & 93.0  & 97.2  & 97.9  & 98.4  & 97.9  & 97.5  & 98.0  & 96.9 \\
        ME-BM & & 92.9  & 95.7  & 98.6  & 99.3  & 97.5  & 97.5  & 97.4  & 96.1 \\
        Ind49 & & 75.3  & 92.6  & 94.8  & 97.7  & 90.2  & 90.8  & 92.8  & 86.0 \\
        
        \bottomrule
        
    \end{tabular}%
    
    \end{center}
    }
\vspace{-0.3cm}    
\end{table}

First, the P-Tree1 factor (from the first P-Tree) explains 98.3\% of the cross-sectional return variation of the P-Tree1 leaf basis portfolios (10 assets), whereas the benchmark factor models provide much lower performance.
This finding is not surprising, since the first P-Tree factor is constructed jointly with these test assets.
However, the FF5, Q5, and RP-PCA five-factor models have much smaller $R^2$s than the P-Tree1 factor, indicating large pricing errors in explaining the expected returns of the highly diversified P-Tree1 test assets. One exception is that the IPCA five-factor model performs well in pricing the P-Tree1 test assets, with a 73.5\% $R^2$.

Second, the P-Tree five-factor model (P-Tree1-5, read as a model containing from the first to the fifth factors), which includes five P-Tree factors from the first five P-Trees, demonstrates tremendous asset pricing performance. 
It outperforms the benchmark five-factor models in pricing the first 10 P-Trees' (P-Tree1-10) test assets and $5\times 5$ ME-BM portfolios. 
For other specifications of test assets, P-Tree1-5 shows comparable levels of cross-sectional $R^2$ among all the five-factor models for a fair comparison. The superior pricing performance remains consistent across Panels A to C for the entire sample and subsample analysis.

Third, these endogenously generated P-Tree test assets better expand the MVE frontier and raise the bar for testing asset pricing models.
The commonly used test assets set a low bar for testing factor models, so benchmark factor models seem sufficient to price them. 
As we see from Panel A of Table \ref{tab:csr2_a}, the listed benchmark five-factor models have over 90\% cross-sectional $R^2$s on the benchmark test assets, and these numbers are consistently large in Panels B and C for subsample analyses. 
However, their $R^2$s decline dramatically when tested against P-Tree test assets. 
P-Tree test assets use nonlinear and asymmetric interactions among a large set of characteristics, whereas benchmark test assets involve only up to two characteristics.\footnote{Industry portfolios are an exception because they are not sorted based on characteristics.}

In summary, P-Tree factor models accurately price P-Tree test assets, while benchmark factors do not, and perform similarly or better than benchmark factor models in pricing benchmark test assets. P-Tree test assets set a high standard for testing asset pricing models, surpassing common sorted and industry portfolios.

\subsection{P-Tree Factors for Investment}

The P-Tree framework clusters thousands of assets into several leaf basis portfolios, reducing portfolio optimization complexity.
These P-Tree test assets can be used as factors directly or as building blocks toward MVE portfolios for investment.
Additionally, the frequency of rebalancing P-Tree investment strategies can be reduced to quarterly or annually to decrease transaction costs.

\begin{table}[h!]
  
  \caption{\bf Factor Investing by Boosted P-Trees}
  \label{tab:factors_a}%

  {\footnotesize
  
  This table presents the investment performance of the initial factors produced by P-Tree and the MVE portfolio of multi-factors generated by boosted P-Trees. The reported statistics include the annualized Sharpe ratio and alphas with respect to CAPM, FF5, Q5, RP-PCA, and IPCA five-factor models.
  Panel A shows the 40-year sample from 1981 to 2020. Panels B1 and B2 present in- and out-of-sample results from 1981 to 2000 and 2001 to 2020, respectively. Panels C1 and C2 display in- and out-of-sample results from 2001 to 2020 and 1981 to 2000, respectively. We find all the $\alpha$'s are significant at 1\% confidence level.

  \begin{center}
    
    \begin{tabular}{l cccccc}
    
    \toprule
          & SR    & $\alpha_{CAPM}$ & $\alpha_{FF5}$ & $\alpha_{Q5}$ & $\alpha_{RP5}$ & $\alpha_{IP5}$ \\
          \hline
            \\
          \multicolumn{7}{c}{\underline{Panel A: 40 Years (1981-2020)}}\\
          \\
                
    P-Tree1    &  6.37 & 1.39  & 1.37  & 1.36  & 1.28  & 1.12 \\
    P-Tree1-5  &  9.19 & 0.97  & 0.95  & 0.93  & 0.87  & 0.82 \\
    P-Tree1-10 & 11.21 & 1.01  & 1.00  & 0.98  & 0.93  & 0.89 \\
    P-Tree1-15 & 13.83 & 0.95  & 0.94  & 0.93  & 0.90  & 0.87 \\
    P-Tree1-20 & 15.64 & 0.97  & 0.96  & 0.95  & 0.93  & 0.90 \\

    \hline
          \\
          \multicolumn{7}{c}{\underline{Panel B1: 20 Years In-Sample (1981-2000)}}\\
          \\
          
    P-Tree1    &  7.13 & 1.86  & 1.78  & 1.72  & 1.62  & 1.59 \\
    P-Tree1-5  & 12.74 & 1.54  & 1.51  & 1.48  & 1.37  & 1.41 \\
    P-Tree1-10 & 19.22 & 1.51  & 1.49  & 1.49  & 1.43  & 1.43 \\
    P-Tree1-15 & 28.43 & 1.42  & 1.41  & 1.40  & 1.37  & 1.39 \\
    P-Tree1-20 & 38.01 & 1.36  & 1.35  & 1.34  & 1.32  & 1.34 \\
    
          \\
          \multicolumn{7}{c}{\underline{Panel B2: 20 Years Out-of-Sample (2001-2020)}}\\
          \\
    
    P-Tree1    & 3.23  & 1.35  & 1.31  & 1.23  & 1.04  & 0.93 \\
    P-Tree1-5  & 3.41  & 1.02  & 1.00  & 0.95  & 0.77  & 0.62 \\
    P-Tree1-10 & 3.21  & 0.95  & 0.94  & 0.89  & 0.74  & 0.56 \\
    P-Tree1-15 & 3.12  & 0.89  & 0.89  & 0.83  & 0.69  & 0.48 \\
    P-Tree1-20 & 3.13  & 0.85  & 0.84  & 0.78  & 0.66  & 0.49 \\

    \hline
          \\
          \multicolumn{7}{c}{\underline{Panel C1: 20 Years In-Sample (2001-2020)}}\\
          \\
                    
    P-Tree1    &  5.83 & 1.51  & 1.47  & 1.50  & 1.52  & 1.69 \\
    P-Tree1-5  &  9.32 & 1.30  & 1.29  & 1.28  & 1.30  & 1.31 \\
    P-Tree1-10 & 14.35 & 1.12  & 1.11  & 1.11  & 1.11  & 1.09 \\
    P-Tree1-15 & 20.64 & 1.08  & 1.07  & 1.08  & 1.10  & 1.05 \\
    P-Tree1-20 & 26.57 & 1.09  & 1.08  & 1.08  & 1.10  & 1.11 \\

          \\
          \multicolumn{7}{c}{\underline{Panel C2: 20 Years Out-of-Sample (1981-2000)}}\\
          \\
    
    P-Tree1    & 4.35  & 1.50  & 1.42  & 1.35  & 1.60  & 1.58 \\
    P-Tree1-5  & 3.87  & 1.18  & 1.05  & 0.96  & 1.23  & 1.24 \\
    P-Tree1-10 & 4.29  & 1.02  & 0.93  & 0.85  & 1.14  & 1.10 \\
    P-Tree1-15 & 4.03  & 0.96  & 0.86  & 0.80  & 1.07  & 1.02 \\
    P-Tree1-20 & 3.88  & 0.96  & 0.87  & 0.81  & 1.08  & 1.03 \\

    \bottomrule

    \end{tabular}
    
    \end{center}
    }
\vspace{-0.3cm} 
\end{table}%

\paragraph{Full-sample evaluation.}
We evaluate the investment performance of P-Tree factors by combining them into one --— the tangency portfolio of P-Tree factors. 
Table \ref{tab:factors_a}, Panel A, reports the P-Tree investment strategy's annualized Sharpe ratios and monthly alphas in the 40-year sample. The Sharpe ratio increases from 6.37 to 9.19, 11.21, 13.83, and 15.64 for the one-factor strategy to 5, 10, 15, and 20 factors, respectively. 
To assess their additional investment information, we further evaluate these P-Tree investment strategies over benchmark factor models. These model-adjusted alphas are all greater than 0.80\% and highly statistically significant.

The existing research finds that P-Tree investment strategies have highly competitive Sharpe ratios.
The correlation-based clustering in \cite{ahn2009basis} generates basis portfolios that underperform bivariate-sorted portfolios on \texttt{ME} and \texttt{BM}.
Furthermore, \cite{daniel2020cross} nearly doubles the Sharpe ratio of the characteristics-sorted portfolio from 1.17 to 2.13 by constructing the corresponding characteristic-efficient portfolios.
However, a single P-Tree delivers a 6.37 Sharpe ratio over 40 years from 1981 to 2020, representing a significant improvement over \cite{ahn2009basis} and \cite{daniel2020cross}, although the empirical samples differ.

\paragraph{Out-of-sample evaluation.}
The results in Panel A of Table \ref{tab:factors_a} are based on a full-sample analysis, which may raise concerns about in-sample overfitting. Therefore, we provide two exercises on OOS investment strategies. We follow \cite{kozak2020shrinking} and perform a half-half split to construct the training and test samples. It suffices to demonstrate the OOS performance when the past and future samples predict each other. This approach further mitigates concerns regarding the look-ahead bias when models trained by the future sample predict the past.

On the one hand, the past-predicting-future results are reported in Panels B1 and B2. The first twenty-year in-sample Sharpe ratios are even higher than the forty-year ones in Panel A, and the OOS Sharpe ratios are over 3.
Despite decreasing performance, many OOS alphas, adjusted by benchmark models, are still close to 1\% and highly significant.
On the other hand, for the future-predicting-past result, the OOS Sharpe ratios are larger than 3.87, and the model-adjusted alphas are over 0.80\%. These findings demonstrate strong OOS evidence for P-Tree test assets and factors for spanning the MVE frontier.

The OOS investment performance results reported here are similar to those in recent studies.
The deep reinforcement learning strategy in \cite{cong2020alphaportfolio} achieves a Sharpe ratio consistently above 2 from 1990 to 2021 even after excluding small stocks.
Based on sorted portfolios on size, operating profitability, and investment in \cite{bryzgalova2023forest}, the regularized portfolio achieves an annualized Sharpe ratio of 2.39 from 2004 to 2016.
The deep neural network approach in \cite{feng2024deep} reports Sharpe ratios ranging from 2.95 to 3.00 from 2002 to 2021.
P-Tree strategies stand out for their transparency, ease of computation, and flexibility as an all-around framework for test asset generation, asset pricing, and investment management.

\section{Model Extensions and Discussion}\label{sec:chars_and_macro}	
We extend the P-Tree framework and applications along several directions, covering interpretability, model complexity, characteristic utility, and macroeconomic regimes.
\subsection{Random P-Forest and P-Tree Interpretability}
\label{sec:random_p_forest}

Random forest \citep{breiman2001random} grows multiple "decorrelated" trees on bootstrap training samples, which are random samples of observations with replacement and a subset of variables from the complete training set. This is called bagging, which helps mitigate model overfitting and quantify the uncertainty of model estimation.
We can adopt this ensemble scheme and grow multiple "decorrelated" P-Trees on bootstrap training samples to form a random P-Forest.
Consequently, there are two direct applications for the random P-Forest: assessing the characteristic importance and recovering a robust SDF using the large number of "decorrelated" P-Trees.

\paragraph{Characteristic importance.}
The simulation evidence in Internet Appendix \ref{sec:simulation} shows how the nonlinear and interactive P-Tree can recover the true set of characteristics, despite redundant or useless characteristics, leading to improved risk proxies for efficient frontier estimation. This contrasts with a linear factor model with ad hoc factors that might disregard these characteristics.
The traditional sequential or independent asset sorting scheme usually focuses on up to three characteristics. Enumerating all possible sorting cases with multiple characteristics is NP-hard, but P-Tree's iterative growth algorithm efficiently overcomes this problem, making computation feasible.

The ensemble steps for constructing the random P-Forest are as follows:
(i) Bootstrap the data on the time-series horizon with replacement and preserve the complete cross section of the panel data for the selected time periods to exploit the low serial correlations among asset returns.
(ii) Randomly select 20 characteristics and independently grow P-Trees on each bootstrap sample. 
These two steps are repeated 1,000 times to create a forest of 1,000 P-Trees.

We study how often a characteristic is chosen for a split in the random P-Forest. A characteristic selected more often is seen as more important for maximizing mean-variance efficiency. In each bootstrap sample, 20 characteristics are randomly drawn to grow the tree, with only some chosen as split variables.
We count the number of times a specific $l$-th characteristic $z_l$ is used in the first $J$ splits and the total number of appearances. We define the measure of characteristic importance as follows:
\vspace{-0.2cm}
\begin{equation}
\label{eqn:select_prob}
    \text { Selection Probability }\left(z_l\right)=\frac{\#\left(z_l \text { is selected at first } J \text { splits }\right)}{\#\left(z_l \text { appears in all bootstrap subsamples }\right)}.
\end{equation}
Table \ref{tab:vi_split} summarizes the selection probabilities \eqref{eqn:select_prob} of characteristics for $J=1,2,3$. The selection probability increases as $J$ grows from 1 to 3, so we must compare within the row. For each row, we list the top five selected characteristics.

\begin{table}[h!]
	\caption{\bf Characteristic Importance by Selection Probability}
	\label{tab:vi_split}
	\footnotesize{
		This table reports the most frequently selected characteristics from the random P-Forest of 1,000 trees.
            The "Top 1" rows only count the first split for 1,000 trees.
            The "Top 2" or "Top 3" rows only count the first two or three splits.
            The numbers reported are the selection frequency for these top characteristics selected out of the 1,000 ensembles.
            The descriptions of characteristics are listed in Table \ref{tab:chars}.
		
		\begin{center}
			      
            \begin{tabular}{l ccccc}
				\toprule

                  & 1     & 2     & 3     & 4     & 5 \\
                  \cline{2-6}
                  \\

                \multirow{2}[0]{*}{Top1} & SUE   & SVAR  & CHPM  & RVAR\_CAPM & BASPREAD \\
                      & 0.51  & 0.34  & 0.25  & 0.23  & 0.21 \\
                      \\
                \multirow{2}[0]{*}{Top2} & SUE   & SVAR  & DOLVOL & CHPM  & BM\_IA \\
                      & 0.66  & 0.38  & 0.34  & 0.31  & 0.31 \\
                      \\
                \multirow{2}[0]{*}{Top3} & SUE   & DOLVOL & BM\_IA & SVAR  & ME\_IA \\
                      & 0.71  & 0.53  & 0.48  & 0.41  & 0.41 \\

                \bottomrule
            \end{tabular}
            
		\end{center}
	}
\vspace{-0.3cm} 
\end{table}

The earnings surprise (\texttt{SUE}), one of the most important fundamentals, has a 50\% chance of being the first splitting characteristic and a 66\% chance of being one of the top two splitting characteristics in the bootstrap sample.
While not incorporated into linear factor models like FF5 and Q5, \texttt{SUE} is most valuable in nonlinear interactive modeling for maximizing the mean-variance efficiency.
Other frequently selected characteristics are return volatility (\texttt{SVAR}), dollar trading volume (\texttt{DOLVOL}), idiosyncratic volatility (\texttt{RVAR\_CAPM}), change in profit margin (\texttt{CHPM}), and industry-adjusted book-to-market ratio (\texttt{BM\_IA}). 
The characteristics in Table \ref{tab:vi_split} cover four major categories---momentum, frictions, profitability, value --- in the top splits.
Note that characteristics not selected do not necessarily mean they are useless, as this table only shows the top three splits.

\paragraph{Interpretability V.S. overfitting.} 
The random P-Forest may be over-parameterized and challenging for visualization and interpretation, but it offers several advantages.
(i) Random P-Forest addresses model uncertainty and can evaluate characteristic importance by multiple bootstrap samples. In other words, the single P-Tree is readily interpretable as it visually depicts characteristics for non-linearities and asymmetric interactions, but the model's efficacy may be limited if we cannot demonstrate that it does not overfit. In constrast, random P-Forest utilizes multiple random bootstrap samples, which helps mitigate the risk of overfitting.
(ii) Random P-Forest offers an alternative to boosting for creating multiple test assets. Unlike boosting, which generates assets sequentially, random P-Forest independently creates assets across P-Trees.

In addition to the OOS evaluation in Table \ref{tab:factors_a} for assessing model overfitting, we also compare key characteristics between the random P-Forest and the single P-Tree.
The selected characteristics of the single P-Tree are also shown as important by the random P-Forest, suggesting that the single P-Tree is not overfitted.
The consistent results may be attributed to the P-Tree's global split criteria, which prevent overfitting, similar to ML ensemble methods, while providing greater interpretability and transparent insights into non-linearities and asymmetric interactions.
The significant characteristics obtained from the random P-Forest, such as earning surprise (\texttt{SUE}), dollar trading volume (\texttt{DOLVOL}), and industry-adjusted book-to-market ratio (\texttt{BM\_IA}), are also chosen in the top splits of P-Tree for the 40 years, as illustrated in Figure \ref{fig:tree_a}.
In addition, the important characteristics listed in Table \ref{tab:vi_split} are also selected in the top splits displayed in the P-Tree diagram for the subsample analysis (refer to Figure \ref{fig:tree_a_T1_T2}).

\subsection{Random P-Forest and Complexity}
\label{sec:random_p_forest_complex}

Our random P-Forest also relates to the growing popularity of large models in finance. In this recent literature, researchers usually question whether model performance improves as the model size increases, which joins the recent statistical literature on "Benign Overfitting" and "High-Dimensional Interpolation" \citep[e.g.,][]{belkin2019reconciling,hastie2022surprises} and "the Virtue of Complexity" in financial applications \citep[e.g.,][]{kelly2022virtueEverywhere,kelly2024virtue,didisheim2024complexity}. 
This literature shows that large regularized models, which have a large number of parameters relative to the number of observations, can also perform well out-of-sample, and their performance even improves as the number of parameters increases.
This goes against the traditional wisdom of using parsimonious modeling in statistics and finance.

Random P-Forest constructs numerous uncorrelated P-Trees independently on randomly bootstrapped samples, allowing it to be expanded as a large model with both statistical and economic regularizations. 
These hundreds of P-Trees and thousands of leaf basis portfolios are similar to the randomly generated portfolios in \cite{didisheim2024complexity}.
We can estimate the SDF (tangency portfolio) using all the leaf basis portfolios in the forest and report their OOS performances to demonstrate the large model performance.

We create this random P-Forest SDF by (i) randomly selecting $L$ characteristics for a P-Tree, growing P-Tree with split criteria \eqref{eqn:criteria}. 
(ii) repeating the procedure (i) for $B$ times to form a forest of $B$ P-Trees with parallel computing. (iii) estimating the SDF weight on all the leaf basis portfolios in the random P-Forest with \eqref{eqn:sdf_random_p_forest} based on in-sample data. 
We adopt the \textit{ridge (ridgeless) SDF estimator} in \cite{didisheim2024complexity} to estimate the SDF weights $\hat w$ on leaf basis portfolios:\footnote{\cite{britten1999sampling} introduces a regression approach to estimate the tangency portfolio and \cite{ao2019approaching} extend the framework to the regularized portfolio by allowing a large number of assets.
}
\vspace{-0.2cm}
\begin{equation} \label{eqn:sdf_random_p_forest}
    \hat w(\gamma) = \argmin_w E\left[ \left(1-w^{\prime} \RR_t\right)^2 + \gamma \| w \| ^2 \right], 
\end{equation}
where $\gamma$ is a shrinkage parameter.
We also investigate the large model performance and define $c=P/T$ as the degree of parameterization or \textit{complexity}, where $P$ is the number of leaf basis portfolios, and $T=240$ is the fixed estimation rolling window. 
We examine the effects of extensive parameterization and shrinkage estimation on the random P-Forest SDF across various values for $c$ and $\gamma$.
In addition to the OOS Sharpe ratio, we also report the OOS Pricing Error $ = E_{OOS}[(1-\hat w^\prime \RR_t)^2]$.
The Pricing Error is the OOS HJ distance \citep{hansen1997assessing}, which has the above expression when $P > T_{OOS}$ and both are sufficiently large.

\begin{figure}[h!]

\caption{\textbf{OOS Performance of Random P-Forest SDF}}\label{fig: sr_split_rpf_10}
    {\footnotesize
    This figure reports the Sharpe ratio, and pricing error (HJD) of the realized OOS SDF portfolio. 
    Each P-Tree in the Random P-Forest is fed with 10 characteristics, randomly drawn without replacement from 61 ones, and generates 10 leaf basis portfolios.
    These P-Trees are independently trained.
    P-Trees are split with goal-oriented criteria to maximize the collective Sharpe ratio of leaf basis portfolios.
    The total number of leaf basis portfolios is denoted $P$.
    The horizontal axis shows model complexity $c = P/T$, with $c$ ranging from 0.1 to 100 and $T = 240$ months.
    We report five specifications with shrinkage parameter $\gamma$ in $[10^{-5}, 10^{-1}, 1, 10, 1000]$.
    }
    
    \vspace{-0.2cm}
    
    \begin{center}
        
    \begin{subfigure}{0.45\textwidth}
        \begin{center} 
            \includegraphics[height=0.95\textwidth]{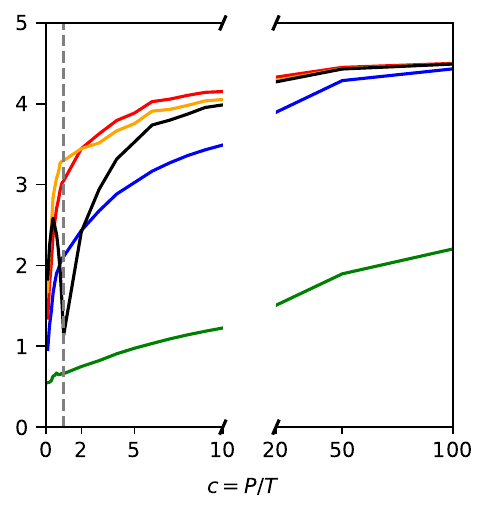}
            \caption{\footnotesize{Sharpe Ratio}}
        \end{center} 
    \end{subfigure}
    \hfill
    
    \begin{subfigure}{0.45\textwidth}
        \begin{center} 
            \includegraphics[height=0.95\textwidth]{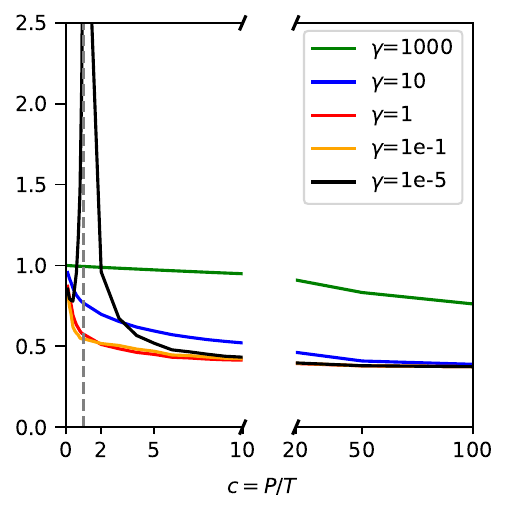}
            \caption{\footnotesize{Pricing Error}}
        \end{center}
    \end{subfigure}
    
    \end{center}
    
\vspace{-0.8cm}
    
\end{figure}

Figure \ref{fig: sr_split_rpf_10} displays the OOS performances of the random P-Forest SDF with $L=10$ random characteristics from 2001 to 2020.
As $c$ increases, we find
(i) the Sharpe ratio exhibits double ascent for low shrinkage cases and permanent ascent for high shrinkage cases, and 
(ii) the pricing error decreases for high shrinkage cases, with a spike around $c = 1$ for low shrinkage cases.
These patterns are similar to those in \cite{didisheim2024complexity}, demonstrating the benefit of randomness and large model size in the OOS performance of the random P-Forest SDF. 
For $c=10$, the Sharpe ratio is about 4.0, and the pricing error is below 0.44. For $c=100$, the Sharpe ratio becomes about 4.5, and the pricing error is below 0.38.

\paragraph{Goal-oriented P-Tree v.s. random split P-Tree.}
The random P-Forest SDF involves randomness as a subset of characteristics is randomly drawn for each P-Tree, but split decisions are not made randomly. Goal-oriented search in P-Tree helps the leaf basis portfolios to maximize the collective Sharpe ratio efficiently.
However, \cite{didisheim2024complexity} show that random portfolios can also generate the SDF with excellent OOS performances if the number of portfolios is large enough.
We show the significance of goal-oriented search in maximizing mean-variance efficiency in an over-parameterized setting by presenting an alternative SDF specification for comparison. 
Specifically, the \textit{random split SDF} uses random split criteria to create leaf basis portfolios, with no specific split criterion and characteristics and splitting values being randomly chosen for each decision to generate ten leaf basis portfolios.

The random P-Forest SDF exhibits comparable patterns to the random split SDF, but the random P-Forest is significantly more efficient regarding parameterization and computational expense.
Specifically, it requires less complexity to achieve the same level of Sharpe ratio and pricing error as the random split SDF. 
Therefore, goal-oriented search improves the efficiency of large models with both statistical and economic regularizations, aligning with the idea of combining economic objectives with ML models in finance research.
Internet Appendix \ref{sec:random_split_SDF} elaborates further.

Finally, from the test asset construction perspective, the goal-oriented P-Tree provides practically useful test assets for asset pricing model estimation and evaluation.
The number of test assets is typically less than 100 because the statistical power of the asset pricing test is low when this number is large \citep{fan2015power}.

\subsection{Evaluating a Characteristic with P-Tree.}
\label{sec:evaluate_char}

As an alternative pricing kernel, P-Tree can directly evaluate a new characteristic's incremental (and nonlinear) contribution over a benchmark to complement the literature. Researchers develop new factors by sorting based on specific characteristics in empirical studies and then test their independent information by controlling benchmark models through spanning regression.
However, empirical studies rarely examine whether the associated characteristics provide new information and may cause information loss.
To address this issue, P-Tree provides a new perspective for evaluating the usefulness of characteristics in generating test assets and latent factor models, given a set of benchmark characteristics, such as \texttt{ME} and \texttt{BM}.

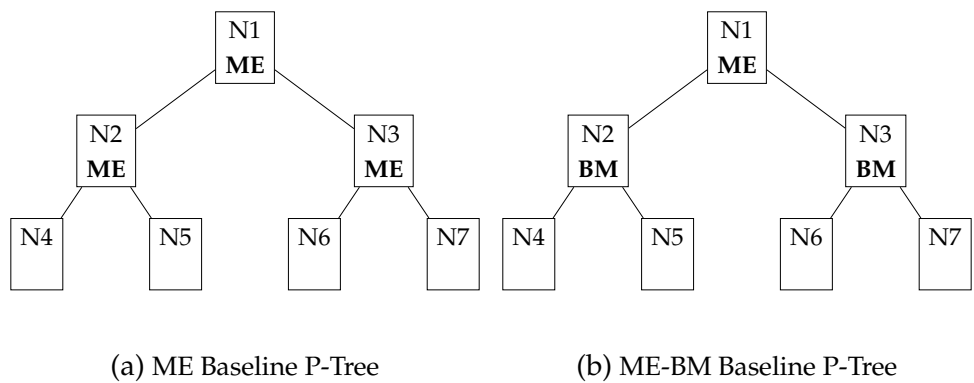
\begin{figure}[h!]
    \centering
    \caption{\bf Demonstration for ME baseline and ME-BM baseline P-Trees}
    \label{fig:me-bm-p-tree-main}
    \begin{subfigure}{0.4\textwidth}
        \resizebox{\textwidth}{!}{
    \begin{tikzpicture}
        [
        grow                    = down,
        edge from parent/.style = {draw},
        ]
    
    \node [env] { \fs{N1} \\ \textbf{\fs{ME}} } 
      child { node [env] { \fs{N2} \\ \textbf{\fs{ME}} } 
            child { node [env] { \fs{N4} \\ } 
                  edge from parent node [left] {} 
            }
            child { node [env] { \fs{N5} \\ } 
                  edge from parent node [right] {} 
            }
            edge from parent node [left] {} 
      }
      child { node [env] { \fs{N3} \\ \textbf{\fs{ME}} } 
            child { node [env] { \fs{N6} \\ } 
                  edge from parent node [left] {} 
            }
            child { node [env] { \fs{N7} \\ } 
                  edge from parent node [right] {} 
            }
            edge from parent node [right] {}
      };
    \end{tikzpicture}
    }
    \vspace{0.1cm}
    \caption{\footnotesize{ME Baseline P-Tree}}
    \end{subfigure}
    \begin{subfigure}{0.4\textwidth}
        \resizebox{\textwidth}{!}{

    \begin{tikzpicture}
        [
        grow                    = down,
        edge from parent/.style = {draw},
        ]

    \node [env] { \fs{N1} \\ \textbf{\fs{ME}} } 
      child { node [env] { \fs{N2} \\ \textbf{\fs{BM}} } 
            child { node [env] { \fs{N4} \\ } 
                  edge from parent node [left] {} 
            }
            child { node [env] { \fs{N5} \\ } 
                  edge from parent node [right] {} 
            }
            edge from parent node [left] {}
      }
      child { node [env] { \fs{N3} \\ \textbf{\fs{BM}} } 
            child { node [env] { \fs{N6} \\ } 
                  edge from parent node [left] {} 
            }
            child { node [env] { \fs{N7} \\ } 
                  edge from parent node [right] {} 
            }
            edge from parent node [right] {} 
      };
    \end{tikzpicture}
    }
    \vspace{0.1cm}
    \caption{\footnotesize{ME-BM Baseline P-Tree}}
    \end{subfigure}
\vspace{-0.4cm} 
\end{figure}

In Figure \ref{fig:me-bm-p-tree-main}, we show two baseline P-Trees: a three-layer ME baseline P-Tree with all splits based on \texttt{ME} and a 3-layer ME-BM baseline P-Tree with the first layer split on \texttt{ME} and the second layer split on \texttt{BM}.
To assess the incremental utility of estimating the efficient frontier of a characteristic against \texttt{ME}, we extend the ME baseline P-Tree with an additional split on the characteristic of interest or \texttt{ME}.
One might expect a split to increase the Sharpe ratio of an MVE portfolio for a shallow P-Tree.
Therefore, We only assess if a characteristic offers new information by comparing the Sharpe ratio improvement when adding it to the baseline P-Tree over a split on \texttt{ME}. This method helps us gauge the incremental impact of the characteristic in question.

\begin{figure}[h!]
	\caption{\bf Evaluating a Characteristic with P-Tree}
	\label{fig:char_increment}
	\footnotesize{
	This figure shows the Sharpe ratio of P-Tree basis portfolios after splitting on a characteristic, based on either ME or ME-BM baseline P-        Trees. Sub-figures (a) and (b) display the results for ME and ME-BM 
        baseline P-Trees, respectively.
        We sort the characteristics by their Sharpe ratios in ascending order. The red lines show the Sharpe ratios of P-Tree basis portfolios after splitting one step further on the controlling characteristics.
        A bar above the red line indicates incremental information against the benchmark characteristics and is colored black. Otherwise, the grey bar indicates the characteristic does not provide incremental information. The characteristics are listed in Table \ref{tab:chars}.
	}
\vspace{-0.2cm}	
	\begin{center}
		\begin{subfigure}[b]{0.9\textwidth}
			
			\centering
			\includegraphics[width=\textwidth]{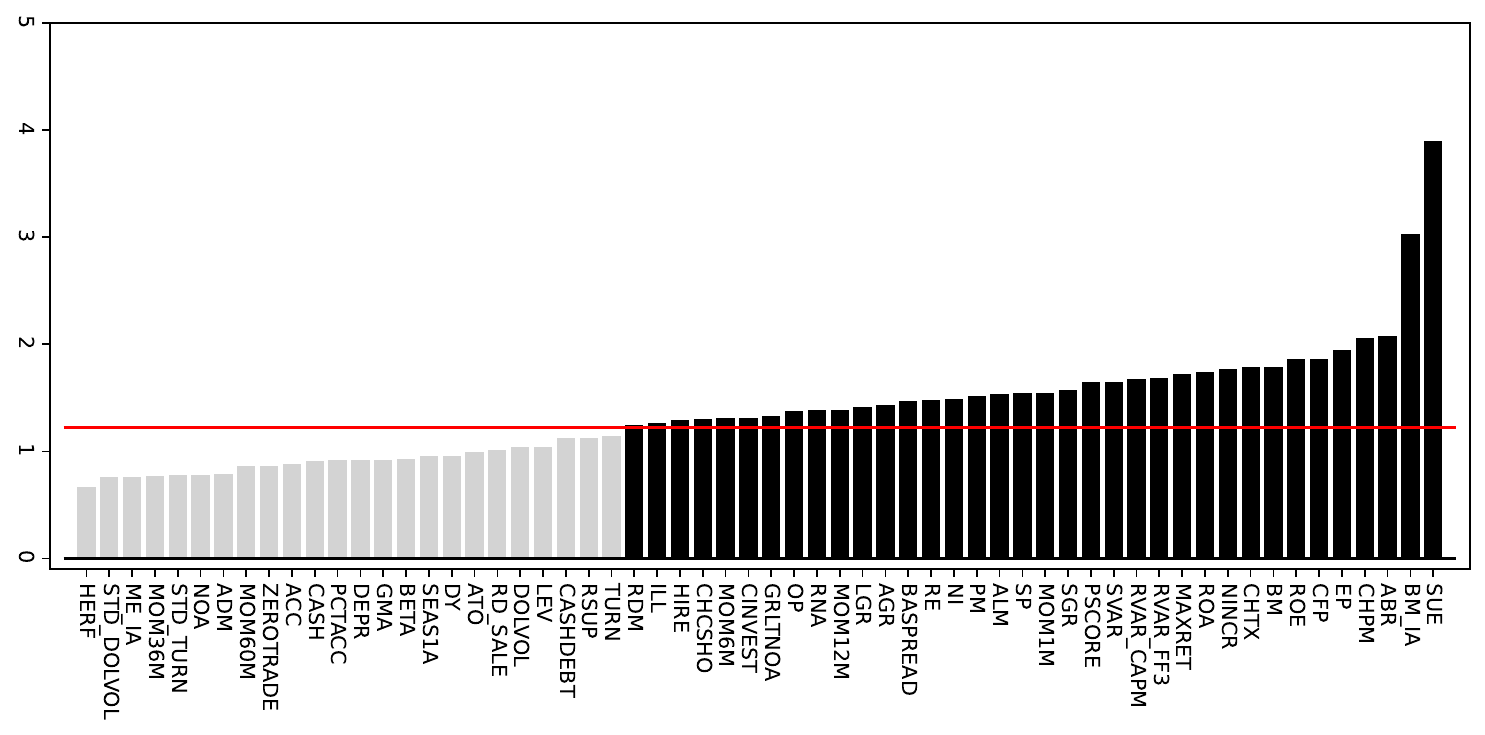}
                \vspace{-0.6cm}
			\caption*{\footnotesize{(a) Controlling ME}}
		\end{subfigure}
		\\
		\begin{subfigure}[b]{0.9\textwidth}
			
			\centering
			\includegraphics[width=\textwidth]{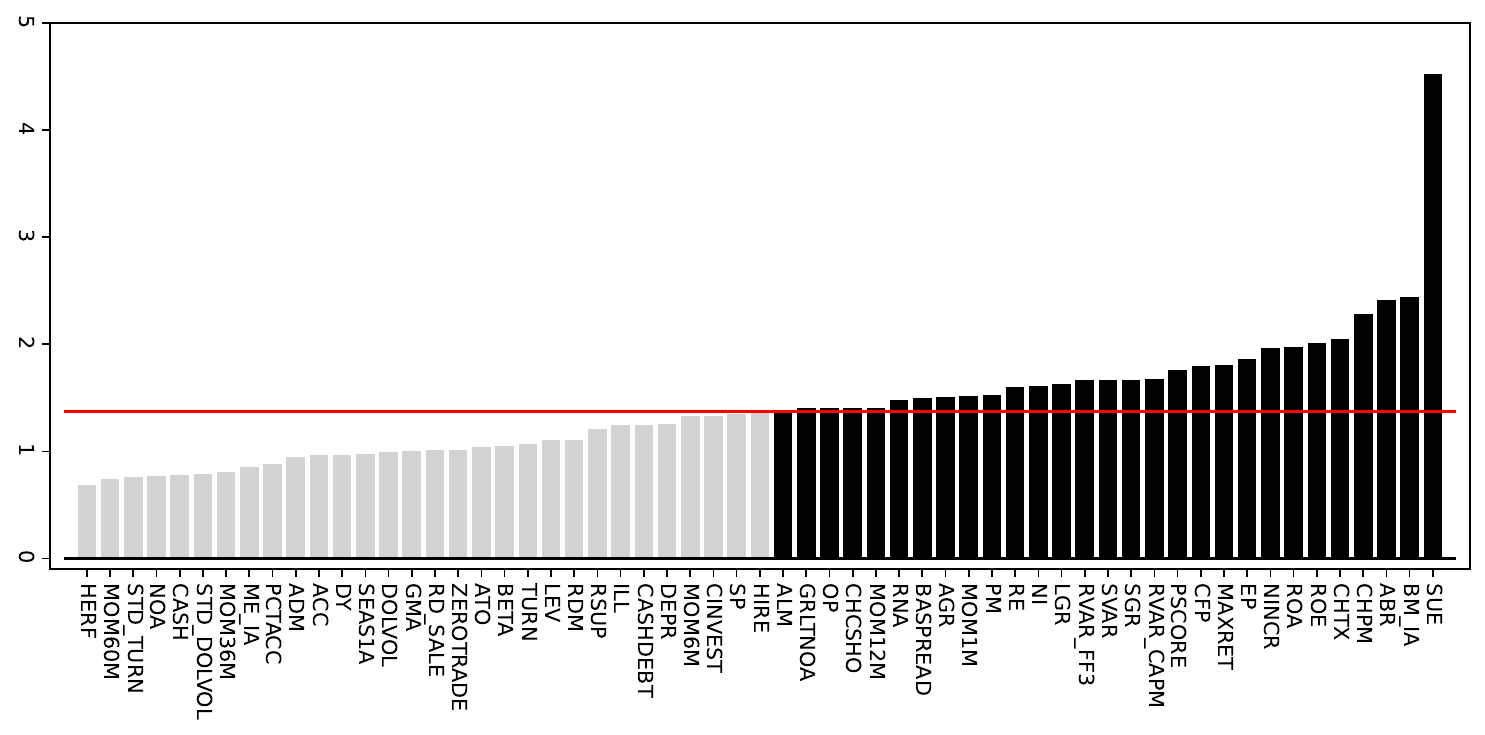}
                \vspace{-0.6cm}
			\caption*{\footnotesize{(b) Controlling ME-BM}}
		\end{subfigure}
	\end{center}
\vspace{-0.5cm} 
\end{figure}

In Figure \ref{fig:char_increment}, subfigure (a) shows the incremental Sharpe ratio of each characteristic on the ME baseline P-Tree. We highlight characteristics outperforming \texttt{ME} in black and those underperforming \texttt{ME} in grey.
We also sort the characteristics by their Sharpe ratios. 
Even for a shallow P-Tree, 24 characteristics do not provide incremental information against the ME baseline P-Tree.
Further, we identify categories other than size, including momentum (\texttt{SUE, ABR, NINCR}), value (\texttt{BM\_IA, BM, CFP, EP}), profitability (\texttt{CHPM, ROA, ROE}), and frictions/volatility (\texttt{MAXRET, SVAR}), that vastly improve investment performance by a simple further split on the ME baseline P-Tree.

In summary, we assess each characteristic on a benchmark P-Tree by controlling \texttt{ME} and/or \texttt{BM} for its marginal contribution to the investment objective. We can expect these marginal contributions to be smaller when considering a deeper tree as the benchmark.
The relative importance of a characteristic may change when the benchmark P-Tree or controlling characteristics are altered, highlighting the adaptable nature of P-Tree evaluation. This evaluation can complement the conventional factor-spanning regression in evaluating characteristics.

Similarly, we assess each characteristic using the ME-BM baseline P-Tree shown in subfigure (b) of Figure \ref{fig:me-bm-p-tree-main}. This P-Tree has three layers and is based on \texttt{ME} and \texttt{BM}.
With a stronger benchmark, some characteristics, such as \texttt{RDM, ILL, HIRE, MOM6M, CINVEST, SP}, do not provide additional information on this upgraded ME-BM baseline P-Tree.
Against the ME-BM baseline P-Tree, we find that 30 characteristics do not offer incremental information for estimating the efficient frontier.
Consistent with the ME baseline P-Tree results, momentum, value, and profitability characteristics are useful in the nonlinear interactive model.

\subsection{P-Tree with Macroeconomic Regimes} \label{sec:macro}

Increasing empirical evidence indicates factor risk premia and loadings can vary significantly under various macroeconomic conditions.
Given that the cross-sectional driving forces may change over time, examining how P-Tree performs in different states and which characteristics are selected for the mean-variance diversification is valuable.
This subsection presents the subsample analysis for implementing P-Trees under various macro regimes and demonstrates their robust performance.

\paragraph{Characteristics evaluation under regimes.} 
We conduct robustness checks by analyzing ten macro variables from Table \ref{tab:macro} for subsample analysis in different regimes. The empirical results of P-Tree models under various regimes are shown in Table \ref{tab:regime_switch}.
We divide the sample into top and bottom regimes using the 50\% 10-year rolling-window percentiles and fit a separate model for each regime. This allows us to assess the effectiveness of P-Tree in both positive and negative macroeconomic environments.\footnote{For identifying macro regime shifts through time-series splits, \cite{feng2024currency} adapt the P-Tree framework to study the currency return dynamics with respect to U.S. macro regimes.}

First, the useful characteristics may vary depending on different macroeconomic conditions. P-Tree performs well, adjusts to macroeconomic changes, and chooses different characteristics for the top three splits.
For example, when the term spread is low, or market volatility is high, the roles of \texttt{SUE} and \texttt{DOVOL} for firm fundamentals and trading volume are more important for cross-sectional mean-variance diversification.
However, characteristics such as \texttt{BASPREAD} and \texttt{RVAR\_CAPM}, representing frictions for liquidity and volatility, are often chosen for scenarios involving high net equity issuance, low default yield, low market liquidity, or low inflation.

\begin{table}[h!]
  
    \caption{\bf P-Tree Performance Under Regime Switches
  }\label{tab:regime_switch}%
  
    {\footnotesize This table shows the performance of P-Tree models across various macroeconomic states. The time series of each macro variable is split into "Top" and "Btm" regimes using the 50\% 10-year rolling-window percentiles. A simple P-Tree is then built for each regime, and similar performance statistics are reported in Table \ref{tab:test_asset_a}, along with the top three splitting characteristics. The rows labeled "Com" report the average return and Sharpe ratio of the combined P-Tree factor of the "Top" and "Btm" regimes.
   
    \begin{center}
    
 \resizebox{\textwidth}{!}{
     
    \begin{tabular}{l c ccccccccccccc}
        \toprule
        & Regime    &       & AVG   & SR    &       
        & GRS & $p$-value & $\overline{|\alpha|}$ & $\sqrt{\overline{\alpha^2}}$ & $\overline{R^2}$  &  
        & \multicolumn{3}{c}{Top Three Characteristics}  \\
        \hline
        \\
        
        \multirow{3}[0]{*}{DFY} & Top   &       & 1.78  & 6.14  &       & 71.84 & 0.00  & 1.27  & 1.57  & 52.69 &       & SUE   & DOLVOL & CFP \\
        & Btm   &       & 1.21  & 4.56  &       & 40.01 & 0.00  & 1.08  & 1.44  & 60.05 &       & ABR   & RVAR\_CAPM & STD\_DOLVOL \\
        & Com   &       & 1.49  & 5.16  &       &       &       &       &       &       &       &       &       &  \\
        \multirow{3}[0]{*}{DY} & Top   &       & 1.45  & 5.84  &       & 43.17 & 0.00  & 1.03  & 1.30  & 73.16 &       & CHPM  & BM\_IA & ME \\
        & Btm   &       & 1.30  & 6.21  &       & 101.56 & 0.00  & 1.05  & 1.21  & 51.45 &       & ABR   & ZEROTRADE & DOLVOL \\
        & Com   &       & 1.35  & 6.03  &       &       &       &       &       &       &       &       &       &  \\
        \multirow{3}[0]{*}{EP} & Top   &       & 0.99  & 5.51  &       & 34.85 & 0.00  & 0.94  & 1.28  & 78.21 &       & RVAR\_CAPM & ME\_IA & ME \\
        & Btm   &       & 1.34  & 6.04  &       & 98.62 & 0.00  & 0.94  & 1.18  & 56.89 &       & SUE   & DOLVOL & BM\_IA \\
        & Com   &       & 1.24  & 5.76  &       &       &       &       &       &       &       &       &       &  \\
        \multirow{3}[0]{*}{ILL} & Top   &       & 1.10  & 5.90  &       & 89.15 & 0.00  & 0.81  & 1.04  & 62.31 &       & RVAR\_CAPM & SUE   & STD\_DOLVOL \\
        & Btm   &       & 1.50  & 7.44  &       & 72.13 & 0.00  & 1.03  & 1.30  & 59.38 &       & SUE   & DOLVOL & SP \\
        & Com   &       & 1.24  & 6.21  &       &       &       &       &       &       &       &       &       &  \\
        \multirow{3}[0]{*}{INFL} & Top   &       & 1.16  & 4.83  &       & 30.10 & 0.00  & 0.99  & 1.29  & 75.20 &       & CFP   & ATO   & ME \\
        & Btm   &       & 1.29  & 6.61  &       & 107.40 & 0.00  & 0.93  & 1.26  & 57.67 &       & SUE   & DOLVOL & CFP \\
        & Com   &       & 1.25  & 5.89  &       &       &       &       &       &       &       &       &       &  \\
        \multirow{3}[0]{*}{LEV} & Top   &       & 1.33  & 5.10  &       & 38.77 & 0.00  & 0.89  & 1.14  & 65.03 &       & CFP   & ALM   & STD\_DOLVOL \\
        & Btm   &       & 1.39  & 5.36  &       & 66.94 & 0.00  & 1.23  & 1.46  & 52.23 &       & NI    & MOM12M & BM\_IA \\
        & Com   &       & 1.37  & 5.26  &       &       &       &       &       &       &       &       &       &  \\
        \multirow{3}[0]{*}{NI} & Top   &       & 1.37  & 5.05  &       & 45.70 & 0.00  & 1.67  & 2.16  & 53.03 &       & RVAR\_CAPM & ME\_IA & RD\_SALE \\
        & Btm   &       & 1.51  & 6.71  &       & 93.02 & 0.00  & 1.18  & 1.38  & 61.79 &       & ROA   & DOLVOL & SUE \\
        & Com   &       & 1.45  & 5.81  &       &       &       &       &       &       &       &       &       &  \\
        \multirow{3}[0]{*}{SVAR} & Top   &       & 1.24  & 6.22  &       & 72.03 & 0.00  & 0.94  & 1.24  & 62.66 &       & SUE   & DOLVOL & CFP \\
        & Btm   &       & 1.23  & 4.78  &       & 43.92 & 0.00  & 1.18  & 1.38  & 59.41 &       & ROA   & ZEROTRADE & ABR \\
        & Com   &       & 1.23  & 5.36  &       &       &       &       &       &       &       &       &       &  \\
        \multirow{3}[0]{*}{TBL} & Top   &       & 1.09  & 7.15  &       & 59.53 & 0.00  & 0.75  & 0.91  & 61.98 &       & CHCSHO & SUE   & DOLVOL \\
        & Btm   &       & 1.30  & 7.05  &       & 134.30 & 0.00  & 0.93  & 1.10  & 63.44 &       & SUE   & DOLVOL & CFP \\
        & Com   &       & 1.24  & 6.97  &       &       &       &       &       &       &       &       &       &  \\
        \multirow{3}[0]{*}{TMS} & Top   &       & 1.17  & 6.01  &       & 67.58 & 0.00  & 0.99  & 1.28  & 63.45 &       & EP    & STD\_DOLVOL & SUE \\
        & Btm   &       & 1.23  & 5.82  &       & 70.00 & 0.00  & 0.86  & 1.17  & 52.52 &       & SUE   & DOLVOL & BM\_IA \\
        & Com   &       & 1.20  & 5.91  &       &       &       &       &       &       &       &       &       &  \\
        
        \bottomrule
        
    \end{tabular}
    } 
    \end{center}
   }
\vspace{-0.3cm} 
\end{table}

Second, P-Tree test assets consistently provide informative results for asset pricing considerations: the GRS tests against Fama-French five factors are rejected for each regime, and the Sharpe ratios of the MVE portfolios spanned by P-Tree test assets are all greater than 7. 
Interestingly, the GRS statistics (weighted pricing errors) are significantly larger for scenarios that involve low inflation or a low t-bill rate, which implies the weakest moments of commonly used factors capturing the cross-sectional signals.
The average alphas are larger than those of ``P-Tree1" in Table \ref{tab:factors_a}, Panel A, implying the test assets under extreme macro regimes are even more challenging to price by common factors. 
Finally, we combine the P-Tree factors in two regimes and find the ``Com" (Combined) factor has larger Sharpe ratios than the first P-Tree factor in Table \ref{tab:factors_a}, Panel A, which implements an unconditional model.

\section{Conclusion} \label{sec:conclusion}

Estimating the mean-variance efficient frontier using individual asset returns and creating diversified test assets for asset pricing model evaluations constitute longstanding empirical challenges \citep[e.g.,][]{Markowitz1952,lewellen2010skeptical,daniel2020cross}.
Our paper introduces a new class of tree-based models, Panel Tree (P-Tree), that effectively addresses these empirical challenges and analyzes the (unbalanced) panel of individual asset returns by generalizing multi-characteristics security sorting and splitting the cross section.
Under the global criteria of mean-variance efficiency, P-Tree utilizes high-dimensional characteristics, which contain rich information on the joint distribution of asset returns, to generate characteristics-managed portfolios 
and recover the stochastic discount factor.
More generally, P-Trees expand tree-based and other ML models beyond pure prediction, effectively analyzing panel data by economic-guided objectives, while maintaining interpretability and handling asymmetric and nonlinear interactions in low signal-to-noise environments.

Our empirical study of U.S. equities shows that P-Tree test assets significantly advance the efficient frontier compared to that spanned by those commonly used test assets (e.g., bivariate- or univariate-sorted portfolios) and exhibit significant unexplained alphas against benchmark models (e.g., Fama-French factors), highlighting the importance of test assets.
These findings remain consistent and robust for the out-of-sample evaluation in the recent two decades.
Second, the P-Tree tangency portfolio is constructed as traded factor models, outperforming popular observable and ML latent factor models for factor investment and cross-sectional pricing performance. 
Third, we demonstrate the versatility and utility of sparse tree-based models for economic interpretation. For example, we identify \texttt{SUE}, \texttt{DOLVOL}, and \texttt{BM\_IA} as key characteristics that interact to explain the cross-section of asset returns. 
Finally, P-Tree captures the complexity of panel stock returns with sparsity, achieving exceptional OOS Sharpe ratios close to those of over-parameterized large models.
Beyond asset pricing, our framework offers an interpretable and computationally efficient alternative to deep-learning-based AI for goal-oriented search in large modeling spaces.

\setstretch{1.2}{\small \bibliography{tree.bib}}

\clearpage
\setstretch{1.6}
\appendix
\appendixpage

\setcounter{section}{0}
\setcounter{table}{0}%
\setcounter{figure}{0}%
\setcounter{page}{1}%

\renewcommand{\thesection}{\Roman{section}}%

\renewcommand\thefigure{A.\arabic{figure}}    
\setcounter{figure}{0}

\renewcommand\thetable{A.\arabic{table}}    
\setcounter{table}{0}

\begin{algorithm}[!h]
	\caption{\bf Growing a Single P-Tree}\label{alg:GT}
	\bigskip
    \small
	\begin{algorithmic}[1]
		\Procedure{GrowTree}{{\tt root}} \\
        \textbf{Input} Asset returns $\rr_{i,t}$ and ranked characteristics $\z_{i,t}$ \\
		\textbf{outcome} Grow P-Tree from the {\tt root} node, form leaf basis portfolios
		\For{$j$ from $1$ to $J$} \Comment{Loop over number of iterations}
		\If{current depth $\geq$ $d_\text{max}$} 
		\State \textbf{return}.
		\Else
		\State Search the tree, find all leaf nodes $\mathcal{N}$
		\For{each leaf node $N$ in $\mathcal{N}$} \Comment{Loop over all current leaf nodes}
		\For{each candidate split $\tilde{\cc}_{k,m,N}$ in $\mathcal{C}_N$}
		\State Partition data temporally in $N$ according to $\tilde{\cc}_{k,m,N}$.
		\If{Either left or right child of $N$ does not satisfy minimal leaf size}
		\State $\mathcal{L}(\tilde{\cc}_{k,m,N}) = - \infty$.
		\Else
		\State Calculate leaf basis portfolios. 
		\State Estimate tangency portfolio using \textit{all} leaf basis portfolios as in (\ref{eqn:SDF}).
		\State Calculate the split criteria $\mathcal{L}(\tilde{\cc}_{k,m,N})$ in (\ref{eqn:criteria}).
		\EndIf
		\EndFor
		\EndFor
		\State Find the best leaf node and split rule that maximize split criteria $$\tilde{\cc}_{j} = \argmax_{N \in \mathcal{N}, \tilde{\cc}_{k,m,N} \in \mathcal{C}_N}\{\mathcal{L}(\tilde{\cc}_{k,m,N})\}$$
		\State Split the node selected with the $j$-th split rule of the tree $\tilde{\cc}_{j}$. 
		\EndIf
		\EndFor
		\State \textbf{return} $\left\{\RR_{t}^{(J)}, f_t^{(J)} \right\}$ 
		\EndProcedure
	\end{algorithmic}
\end{algorithm}


\begin{figure}[ht!]
    \caption{\bf Panel Tree Diagram for Subsamples
    }
    {\footnotesize
    Format follows Figure \ref{fig:tree_a}. Subfigure (a) shows P-Tree in the sample from 1981-2000. Subfigure (b) shows P-Tree in the sample from 2001-2020.
    }
    \label{fig:tree_a_T1_T2}

    \vspace{0.2cm}

    \begin{subfigure}[b]{\textwidth}
        \centering
        \includegraphics[width=\textwidth]{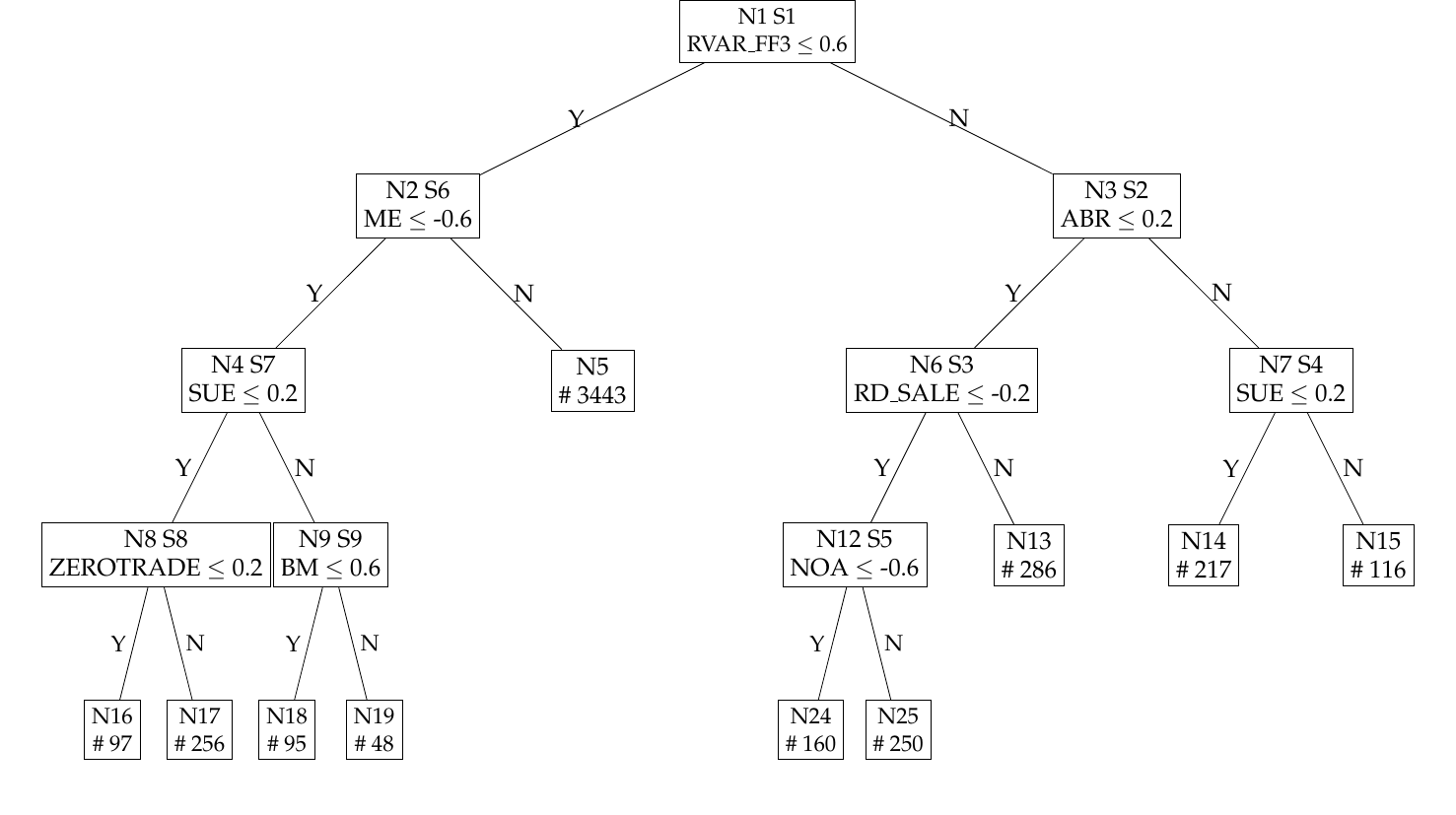}    
        \caption{20-year Sample (1981-2000)}
    \end{subfigure}

    \begin{subfigure}[b]{\textwidth}
        \centering
        \includegraphics[width=\textwidth]{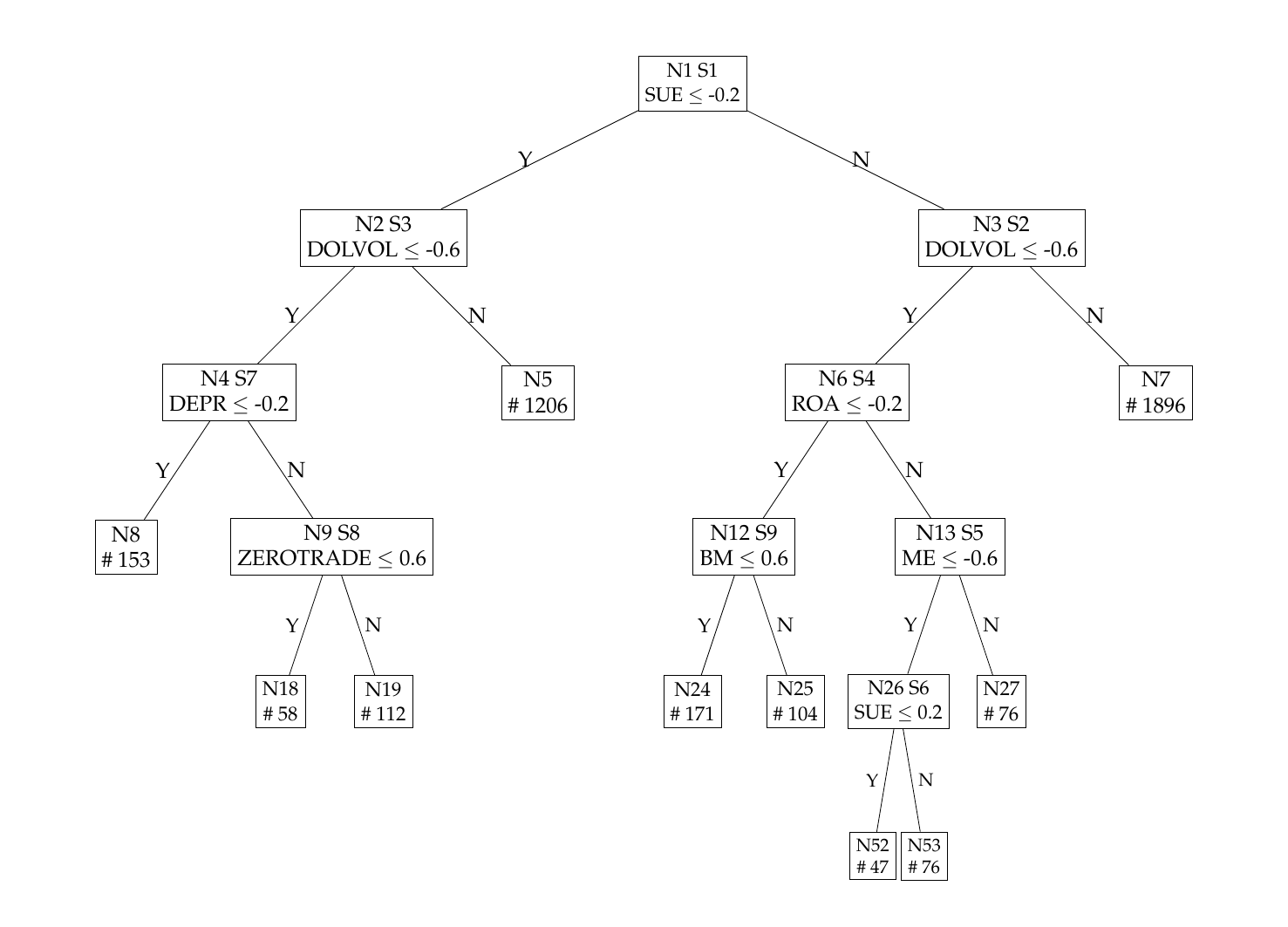}    
        \caption{20-year Sample (2001-2020)}
    \end{subfigure}
    
\end{figure}

\clearpage


\begin{table}[ht!]
	\caption{\bf Equity Characteristics 
    }
	\label{tab:chars}
	\footnotesize{
		This table lists the description of 61 characteristics used in the empirical study. 
		
		\begin{center}
			
    \resizebox{\textwidth}{!}{

            \begin{tabular}{l llllll}
				\toprule
				No.    &  & Characteristics       &  & Description                                   &  & Category\\               \midrule
				1      &        & ABR    &        & Abnormal returns around earnings announcement    &  & Momentum\\
				2      &        & ACC    &        & Operating accruals                               &  & Investment\\
				3      &        & ADM    &        & Advertising expense-to-market                    &  & Intangibles\\
				4      &        & AGR    &        & Asset growth                                     &  & Investment\\
				5      &        & ALM    &        & Quarterly asset liquidity                        &  & Intangibles\\
				6      &        & ATO    &        & Asset turnover                                   &  & Profitability\\
				7      &        &BASPREAD&        & Bid-ask spread (3 months)                        &  & Frictions\\
				8      &        & BETA   &        & Beta (3 months)                                  &  & Frictions\\
				9      &        & BM     &        & Book-to-market equity                            &  & Value-versus-growth\\
				10     &        & BM\_IA &        & Industry-adjusted book to market                 &  & Value-versus-growth\\
				11     &        & CASH   &        & Cash holdings                                    &  & Value-versus-growth\\
				12     &        &CASHDEBT&        & Cash to debt                                     &  & Value-versus-growth\\
				13     &        & CFP    &        & Cashflow-to-price                                &  & Value-versus-growth\\
				14     &        & CHCSHO &        & Change in shares outstanding                     &  & Investment\\
				15     &        & CHPM   &        & Change in Profit margin                          &  & Profitability\\
				16     &        & CHTX   &        & Change in tax expense                            &  & Momentum\\
				17     &        & CINVEST&        & Corporate investment                             &  & Investment\\
				18     &        & DEPR   &        & Depreciation / PP\&E                             &  & Momentum\\
				19     &        & DOLVOL &        & Dollar trading volume                            &  & Frictions\\
				20     &        & DY     &        & Dividend yield                                   &  & Value-versus-growth\\
				21     &        & EP     &        & Earnings-to-price                                &  & Value-versus-growth\\
				22     &        & GMA    &        & Gross profitability                              &  & Investment\\
				23     &        & GRLTNOA&        & Growth in long-term net operating assets         &  & Investment\\
				24     &        & HERF   &        & Industry sales concentration                     &  & Intangibles\\
				25     &        & HIRE   &        & Employee growth rate                             &  & Intangibles\\
				26     &        & ILL    &        & Illiquidity rolling (3 months)                   &  & Frictions\\
				27     &        & LEV    &        & Leverage                                         &  & Value-versus-growth\\
				28     &        & LGR    &        & Growth in long-term debt                         &  & Investment\\
				29     &        & MAXRET &        & Maximum daily returns (3 months)                 &  & Frictions\\
				30     &        & ME     &        & Market equity                                    &  & Frictions\\
				31     &        & ME\_IA &        & Industry-adjusted size                           &  & Frictions\\
				32     &        & MOM12M &        & Cumulative returns in the past (2-12) months     &  & Momentum\\
				33     &        & MOM1M  &        & Previous month return                            &  & Momentum\\
				34     &        & MOM36M &        & Cumulative returns in the past (13-35) months    &  & Momentum\\ 
				35     &        & MOM60M &        & Cumulative returns in the past (13-60) months    &  & Momentum\\ 
				36     &        & MOM6M  &        & Cumulative returns in the past (2-6) months      &  & Momentum\\ 
				37     &        & NI     &        & Net equity issue                                 &  & Investment\\
				38     &        & NINCR  &        & Number of earnings increases                     &  & Momentum\\
				39     &        & NOA    &        & Net operating assets                             &  & Investment\\
				40     &        & OP     &        & Operating profitability                          &  & Profitability\\
				\bottomrule
   
            \end{tabular}
			
   }
		\end{center}
	}
\end{table}

\clearpage

\begin{table}[ht!]
	\caption*{\bf Continue: Equity Characteristics}
	\label{tab:chars2}
	\footnotesize{
		
		\begin{center}
			
            \begin{tabular}{l llllll}

				\toprule
				No.    &  & Characteristics        &  & Description                                 &  & Category\\                \midrule
				41     &        & PCTACC &        & Percent operating accruals                      &  & Investment\\
				42     &        & PM     &        & Profit margin                                   &  & Profitability\\
				43     &        & PS     &        & Performance Score                               &  & Profitability\\
				44     &        &RD\_SALE&        & R\&D-to-sales                                   &  & Intangibles\\
				45     &        & RDM    &        & R\&D-to-market                                  &  & Intangibles\\
				46     &        & RE     &        & Revisions in analysts' earnings forecasts       &  & Intangibles\\
				47     &        & RNA    &        & Return on net operating assets                  &  & Profitability\\
				48     &        & ROA   &        & Return on assets                                 &  & Profitability\\
				49     &        & ROE    &        & Return on equity                                &  & Profitability\\
				50     &        & RSUP   &        & Revenue surprise                                &  & Momentum\\
				51     &        & RVAR\_CAPM &        & Idiosyncratic volatility - CAPM (3 months)         &  & Frictions\\
				52     &        & RVAR\_FF3 &        & Res. var. - Fama-French 3 factors (3 months) &  & Frictions\\
				53     &        & SVAR &        & Return variance (3 months)                        &  & Frictions\\
				54     &        & SEAS1A &        & 1-Year Seasonality                              &  & Intangibles\\
				55     &        & SGR    &        & Sales growth                                    &  & Value-versus-growth\\
				56     &        & SP     &        & Sales-to-price                                  &  & Value-versus-growth\\
				57     &        & STD\_DOLVOL &        & Std of dollar trading volume (3 months)    &  & Frictions\\
				58     &        & STD\_TURN &        & Std. of Share turnover (3 months)            &  & Frictions\\
				59     &        & SUE    &        & Standardized unexpected quarterly earnings                   &  & Momentum\\
				60     &        & TURN   &        & Shares turnover                                 &  & Frictions\\
				61     &        & ZEROTRADE &        & Number of zero-trading days (3 months)       &  & Frictions\\
				\bottomrule
			
            \end{tabular}
            
		\end{center}
	}
\end{table}

\vspace{1cm}

\begin{table}[ht!]
	\caption{\bf Macroeconomic Variables}
	\label{tab:macro}
	\footnotesize{
		This table lists the description of macro variables used in the empirical study. 
		
		\begin{center}

            \begin{tabular}{l llll}
            
				\toprule
				No.    &  & Variable Name        &  & Description                                                \\ \midrule
				1      &        & DFY    &        & Default yield \\				
				2      &        & DY     &        & Dividend yield of S\&P 500 \\
                3      &        & EP     &        & Earnings-to-price of S\&P 500 \\
				4      &        & ILL    &        & Pastor-Stambaugh illiquidity \\
				5      &        & INFL   &        & Inflation \\
                6      &        & LEV    &        & Leverage of S\&P 500 \\
				7      &        & NI     &        & Net equity issuance of S\&P 500 \\
				8      &        & SMVAR  &        & Stock Market (S\&P 500) Volatility  \\
				9      &        & TBL    &        & Three-month treasury bill rate \\
				10     &        & TMS    &        & Term spread \\
				\bottomrule
			
               \end{tabular}
               
		\end{center}
	}
\end{table}

\clearpage

\begin{table}[ht!]
  
  \caption{\bf  Subsample Analysis for Testing the Boosted P-Tree Growth}
  \label{tab:app_factor_dim_a}%
  
  {\footnotesize
    This table reports the subsample Analysis for results in Table \ref{tab:factor_dim_a}. Panel B is for the first 20-year sample from 1981–2000, and panel C is for the recent 20-year sample from 2001-2020.
    
  \begin{center}
    \resizebox{\textwidth}{!}{
      \begin{tabular}{l ccccccccccccccc}

        \toprule
        
        &
        & \multicolumn{2}{c}{Sharpe Ratio} &
        & \multicolumn{2}{c}{CAPM Test} &
        & \multicolumn{2}{c}{FF5 Test}   &
        & \multicolumn{3}{c}{Expanding Factors Test}   &
        & \multicolumn{1}{c}{BS Test} \\
        
        \cline{3-4} \cline{6-7} \cline{9-10} \cline{12-14} \cline{16-16}

        &
        & Single    & Multiple &
        & $\alpha$ (\%) & $t$-stat & 
        & $\alpha$ (\%) & $t$-stat & 
        & $\alpha$ (\%) & $t$-stat & $R^2$ & 
        & $p$-value \\
        \hline

        \\
        \multicolumn{16}{c}{\underline{Panel B: 20 Years (1981-2000)}}\\
        \\
        
        1     &       & 7.12  & 7.12  &       & 1.86  & 29.62 &       & 1.78  & 30.22 &       & -     & -     & -     &       &  \\
        2     &       & 1.09  & 7.75  &       & 0.25  & 3.66  &       & 0.11  & 1.91  &       & -0.83 & -4.80 & 0.28  &       & 0.00 \\
        3     &       & 2.26  & 9.45  &       & 0.39  & 9.72  &       & 0.34  & 8.48  &       & 0.92  & 8.19  & 0.17  &       & 0.00 \\
        4     &       & 6.48  & 11.35 &       & 1.80  & 26.96 &       & 1.76  & 27.06 &       & 1.75  & 10.16 & 0.03  &       & 0.00 \\
        5     &       & 1.59  & 12.71 &       & 0.33  & 6.50  &       & 0.28  & 5.77  &       & 1.11  & 6.85  & 0.36  &       & 0.00 \\
        6     &       & 1.65  & 13.67 &       & 0.37  & 5.87  &       & 0.26  & 4.57  &       & -1.03 & -5.34 & 0.40  &       & 0.00 \\
        7     &       & 1.72  & 14.94 &       & 0.33  & 6.99  &       & 0.32  & 6.17  &       & 1.25  & 5.67  & 0.20  &       & 0.00 \\
        8     &       & 1.52  & 15.78 &       & 0.39  & 6.10  &       & 0.31  & 4.93  &       & -1.32 & -5.12 & 0.35  &       & 0.00 \\
        9     &       & 3.02  & 17.90 &       & 1.03  & 14.00 &       & 0.93  & 13.08 &       & 2.69  & 8.42  & 0.26  &       & 0.00 \\
        10    &       & 1.93  & 19.18 &       & 0.83  & 8.12  &       & 0.67  & 7.14  &       & -2.41 & -6.27 & 0.52  &       & 0.00 \\
        11    &       & 1.34  & 20.72 &       & 0.38  & 4.93  &       & 0.19  & 2.76  &       & -2.18 & -7.33 & 0.45  &       & 0.00 \\
        12    &       & 2.22  & 22.89 &       & 0.36  & 9.38  &       & 0.31  & 8.03  &       & 1.46  & 7.36  & 0.39  &       & 0.00 \\
        13    &       & 1.51  & 24.14 &       & 0.35  & 4.88  &       & 0.23  & 3.69  &       & 1.82  & 6.06  & 0.43  &       & 0.00 \\
        14    &       & 2.11  & 25.71 &       & 0.58  & 7.80  &       & 0.48  & 7.57  &       & -2.43 & -5.93 & 0.40  &       & 0.00 \\
        15    &       & 2.39  & 28.37 &       & 0.87  & 10.14 &       & 0.76  & 8.70  &       & 3.69  & 7.64  & 0.52  &       & 0.00 \\
        16    &       & 2.53  & 30.37 &       & 0.53  & 9.58  &       & 0.44  & 7.91  &       & 2.40  & 7.12  & 0.31  &       & 0.00 \\
        17    &       & 1.50  & 32.42 &       & 0.30  & 5.66  &       & 0.31  & 5.08  &       & 2.65  & 7.18  & 0.35  &       & 0.00 \\
        18    &       & 1.63  & 33.89 &       & 0.37  & 5.95  &       & 0.28  & 4.99  &       & -2.79 & -5.64 & 0.37  &       & 0.00 \\
        19    &       & 1.84  & 35.78 &       & 0.47  & 7.04  &       & 0.37  & 5.93  &       & 3.54  & 7.35  & 0.30  &       & 0.00 \\
        20    &       & 2.70  & 37.94 &       & 0.89  & 9.85  &       & 0.79  & 9.06  &       & -4.67 & -6.91 & 0.44  &       & 0.00 \\

        \\
        \multicolumn{16}{c}{\underline{Panel C: 20 Years (2001-2020)}}\\
        \\
        
        1     &       & 5.82  & 5.82  &       & 1.51  & 24.22 &       & 1.47  & 24.26 &       & -     & -     & -     &       &  \\
        2     &       & 2.14  & 6.64  &       & 0.58  & 8.74  &       & 0.52  & 8.26  &       & 0.91  & 6.44  & 0.03  &       & 0.00 \\
        3     &       & 0.91  & 7.53  &       & 0.18  & 3.28  &       & 0.09  & 2.01  &       & -0.86 & -5.82 & 0.36  &       & 0.00 \\
        4     &       & 1.35  & 8.60  &       & 0.29  & 6.18  &       & 0.31  & 6.04  &       & 1.00  & 8.65  & 0.17  &       & 0.00 \\
        5     &       & 1.08  & 9.30  &       & 0.31  & 4.14  &       & 0.26  & 3.88  &       & -1.02 & -6.38 & 0.42  &       & 0.00 \\
        6     &       & 2.63  & 10.36 &       & 0.60  & 12.15 &       & 0.56  & 11.11 &       & 1.04  & 7.61  & 0.12  &       & 0.00 \\
        7     &       & 1.39  & 11.30 &       & 0.29  & 5.53  &       & 0.26  & 5.16  &       & 1.00  & 7.23  & 0.28  &       & 0.00 \\
        8     &       & 1.60  & 11.93 &       & 0.40  & 5.76  &       & 0.30  & 5.47  &       & -0.85 & -5.62 & 0.44  &       & 0.00 \\
        9     &       & 1.04  & 12.73 &       & 0.21  & 4.53  &       & 0.18  & 3.83  &       & -1.01 & -7.44 & 0.36  &       & 0.00 \\
        10    &       & 1.64  & 14.32 &       & 0.30  & 5.96  &       & 0.25  & 5.31  &       & 1.11  & 7.80  & 0.44  &       & 0.00 \\
        11    &       & 1.69  & 15.73 &       & 0.36  & 6.73  &       & 0.30  & 6.20  &       & 1.43  & 6.79  & 0.33  &       & 0.00 \\
        12    &       & 1.58  & 16.82 &       & 0.60  & 6.08  &       & 0.51  & 5.36  &       & 2.48  & 5.72  & 0.24  &       & 0.00 \\
        13    &       & 1.48  & 18.06 &       & 0.41  & 5.29  &       & 0.29  & 4.47  &       & -1.77 & -5.52 & 0.47  &       & 0.00 \\
        14    &       & 1.39  & 19.46 &       & 0.32  & 5.22  &       & 0.26  & 4.62  &       & 1.68  & 6.05  & 0.40  &       & 0.00 \\
        15    &       & 1.48  & 20.60 &       & 0.35  & 5.51  &       & 0.21  & 4.34  &       & -1.38 & -6.25 & 0.58  &       & 0.00 \\
        16    &       & 1.40  & 21.91 &       & 0.34  & 5.84  &       & 0.27  & 4.82  &       & 1.84  & 6.29  & 0.46  &       & 0.00 \\
        17    &       & 1.27  & 23.41 &       & 0.39  & 4.82  &       & 0.25  & 3.46  &       & -2.55 & -6.99 & 0.51  &       & 0.00 \\
        18    &       & 1.44  & 24.36 &       & 0.34  & 5.16  &       & 0.28  & 4.72  &       & -1.61 & -4.97 & 0.48  &       & 0.00 \\
        19    &       & 1.58  & 25.45 &       & 0.33  & 6.59  &       & 0.28  & 6.08  &       & -1.74 & -5.20 & 0.42  &       & 0.00 \\
        20    &       & 1.88  & 26.52 &       & 0.58  & 7.12  &       & 0.50  & 6.53  &       & -2.76 & -4.96 & 0.39  &       & 0.00 \\
      
      \\
      
      \bottomrule
      
      \end{tabular}%
    
    } 

    \end{center}

  }
  
\end{table}%


\clearpage

\setcounter{section}{0}
\setcounter{page}{1}%

\renewcommand\theequation{I.\arabic{equation}}    
\setcounter{equation}{0}%

\renewcommand\thefigure{I.\arabic{figure}}    
\setcounter{figure}{0}

\renewcommand\thetable{I.\arabic{table}}    
\setcounter{table}{0}

\renewcommand{\thesection}{\Roman{section}} 
\renewcommand{\thesubsection}{\thesection.\Alph{subsection}}
\renewcommand{\thesubsubsection}{\thesubsection.\arabic{subsection}}

\begin{center}

{\Large Internet Appendices for \\ ``Growing the Efficient Frontier on Panel Trees''}

\vspace{0.5cm}

{\normalsize
Lin William Cong \hspace{1cm} Guanhao Feng \hspace{1cm} Jingyu He \hspace{1cm} Xin He
}

\vspace{0.5cm}

\end{center}

The Internet appendix is organized as follows. Section \ref{sec:simulation} uses simulation to demonstrate P-Tree's advantage. Section \ref{sec:market_benchmark} illustrates the flexibility of P-Tree in adjusting for benchmark factors such as the market factor. 
Section \ref{sec:random_split_SDF} provides the empirical SDF spanned by random split P-Trees in an over-parameterized environment.

\section{Simulation} 
\label{sec:simulation}

Simulation studies demonstrate the practical capability of P-Tree in selecting characteristics and the efficiency of P-Tree test assets in capturing non-linear interactions among characteristics. The positive simulation evidence provides further support for the modeling ability of P-Tree. The return-generating process is calibrated using the same empirical data sample, considering three genuine characteristics and their interactions and nonlinear terms without loss of generality. To simulate the true investor data environment, a large set of useless or noisy characteristics that correlate with the true characteristics but do not impact generating returns is considered.

\paragraph{Return-generating process.}
The ranked asset characteristics $\z_{i,t}$ (a vector of size $61 \times 1$) are simulated from a VAR(1) model, ensuring that even redundant characteristics still correlate with genuine ones. Redundant characteristics may be selected in scenarios of low signal-to-noise ratios or small sample sizes. The true return-generating process is assumed to involve only three characteristics among all 61 ($\texttt{ME}$, $\texttt{BM}$, and $\texttt{MOM12M}$), incorporating simple nonlinearity and interaction.
\vspace{-0.2cm}
\begin{equation} \label{eqn:dgp_return_20231012}
    r_{i,t} = \texttt{mkt}_t + \kappa \Big[ c_1 \texttt{ME}_{i,t-1} + c_2 \texttt{BM}_{i,t-1} + c_3 \texttt{MEBM}_{i,t-1} + c_4 \texttt{MOM12M}_{i,t-1} + c_5 \texttt{MOM12M}^2_{i,t-1} \Big] + \epsilon_{i,t},
\end{equation}
where $r_{i,t}$ represents the stock excess return, $\texttt{mkt}_t$ denotes the market excess return, $\texttt{MEBM}$ is the interaction of $\texttt{ME}$ and $\texttt{BM}$, $\texttt{MOM12M}^2$ is the quadratic term, and the error terms are simulated from $\epsilon_{i,t} \sim i.i.d. N(0,\sigma^2)$. 
The parameter $\sigma$ is calibrated from real data, and so are $[c_1, \cdots, c_5]$.
The parameter $\kappa = [0.5, 1, 2]$ allows us to adjust the strength of the characteristic signal, with the value of $\kappa = 1$ representing our actual data in the empirical study.

\begin{table}[ht!]
    \caption{\bf Simulation Performance Comparison}
    \label{tab:simu_20231012}%
    {\footnotesize
    The table presents the maximal Sharpe ratios, average absolute market $\alpha$ (\%), and root mean squared $\alpha$ (\%) for different sets of test assets. 
    The average results are reported based on 10 simulations. 
    We test up to 20 P-Trees, each containing 10 leaf basis portfolios.
    The numbers in parentheses represent the average number of test assets. 
    Benchmark results include the following specifications:
    (1) market factor;
    (2) average results for 3 sets of decile portfolios of true char.;
    (3) average results for 61 sets of decile portfolios;
    and
    (4 - 6) bivariate-sort ($5\times5$) portfolios on either two true char., including $\texttt{ME}$ and $\texttt{BM}$, $\texttt{ME}$ and $\texttt{MOM12M}$, and $\texttt{BM}$ and $\texttt{MOM12M}$. 

    \begin{center}
    
    \begin{tabular}{l cccccccccccc}

    \toprule
    &       & \multicolumn{3}{c}{$\kappa=0.5$} 
    &       & \multicolumn{3}{c}{$\kappa=1.0$} 
    &       & \multicolumn{3}{c}{$\kappa=2.0$} \\
    \cline{3-5} \cline{7-9} \cline{11-13}

    &       & SR & $\overline{|\alpha|}$ &  $\sqrt{\overline{\alpha^2}}$ 
    &       & SR & $\overline{|\alpha|}$ &  $\sqrt{\overline{\alpha^2}}$ 
    &       & SR & $\overline{|\alpha|}$ &  $\sqrt{\overline{\alpha^2}}$ 
    \\
    \hline

    \\
    \multicolumn{13}{c}{\underline{Panel A: In-Sample Results}} \\
    \\
    
    P-Tree1 &       & 3.30  & 0.43  & 0.54  &       & 6.14  & 0.73  & 0.94  &       & 12.00 & 1.54  & 1.89 \\
    P-Tree1-5 &       & 4.48  & 0.24  & 0.33  &       & 7.59  & 0.34  & 0.53  &       & 14.23 & 0.63  & 1.00 \\
    P-Tree1-10 &       & 6.13  & 0.19  & 0.27  &       & 9.62  & 0.26  & 0.42  &       & 17.60 & 0.48  & 0.78 \\
    P-Tree1-15 &       & 7.85  & 0.17  & 0.24  &       & 11.96 & 0.24  & 0.37  &       & 21.38 & 0.41  & 0.68 \\
    P-Tree1-20 &       & 9.84  & 0.15  & 0.22  &       & 14.29 & 0.22  & 0.34  &       & 25.69 & 0.37  & 0.61 \\
    
    \\
    MKT(1) &       & 0.55  & 0.00  & 0.00  &       & 0.55  & 0.00  & 0.00  &       & 0.55  & 0.00  & 0.00 \\
    DECILE AVG(30) &       & 1.73  & 0.21  & 0.25  &       & 3.12  & 0.39  & 0.47  &       & 5.97  & 0.76  & 0.92 \\
    DECILE AVG(610) &       & 0.81  & 0.08  & 0.09  &       & 0.97  & 0.10  & 0.12  &       & 1.36  & 0.15  & 0.19 \\
    ME-BM(25) &       & 2.63  & 0.31  & 0.40  &       & 4.92  & 0.57  & 0.78  &       & 9.60  & 1.12  & 1.54 \\
    ME-MOM12M(25) &       & 2.42  & 0.29  & 0.37  &       & 4.49  & 0.55  & 0.70  &       & 8.74  & 1.09  & 1.37 \\
    BM-MOM12M(25) &       & 2.95  & 0.35  & 0.46  &       & 5.60  & 0.68  & 0.89  &       & 11.00 & 1.35  & 1.75 \\

    \hline
    \\
    \multicolumn{13}{c}{\underline{Panel B: Out-of-Sample Results}} \\
    \\
    
    P-Tree1 &       & 2.66  & 0.61  & 0.72  &       & 5.60  & 0.88  & 1.06  &       & 11.37 & 1.56  & 1.90 \\
    P-Tree1-5 &       & 2.34  & 0.58  & 0.64  &       & 5.29  & 0.68  & 0.77  &       & 10.79 & 0.88  & 1.14 \\
    P-Tree1-10 &       & 2.09  & 0.59  & 0.62  &       & 4.94  & 0.65  & 0.72  &       & 10.19 & 0.76  & 0.97 \\
    P-Tree1-15 &       & 1.96  & 0.59  & 0.62  &       & 4.74  & 0.63  & 0.70  &       & 9.87  & 0.72  & 0.91 \\
    P-Tree1-20 &       & 1.90  & 0.59  & 0.62  &       & 4.62  & 0.63  & 0.69  &       & 9.61  & 0.70  & 0.86 \\
    
    \\
    MKT(1) &       & 0.45  & 0.00  & 0.00  &       & 0.45  & 0.00  & 0.00  &       & 0.45  & 0.00  & 0.00 \\
    DECILE AVG(30) &       & 1.55  & 0.20  & 0.25  &       & 3.03  & 0.39  & 0.48  &       & 6.00  & 0.77  & 0.94 \\
    DECILE AVG(610) &       & 0.49  & 0.08  & 0.09  &       & 0.70  & 0.10  & 0.13  &       & 1.17  & 0.16  & 0.20 \\
    ME-BM(25) &       & 2.23  & 0.29  & 0.41  &       & 4.58  & 0.56  & 0.78  &       & 9.20  & 1.12  & 1.54 \\
    ME-MOM12M(25) &       & 2.20  & 0.31  & 0.39  &       & 4.40  & 0.58  & 0.72  &       & 8.70  & 1.12  & 1.41 \\
    BM-MOM12M(25) &       & 2.72  & 0.36  & 0.46  &       & 5.41  & 0.69  & 0.89  &       & 10.73 & 1.35  & 1.76 \\
    
    \bottomrule
    
    \end{tabular}%

    \end{center}
    
    }
    
\vspace{-0.5cm}
\end{table}%

Each simulation sample consists of 1,000 assets, 500 periods for in-sample analysis, and another 500 periods for OOS analysis. Table \ref{tab:simu_20231012} presents the average results of 10 repetitions, including the maximal Sharpe ratios and market alphas of P-Tree test assets. The former displays the tangency portfolio on the efficient frontier, whereas the latter demonstrates the pricing difficulty of the market factor. 

\paragraph{Characteristics-sorted portfolios.}
In Table \ref{tab:simu_20231012}, ``DECILE AVG(30)" presents the average performance of three MVE portfolios and test assets for decile portfolios sorted on three true characteristics. This configuration yields a Sharpe ratio of 3.12, an average absolute $\alpha$ of 0.39\%, and a root mean squared $\alpha$ of 0.47\% when $\kappa = 1$. Meanwhile, the performance of  ``DECILE AVG(30)" is quite consistent in OOS results, because of the pre-specified true characteristics. 
Without knowing the true characteristics, "DECILE AVG(610)" on 61 characteristics reduces the Sharpe ratio to 0.97 and the alphas to 0.10\% and 0.12\%, whereas the OOS results reveal lower Sharpe ratios. Across the different $\kappa$'s, we find higher Sharpe ratios and alphas for large $\kappa$ than for small $\kappa$ cases.

The return-generating process described in \eqref{eqn:dgp_return_20231012} involves interaction and quadratic terms. Using the true characteristics, we also examine bivariate-sorted portfolios ($5\times 5$).
These portfolios consistently yield higher Sharpe ratios and alphas than decile portfolios, demonstrating superior performance. Our findings indicate that accounting for potential characteristic interactions can lead to pricing information gains.

\paragraph{P-Tree test assets.}
P-Trees are an endogenous solution for cross-sectional splitting, test asset creation, and efficient frontier spanning. In Table \ref{tab:simu_20231012}, the P-Tree test assets (``P-Tree1") exhibit an impressive in-sample Sharpe ratio (6.14), absolute alpha (0.73\%), root-mean-squared alpha (0.94\%), OOS Sharpe ratio (5.61), absolute alpha (0.88\%), and root-mean-squared alpha (1.06\%) when $\kappa = 1$. When the $\kappa$ increases to 2, we observe larger Sharpe ratios and alphas for in- and out-of-sample results. Meanwhile, the first P-Tree is parsimonious, requiring fewer assets than alternative approaches to achieve comparable Sharpe ratios; for example, the first P-Tree only needs 10 assets to achieve an OOS Sharpe ratio of 5.61, whereas the alternative bivariate-sorted portfolios on true characteristics require 25 assets to get comparable Sharpe ratios. Further, the test assets of the first P-Tree have larger OOS alphas than other benchmark test assets.

The boosted P-Trees are denoted as ``P-Tree1-5'' to ``P-Tree1-20.'' Although the in-sample Sharpe ratio increases when adding boosted factors, the alphas are lower from the first to the 20-th P-Tree. Therefore, P-Trees capture the important information in early boosting steps, whereas the subsequent P-Trees provide declining information. We observe lower Sharpe ratios and higher alphas for OOS results for all P-Tree specifications. Meanwhile, ``P-Tree1'' provides the largest OOS Sharpe ratios and alphas. This finding confirms the early P-Trees capture more information, given such a simple return-generating process involving only three true characteristics as in (\ref{eqn:dgp_return_20231012}).

In conclusion, the P-Tree test assets outperform most pre-specified sorted portfolio specifications in risk-adjusted or model-based investment criteria, such as Sharpe ratios and alphas. The outperformance is more pronounced when the predictive signal $\kappa$ is larger. In this interactive and nonlinear return-generating scenario \eqref{eqn:dgp_return_20231012}, P-Tree performances are competitive with all the conventional sorted portfolios.

\paragraph{Limits to learning.}
A gap between in- and out-of-sample performance is frequently observed in financial machine learning applications. As \cite{didisheim2024complexity} suggests, this gap comprises two main components: \textit{Overfit} and \textit{Limits to Learning}.
\vspace{-0.2cm}
\begin{equation}
    \label{eqn:in-out-gap}
    \begin{aligned}
        \text { Gap }= & \text { In-sample Performance }- \text { Out-of-sample Performance } \\
        = & (\text { In-sample Performance }- \text { True Predictability }) \\
          & +(\text { True Predictability } - \text { Out-of-sample Performance }) \\
        = & \text { Overfitting }+ \text { Limits to Learning },
    \end{aligned}
\end{equation}
where \textit{True Predictability} is a pivotal element in this decomposition. However, we find it difficult to observe or estimate the True Predictability in real data because we are agnostic to the true return-generating process. To illustrate this concept, we present a simulation study examining the characteristics of Panel Tree models under a specific return-generating process \eqref{eqn:dgp_return_20231012}.
True Predictability in this context is defined as the OOS Sharpe ratio of a P-Tree model using only the oracle characteristics.

Table \ref{tab:simu_gap} details the simulation results for true predictability, overfit, and limits to learning, measured by the Sharpe ratios. First, we confirm that true predictability exceeds OOS predictability and falls below in-sample predictability.
Additionally, overfitting tends to increase with the number of boosted P-Trees, suggesting that P-Tree models with more boosting steps are more prone to overfitting the in-sample data. Similarly, we observe an increase in limits to learning with more boosting, corroborated by the lower OOS Sharpe ratios for multiple boosted P-Tree factors compared to a single factor, as reported in Table \ref{tab:simu_20231012}. Overall, these findings support the statement in \cite{didisheim2024complexity} that overfitting and limits to learning pose greater challenges as the number of model parameters increases, particularly when the number of observations remains fixed and relatively small.

\begin{table}[ht!]
    \caption{\bf Gap between In- and Out-of-Sample Sharpe Ratio}
    \label{tab:simu_gap}%
    {\footnotesize
    The table reports the true predictability (Panel A), overfitting (Panel B), and limits to learning (Panel C) in terms of the Sharpe ratios in Table \ref{tab:simu_20231012}.
    Specifically, we decompose the gap between in- and out-of-sample Sharpe ratio into two parts: overfitting and limits to learning, see Eq. \eqref{eqn:in-out-gap}.
    }
    
    {\footnotesize

    \begin{center}
    
    \begin{tabular}{l cccccc}
    
    \toprule

    &       & $\kappa=0.5$ 
    &       & $\kappa=1.0$ 
    &       & $\kappa=2.0$ \\

    \hline
    \\
    \multicolumn{7}{c}{\underline{Panel A: True Predictability}} \\
    \\
    
    P-Tree1 &       & 2.87  &       & 5.73  &       & 11.62 \\
    P-Tree1-5 &       & 3.05  &       & 6.13  &       & 12.33 \\
    P-Tree1-10 &       & 3.05  &       & 6.18  &       & 12.40 \\
    P-Tree1-15 &       & 3.05  &       & 6.17  &       & 12.37 \\
    P-Tree1-20 &       & 3.03  &       & 6.15  &       & 12.34 \\

    \hline
    \\
    \multicolumn{7}{c}{\underline{Panel B: Overfitting}} \\
    \\
    
    P-Tree1 &       & 0.44  &       & 0.41  &       & 0.38 \\
    P-Tree1-5 &       & 1.43  &       & 1.46  &       & 1.91 \\
    P-Tree1-10 &       & 3.08  &       & 3.44  &       & 5.20 \\
    P-Tree1-15 &       & 4.80  &       & 5.79  &       & 9.01 \\
    P-Tree1-20 &       & 6.81  &       & 8.15  &       & 13.35 \\

    \hline
    \\
    \multicolumn{7}{c}{\underline{Panel C: Limits to Learning}} \\
    \\
    
    P-Tree1 &       & 0.21  &       & 0.13  &       & 0.24 \\
    P-Tree1-5 &       & 0.71  &       & 0.84  &       & 1.53 \\
    P-Tree1-10 &       & 0.96  &       & 1.24  &       & 2.21 \\
    P-Tree1-15 &       & 1.09  &       & 1.43  &       & 2.50 \\
    P-Tree1-20 &       & 1.13  &       & 1.53  &       & 2.73 \\
    
    \bottomrule

    \end{tabular}

    \end{center}
    
    }
\vspace{-0.8cm}    
\end{table}%

\paragraph{P-Tree with incomplete information.} The literature documents hundreds of characteristics, factors, and anomalies \citep{cochrane2011presidential, green2017characteristics, hou2020replicating, feng2020taming}. 
We include all characteristics as an attempt to span the efficient frontier fully. However, we still do not know whether all the characteristics are helpful. Also, we are agnostic if more valuable characteristics are still not found.

To this end, we investigate the performance of P-Tree without observing the full set of true characteristics, given the return-generating process (\ref{eqn:dgp_return_20231012}).
Specifically, we ignore \texttt{MOM12M} and re-train P-Trees. Table \ref{tab:simu_incomplete_20231012} reports the simulation results. Once we ignore one of the true characteristics, we find a considerable decline in Sharpe ratios and alphas for both in- and out-of-sample analysis. Also, the alphas of P-Tree test assets are smaller, ignoring \texttt{MOM12M}. Therefore, P-Tree would be inefficient if it were missing some key characteristics. In empirical applications, we are agnostic regarding the return-generating process of the real economy. Therefore, we cover many characteristics to relieve the concern about missing key ones.

\begin{table}[h!]
    \caption{\bf Simulation Performance for P-Tree Omitting a Characteristic \texttt{MOM12M}}
    \label{tab:simu_incomplete_20231012}%
    {\footnotesize
    
    This table shows the Sharpe ratios and alphas of P-Tree test assets omitting a true characteristic \texttt{MOM12M}.
    Table format follows Table \ref{tab:simu_20231012}.

    \begin{center}
    
    \begin{tabular}{l cccccccccccc}

    \toprule
    &       & \multicolumn{3}{c}{$\kappa=0.5$} 
    &       & \multicolumn{3}{c}{$\kappa=1.0$} 
    &       & \multicolumn{3}{c}{$\kappa=2.0$} \\
    \cline{3-5} \cline{7-9} \cline{11-13}

    &       & SR & $\overline{|\alpha|}$ &  $\sqrt{\overline{\alpha^2}}$ 
    &       & SR & $\overline{|\alpha|}$ &  $\sqrt{\overline{\alpha^2}}$ 
    &       & SR & $\overline{|\alpha|}$ &  $\sqrt{\overline{\alpha^2}}$ 
    \\
    \hline

    \\
    \multicolumn{13}{c}{\underline{Panel A: In-Sample Results}} \\
    \\

    P-Tree1 &       & 2.73  & 0.32  & 0.41  &       & 4.86  & 0.60  & 0.77  &       & 9.42  & 1.30  & 1.66 \\
    P-Tree1-5 &       & 4.01  & 0.18  & 0.26  &       & 6.18  & 0.25  & 0.40  &       & 11.52 & 0.47  & 0.81 \\
    P-Tree1-10 &       & 5.60  & 0.15  & 0.22  &       & 8.17  & 0.20  & 0.32  &       & 14.31 & 0.35  & 0.62 \\
    P-Tree1-15 &       & 7.23  & 0.14  & 0.20  &       & 10.24 & 0.19  & 0.29  &       & 17.07 & 0.32  & 0.54 \\
    P-Tree1-20 &       & 9.09  & 0.14  & 0.19  &       & 12.52 & 0.19  & 0.28  &       & 20.50 & 0.30  & 0.50 \\

    \hline
    \\
    \multicolumn{13}{c}{\underline{Panel B: Out-of-Sample Results}} \\
    \\
    P-Tree1 &       & 1.91  & 0.59  & 0.68  &       & 4.23  & 0.76  & 0.94  &       & 8.91  & 1.36  & 1.72 \\
    P-Tree1-5 &       & 1.54  & 0.59  & 0.62  &       & 3.79  & 0.62  & 0.70  &       & 8.28  & 0.77  & 0.99 \\
    P-Tree1-10 &       & 1.39  & 0.60  & 0.62  &       & 3.48  & 0.62  & 0.67  &       & 7.81  & 0.68  & 0.85 \\
    P-Tree1-15 &       & 1.34  & 0.60  & 0.62  &       & 3.31  & 0.61  & 0.66  &       & 7.56  & 0.67  & 0.81 \\
    P-Tree1-20 &       & 1.29  & 0.60  & 0.62  &       & 3.24  & 0.61  & 0.66  &       & 7.36  & 0.66  & 0.78 \\

    \bottomrule

    \end{tabular}%

    \end{center}
    
    }
\vspace{-0.8cm}
\end{table}%

\paragraph{P-Tree with true characteristics.}
Even without pre-specifying the true characteristics, P-Tree can produce informative test assets to achieve high Sharpe ratios and large pricing errors against the market factor. 
Thus, P-Tree can be used to select the true characteristics.
Table \ref{tab:simu_selection_probability_20231012} shows the probability of selecting each characteristic in P-Tree splits. We only report for the first P-Tree over 10 simulation repetitions for simplicity. We find the true characteristics, \texttt{ME}, \texttt{BM}, and \texttt{MOM12M}, are selected with high probabilities in the top three, top five, and all nine splits.

When $\kappa$ is large, the probability of selecting the true characteristics is higher. For low signal-to-noise setting $\kappa=0.5$, the noisy characteristics are selected in top splits; for example, \texttt{NI} and \texttt{BASPREAD} are selected with non-zero probabilities in the top five splits. However, for high signal-to-noise setting $\kappa=2$, only the true characteristics are selected in the top five splits, over 10 simulation repetitions.
Therefore, we confirm that the superb performance of P-Trees originates from correctly selecting the true characteristics in the return-generating process (\ref{eqn:dgp_return_20231012}).

\begin{table}[h!]
    \caption{\bf Characteristics Selection Probability for the First P-Tree}
    \label{tab:simu_selection_probability_20231012}

    {\footnotesize

    This table shows the selected characteristics of the first P-Tree with its selection probability over 10 repetitions. Each P-Tree contains nine splits and, thus, ten leaves.
    The ``Top3'' columns specify that we only count the first three splits of the P-Tree model, and similarly, we defined ``Top5'' and ``Top9.'' 
    The rows only show the five most frequently selected characteristics.
    For example, in Panel A, the characteristic \texttt{BM} is the most frequently selected, with a 40\% probability, in the top three splits.
    Panels A, B, and C report for $\kappa = 0.5, 1, 2$.
    The left panel reports for P-Trees with complete characteristics information as in Table \ref{tab:simu_20231012}, whereas the right panel is for P-Trees omitting \texttt{MOM12M} as in Table \ref{tab:simu_incomplete_20231012}.
    \\

        \resizebox{\textwidth}{!}{
        
            \begin{tabular}{l p{1cm}p{1.5cm}p{1.5cm}p{1.5cm}p{1.5cm} c p{1cm}p{1cm}p{1.75cm}p{1.25cm}p{1.5cm}}
            \toprule

                & \multicolumn{5}{c}{Complete Characteristics Set} &       & \multicolumn{5}{c}{Remove \texttt{MOM12M}}  \\
                
                \cline{2-6} \cline{8-12}
                
                & 1     & 2     & 3     & 4     & 5     &       & 1     & 2     & 3     & 4     & 5 \\
                
                \hline
                
                \\
                \multicolumn{12}{c}{\underline{Panel A: $\kappa=0.5$}} \\
                \\
                            
                Top3  & BM    & MOM12M & ME    &       &       &       & ME    & BM    &       &       &  \\
                      & 0.40  & 0.33  & 0.27  &       &       &       & 0.63  & 0.37  &       &       &  \\
                    Top5  & BM    & ME    & MOM12M & NI    & BASPREAD &       & ME    & BM    & BASPREAD & HIRE    & SEAS1A \\
                      & 0.40  & 0.34  & 0.22  & 0.02  & 0.02  &       & 0.46  & 0.38  & 0.02  & 0.02  & 0.02 \\
                Top9  & BM    & ME    & MOM12M & ALM   & CFP   &       & ME    & BM    & ILL & RNA   & CASHDEBT \\
                      & 0.33  & 0.26  & 0.12  & 0.02  & 0.02  &       & 0.29  & 0.26  & 0.03  & 0.02  & 0.02 \\

                \hline
                
                \\
                \multicolumn{12}{c}{\underline{Panel B: $\kappa=1$}} \\
                \\

                Top3  & BM    & MOM12M & ME    &       &       &       & ME    & BM    &       &       &  \\
                      & 0.37  & 0.33  & 0.30  &       &       &       & 0.60  & 0.40  &       &       &  \\
                Top5  & BM    & ME    & MOM12M &       &       &       & BM    & ME    & DEPR    & TURN & SVAR \\
                      & 0.44  & 0.32  & 0.24  &       &       &       & 0.48  & 0.44  & 0.02  & 0.02  & 0.02 \\
                Top9  & BM    & ME    & MOM12M & CHCSHO & RE    &       & ME    & BM    & TURN & DEPR    & ILL \\
                      & 0.36  & 0.29  & 0.16  & 0.03  & 0.02  &       & 0.40  & 0.34  & 0.03  & 0.02  & 0.02 \\

                \hline
                
                \\
                \multicolumn{12}{c}{\underline{Panel C: $\kappa=2$}} \\
                \\
                            
                Top3  & BM    & MOM12M & ME    &       &       &       & ME    & BM    &       &       &  \\
                      & 0.37  & 0.33  & 0.30  &       &       &       & 0.67  & 0.33  &       &       &  \\
                Top5  & BM    & ME    & MOM12M &       &       &       & BM    & ME    &       &       &  \\
                      & 0.48  & 0.28  & 0.24  &       &       &       & 0.60  & 0.40  &       &       &  \\
                Top9  & BM    & ME    & MOM12M & MOM36M & CFP   &       & ME    & BM    & CFP & SP  & TURN \\
                      & 0.43  & 0.27  & 0.19  & 0.02  & 0.01  &       & 0.47  & 0.42  & 0.02  & 0.01  & 0.01 \\

            \bottomrule
                
            \end{tabular}%
    
        }

    }
\vspace{-0.3cm}    
\end{table}

Further, we investigate the characteristics selection results when omitting an important characteristic \texttt{MOM12M}, the same as the treatment in Table \ref{tab:simu_incomplete_20231012}. We find \texttt{BM} and \texttt{ME} are selected with an even higher probability, once we omit \texttt{MOM12M}. More importantly, we find noisy characteristics are selected more frequently. However, the noisy characteristics cannot help P-Tree fully span the efficient frontier, as shown in Table \ref{tab:simu_incomplete_20231012}. This phenomenon alerts us that we must feed all true characteristics to P-Tree.

\clearpage
\section{Benchmark-Adjusted P-Trees} 
\label{sec:market_benchmark}


As described in section \ref{sec:boosting}, P-Tree boosting is a flexible approach to include any benchmark factor(w), such as CAPM, Fama-French three factors, and PCA factors. Here, we demonstrate an example of fitting boosted P-Trees with the market factor as a benchmark factor model. Specifically, we define $\F = [\text{mkt}_t, f_{1,t}, \cdots, f_{K,t}]$ and apply to the split criteria \eqref{eqn:criteria}.

The initial P-Tree remains unchanged in the empirical findings, even when the market factor is included as a benchmark, suggesting the information usefulness of the cross-sectional characteristics beyond the market factor. Therefore, the P-Tree diagram remains identical to Figure \ref{fig:tree_a}. 
However, the subsequent P-Trees change significantly, and the empirical performances differ from those in section \ref{sec:empirical_example}. The following tables and figures present the empirical results for the P-Tree test assets and factors with the market factor benchmark.

\begin{figure}[!h]
    \caption{\bf Market Factor Benchmark: Characterizing the Efficient Frontier}
    \label{fig:ef_b}

    {\footnotesize
    This figure presents complementary results to Figure \ref{fig:ef_a} for considering the market factor benchmark while growing P-Trees.
    }
    
    \begin{center}
        \includegraphics[width=0.95\textwidth]{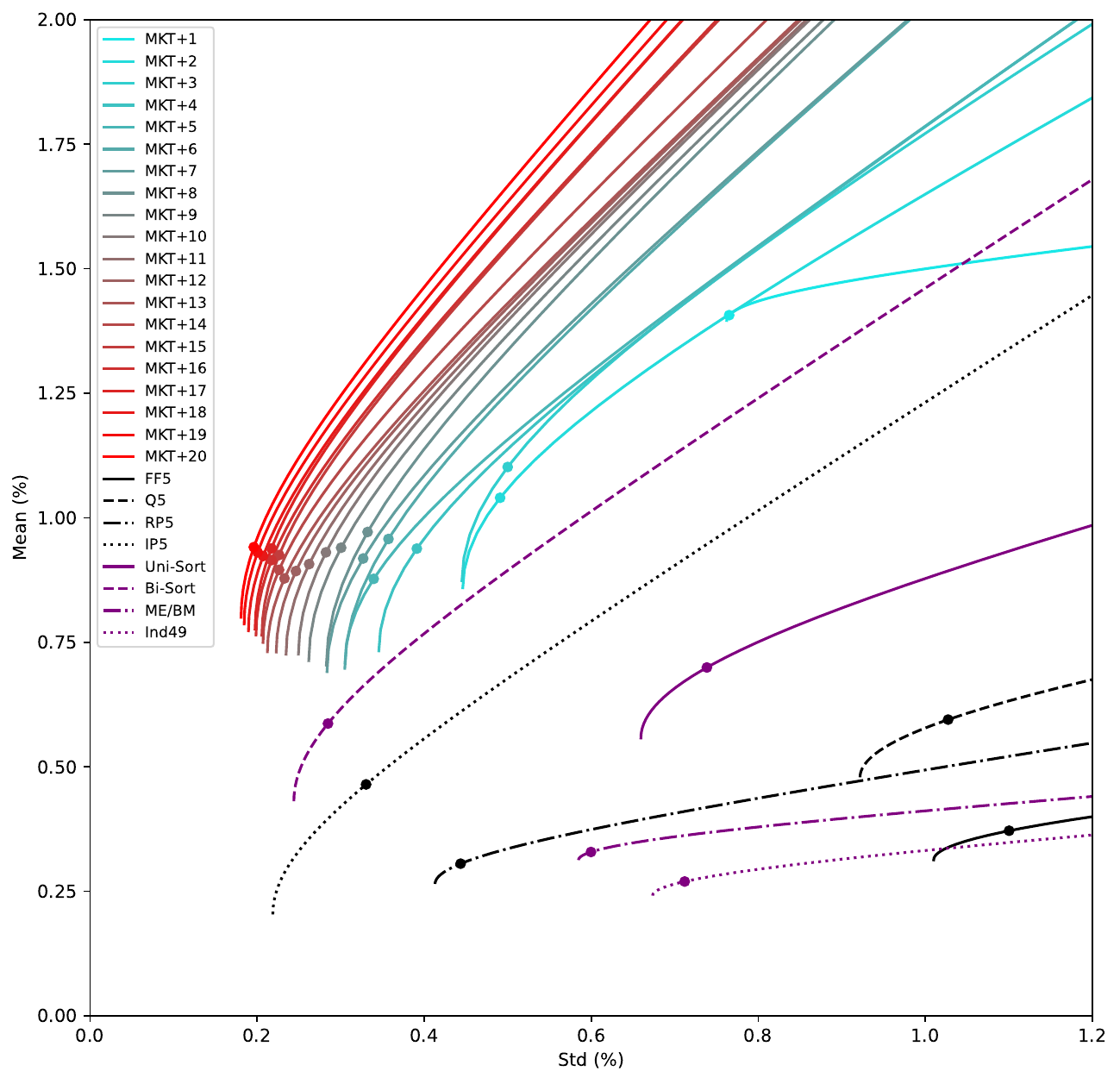} 
    \end{center}
    
\end{figure}

Figure \ref{fig:ef_b} depicts the efficient frontiers spanned by P-Tree factors and the market factor. We find similar patterns as our main results without a market benchmark in Figure \ref{fig:ef_a}. The P-Tree frontiers move to the top-left corner of the mean-variance diagram, as more boosted P-Trees are generated. However, the market factor only takes a small proportion of the tangency portfolio, and the frontier is flat for the first P-Tree and the market factor.

Table \ref{tab:test_asset_b} shows the asset pricing performance of P-Tree test assets. Consistent with our main results in Table \ref{tab:test_asset_a}, we find the test assets of the first P-Tree have larger GRS statistics, higher average alphas, and higher proportions of significant alphas than the test assets generated by the follow-up P-Trees. Additionally, the GRS and PY tests always reject the null hypothesis, except for P-Tree11-15 in subsample 2001-2020. In general, the test assets of P-Trees with a market factor benchmark are challenging to be priced by FF5.

Table \ref{tab:factor_dim_b} examines the performance of each P-Tree factor. Consistent with our main results in Table \ref{tab:factor_dim_a}, the boosted P-Tree factors cannot be explained by the previous P-Tree factors, as indicated by the low $p$-values. In contrast to the results in Table \ref{tab:factor_dim_a}, where the first factor could not be tested, we test the first P-Tree factor against the market benchmark. As expected, the first P-Tree factor cannot be explained by the market factor.

Table \ref{tab:factors_b} shows the investment performances. Including the market factor improves the risk-adjusted investment performance slightly. The investment Sharpe ratios are over 3 for the OOS periods 2001-2020 and 1981-2020. The observable and latent factor models listed cannot explain our P-Tree factors with large unexplained alphas. In conclusion, this section demonstrates the flexibility of boosting in P-Trees: one can add any given factor model as a control when building P-Tree test assets and factors.

\begin{table}[!h]
  
  \caption{\bf Market Factor Benchmark: Test Assets Generated by P-Tree
  }
  \label{tab:test_asset_b}%

  {\footnotesize

  This table presents complementary results to Table \ref{tab:test_asset_a} for considering the market factor benchmark while growing P-Trees. 

    \begin{center}
        
    \begin{tabular}{l cccc cccccc}
    
    \toprule
    \\
    & N & \multicolumn{1}{l}{GRS} & \multicolumn{1}{l}{$p$-GRS} 
    & \multicolumn{1}{l}{$p$-PY}
    & \multicolumn{1}{l}{ $\overline{|\alpha|}$} 
    & \multicolumn{1}{l}{ $\sqrt{\overline{\alpha^2}}$}
    & \multicolumn{1}{l}{$\overline{R^2}$}
    & $\%\alpha_{10\%}$ & $\%\alpha_{5\%}$ & $\%\alpha_{1\%}$ \\

    \hline
    
    \\
    \multicolumn{11}{c}{\underline{Panel A: 40 Years (1981-2020)}} \\
    \\
              
    P-Tree1 & 10    & 141.27 & 0.00  & 0.00  & 0.92  & 1.11  & 75    & 100   & 90    & 80 \\
    P-Tree1-5 & 50    & 57.09 & 0.00  & 0.00  & 0.39  & 0.59  & 78    & 64    & 52    & 46 \\
    P-Tree6-10 & 50    & 4.99  & 0.00  & 0.00  & 0.24  & 0.31  & 79    & 48    & 38    & 24 \\
    P-Tree11-15 & 50    & 6.82  & 0.00  & 0.00  & 0.39  & 0.52  & 71    & 50    & 40    & 30 \\
    P-Tree16-20 & 50    & 3.74  & 0.00  & 0.00  & 0.27  & 0.36  & 72    & 48    & 40    & 32 \\
    P-Tree1-20 & 200   & 39.88 & 0.00  & 0.00  & 0.32  & 0.46  & 75    & 53    & 43    & 33 \\
    
    \\
    \hline
    
    \\
    \multicolumn{11}{c}{\underline{Panel B: 20 Years (1981-2000)}} \\
    \\

    P-Tree1 & 10    & 84.36 & 0.00  & 0.00  & 1.58  & 1.95  & 69.62 & 90    & 80    & 80 \\
    P-Tree1-5 & 50    & 50.84 & 0.00  & 0.00  & 0.79  & 1.26  & 76.16 & 62    & 60    & 52 \\
    PTree6-10 & 50    & 7.27  & 0.00  & 0.00  & 0.58  & 0.87  & 74.90 & 56    & 44    & 38 \\
    P-Tree11-15 & 50    & 6.39  & 0.00  & 0.00  & 0.55  & 0.82  & 76.10 & 66    & 60    & 42 \\
    P-Tree16-20 & 50    & 8.42  & 0.00  & 0.00  & 0.52  & 0.74  & 76.00 & 62    & 54    & 50 \\
    P-Tree1-20 & 200   & 112.90 & 0.00  & 0.00  & 0.61  & 0.95  & 75.79 & 62    & 55    & 46 \\

    \\
    \hline
    
    \\
    \multicolumn{11}{c}{\underline{Panel C: 20 Years (2001-2020)}} \\
    \\

    P-Tree1 & 10    & 56.76 & 0.00  & 0.00  & 1.09  & 1.35  & 68.45 & 90    & 90    & 90 \\
    P-Tree1-5 & 50    & 30.35 & 0.00  & 0.00  & 0.43  & 0.68  & 76.44 & 52    & 38    & 24 \\
    PTree6-10 & 50    & 5.17  & 0.00  & 0.00  & 0.29  & 0.37  & 74.78 & 34    & 28    & 14 \\
    P-Tree11-15 & 50    & 2.20  & 0.00  & 0.00  & 0.27  & 0.35  & 75.22 & 30    & 22    & 10 \\
    P-Tree16-20 & 50    & 2.52  & 0.00  & 0.00  & 0.31  & 0.40  & 75.82 & 42    & 28    & 10 \\
    P-Tree1-20 & 200   & 83.91 & 0.00  & 0.00  & 0.33  & 0.47  & 75.56 & 40    & 29    & 14 \\

    \bottomrule
    \end{tabular}

    \end{center}
  }
\end{table}%

\clearpage

\begin{table}[ht!]
  
  \caption{\bf Market Factor Benchmark: Testing the Boosted P-Tree Growth}
  \label{tab:factor_dim_b}%
  
  {\footnotesize
  This table presents complementary results to Table \ref{tab:factor_dim_a} for considering the market factor benchmark while growing P-Trees. 

  \vspace{-0.3cm}
  
  \begin{center}
    \resizebox{\textwidth}{!}{
      \begin{tabular}{l ccccccccccccccc}

      \toprule
        
        &
        & \multicolumn{2}{c}{Sharpe Ratio} &
        & \multicolumn{2}{c}{CAPM Test} &
        & \multicolumn{2}{c}{FF5 Test}   &
        & \multicolumn{3}{c}{Expanding Factors Test}   &
        & \multicolumn{1}{c}{BS Test} \\
        
        \cline{3-4} \cline{6-7} \cline{9-10} \cline{12-14} \cline{16-16}

        &
        & Single    & Cumu. &
        & $\alpha$ (\%) & $t$-stat & 
        & $\alpha$ (\%) & $t$-stat & 
        & $\alpha$ (\%) & $t$-stat & $R^2$ & 
        & $p$-value \\
        \hline
        
        \\
        \multicolumn{16}{c}{\underline{Panel A: 40 Years (1981-2020)}}\\
        \\
            
        1     &       & 6.37  & 6.37  &       & 1.39  & 35.36 &       & 1.37  & 35.81 &       & 1.39  & 35.36 & 0.01  &       & 0.00 \\
        2     &       & 2.24  & 7.33  &       & 0.40  & 12.93 &       & 0.37  & 11.47 &       & 0.66  & 8.90  & 0.13  &       & 0.00 \\
        3     &       & 0.65  & 7.61  &       & 0.12  & 1.75  &       & -0.03 & -0.72 &       & -0.62 & -4.76 & 0.66  &       & 0.00 \\
        4     &       & 1.57  & 8.29  &       & 0.27  & 9.09  &       & 0.26  & 8.20  &       & 0.58  & 8.39  & 0.25  &       & 0.00 \\
        5     &       & 1.63  & 8.92  &       & 0.46  & 9.61  &       & 0.42  & 8.98  &       & 0.97  & 7.99  & 0.17  &       & 0.00 \\
        6     &       & 0.72  & 9.24  &       & 0.10  & 2.70  &       & 0.05  & 1.61  &       & -0.47 & -5.54 & 0.70  &       & 0.00 \\
        7     &       & 1.40  & 9.66  &       & 0.36  & 7.98  &       & 0.29  & 6.16  &       & 0.74  & 6.08  & 0.29  &       & 0.00 \\
        8     &       & 0.83  & 10.04 &       & 0.17  & 3.52  &       & 0.09  & 2.04  &       & -0.66 & -5.56 & 0.55  &       & 0.00 \\
        9     &       & 1.39  & 10.72 &       & 0.29  & 7.24  &       & 0.25  & 6.85  &       & 0.77  & 7.83  & 0.36  &       & 0.00 \\
        10    &       & 1.37  & 11.27 &       & 0.36  & 8.10  &       & 0.27  & 6.64  &       & 0.93  & 6.31  & 0.31  &       & 0.00 \\
        11    &       & 1.22  & 11.82 &       & 0.37  & 6.72  &       & 0.40  & 7.21  &       & 1.14  & 6.48  & 0.31  &       & 0.00 \\
        12    &       & 1.24  & 12.36 &       & 0.36  & 6.45  &       & 0.28  & 4.28  &       & 1.09  & 6.06  & 0.35  &       & 0.00 \\
        13    &       & 1.27  & 12.87 &       & 0.40  & 6.38  &       & 0.25  & 4.29  &       & -1.02 & -5.92 & 0.46  &       & 0.00 \\
        14    &       & 1.64  & 13.47 &       & 0.53  & 9.27  &       & 0.48  & 8.63  &       & 1.12  & 6.99  & 0.44  &       & 0.00 \\
        15    &       & 0.69  & 13.88 &       & 0.10  & 2.36  &       & 0.05  & 1.42  &       & -0.68 & -6.16 & 0.73  &       & 0.00 \\
        16    &       & 1.16  & 14.31 &       & 0.41  & 6.13  &       & 0.29  & 5.47  &       & -1.00 & -5.07 & 0.58  &       & 0.00 \\
        17    &       & 0.80  & 14.67 &       & 0.15  & 3.75  &       & 0.11  & 2.58  &       & -0.79 & -5.01 & 0.55  &       & 0.00 \\
        18    &       & 1.69  & 15.09 &       & 0.34  & 10.25 &       & 0.28  & 9.78  &       & 0.71  & 5.09  & 0.33  &       & 0.00 \\
        19    &       & 1.37  & 15.59 &       & 0.33  & 7.60  &       & 0.29  & 6.86  &       & -1.00 & -7.11 & 0.40  &       & 0.00 \\
        20    &       & 1.08  & 16.09 &       & 0.23  & 5.31  &       & 0.15  & 4.00  &       & -0.87 & -5.60 & 0.52  &       & 0.00 \\

      \bottomrule
      
      \end{tabular}%
    } 
    \end{center}
    } 
\end{table}%

\clearpage

\begin{table}[ht!]
  
  \caption*{\bf Table Continued: Market Factor Benchmark: Testing the Boosted P-Tree Growth}

  \vspace{-0.3cm}
  
  {\footnotesize

  \begin{center}
    \resizebox{\textwidth}{!}{
      \begin{tabular}{l ccccccccccccccc}
      
      \toprule
        
        &
        & \multicolumn{2}{c}{Sharpe Ratio} &
        & \multicolumn{2}{c}{CAPM Test} &
        & \multicolumn{2}{c}{FF5 Test}   &
        & \multicolumn{3}{c}{Expanding Factors Test}   &
        & \multicolumn{1}{c}{BS Test} \\
        
        \cline{3-4} \cline{6-7} \cline{9-10} \cline{12-14} \cline{16-16}

        &
        & Single    & Cumu. &
        & $\alpha$ (\%) & $t$-stat & 
        & $\alpha$ (\%) & $t$-stat & 
        & $\alpha$ (\%) & $t$-stat & $R^2$ & 
        & $p$-value \\
        \hline
        
        \\
        \multicolumn{16}{c}{\underline{Panel B: 20 Years (1981-2000)}}\\
        \\
        
        1     &       & 7.12  & 7.12  &       & 1.86  & 29.62 &       & 1.78  & 30.22 &       & 1.86  & 29.62 & 0.01  &       & 0.00 \\
        2     &       & 1.09  & 8.03  &       & 0.25  & 3.66  &       & 0.11  & 1.91  &       & -0.83 & -4.81 & 0.50  &       & 0.00 \\
        3     &       & 2.26  & 9.52  &       & 0.39  & 9.72  &       & 0.34  & 8.48  &       & 0.86  & 8.02  & 0.20  &       & 0.00 \\
        4     &       & 6.48  & 11.41 &       & 1.80  & 26.96 &       & 1.76  & 27.06 &       & 1.75  & 10.55 & 0.03  &       & 0.00 \\
        5     &       & 1.59  & 12.71 &       & 0.33  & 6.50  &       & 0.28  & 5.77  &       & 1.08  & 6.54  & 0.37  &       & 0.00 \\
        6     &       & 1.65  & 13.70 &       & 0.37  & 5.87  &       & 0.26  & 4.57  &       & -1.04 & -5.43 & 0.41  &       & 0.00 \\
        7     &       & 1.72  & 14.97 &       & 0.33  & 6.99  &       & 0.32  & 6.17  &       & 1.19  & 5.87  & 0.28  &       & 0.00 \\
        8     &       & 1.52  & 15.78 &       & 0.39  & 6.10  &       & 0.31  & 4.93  &       & -1.29 & -5.06 & 0.36  &       & 0.00 \\
        9     &       & 3.02  & 17.89 &       & 1.03  & 14.00 &       & 0.93  & 13.08 &       & 2.69  & 8.42  & 0.26  &       & 0.00 \\
        10    &       & 1.93  & 19.18 &       & 0.83  & 8.12  &       & 0.67  & 7.14  &       & -2.41 & -6.25 & 0.52  &       & 0.00 \\
        11    &       & 1.34  & 20.74 &       & 0.38  & 4.93  &       & 0.19  & 2.76  &       & -2.15 & -7.68 & 0.47  &       & 0.00 \\
        12    &       & 2.22  & 22.89 &       & 0.36  & 9.38  &       & 0.31  & 8.03  &       & 1.45  & 7.23  & 0.39  &       & 0.00 \\
        13    &       & 1.51  & 24.13 &       & 0.35  & 4.88  &       & 0.23  & 3.69  &       & 1.80  & 5.96  & 0.43  &       & 0.00 \\
        14    &       & 2.11  & 25.70 &       & 0.58  & 7.80  &       & 0.48  & 7.57  &       & -2.43 & -5.92 & 0.40  &       & 0.00 \\
        15    &       & 2.39  & 28.37 &       & 0.87  & 10.14 &       & 0.76  & 8.70  &       & 3.69  & 7.68  & 0.53  &       & 0.00 \\
        16    &       & 2.53  & 30.36 &       & 0.53  & 9.58  &       & 0.44  & 7.91  &       & 2.41  & 7.12  & 0.31  &       & 0.00 \\
        17    &       & 1.50  & 32.40 &       & 0.30  & 5.66  &       & 0.31  & 5.08  &       & 2.57  & 6.94  & 0.39  &       & 0.00 \\
        18    &       & 1.63  & 33.82 &       & 0.37  & 5.95  &       & 0.28  & 4.99  &       & -2.73 & -5.64 & 0.39  &       & 0.00 \\
        19    &       & 1.84  & 35.69 &       & 0.47  & 7.04  &       & 0.37  & 5.93  &       & 3.52  & 7.34  & 0.30  &       & 0.00 \\
        20    &       & 2.70  & 37.86 &       & 0.89  & 9.85  &       & 0.79  & 9.06  &       & -4.67 & -6.93 & 0.44  &       & 0.00 \\
        
        \\
        \multicolumn{16}{c}{\underline{Panel C: 20 Years (2001-2020)}}\\
        \\

        1     &       & 5.82  & 5.82  &       & 1.51  & 24.22 &       & 1.47  & 24.26 &       & 1.51  & 24.22 & 0.01  &       & 0.00 \\
        2     &       & 2.14  & 6.72  &       & 0.58  & 8.74  &       & 0.52  & 8.26  &       & 0.91  & 6.54  & 0.12  &       & 0.00 \\
        3     &       & 0.91  & 7.62  &       & 0.18  & 3.28  &       & 0.09  & 2.01  &       & -0.71 & -7.09 & 0.58  &       & 0.00 \\
        4     &       & 1.35  & 8.61  &       & 0.29  & 6.18  &       & 0.31  & 6.04  &       & 0.88  & 8.04  & 0.30  &       & 0.00 \\
        5     &       & 1.08  & 9.32  &       & 0.31  & 4.14  &       & 0.26  & 3.88  &       & -0.99 & -6.40 & 0.48  &       & 0.00 \\
        6     &       & 2.63  & 10.36 &       & 0.60  & 12.15 &       & 0.56  & 11.11 &       & 1.03  & 7.34  & 0.12  &       & 0.00 \\
        7     &       & 1.39  & 11.29 &       & 0.29  & 5.53  &       & 0.26  & 5.16  &       & 0.96  & 7.15  & 0.32  &       & 0.00 \\
        8     &       & 1.60  & 11.95 &       & 0.40  & 5.76  &       & 0.30  & 5.47  &       & -0.86 & -5.58 & 0.45  &       & 0.00 \\
        9     &       & 1.04  & 12.73 &       & 0.21  & 4.53  &       & 0.18  & 3.83  &       & -0.95 & -7.70 & 0.45  &       & 0.00 \\
        10    &       & 1.64  & 14.33 &       & 0.30  & 5.96  &       & 0.25  & 5.31  &       & 1.13  & 7.94  & 0.45  &       & 0.00 \\
        11    &       & 1.69  & 15.67 &       & 0.36  & 6.73  &       & 0.30  & 6.20  &       & 1.40  & 6.58  & 0.34  &       & 0.00 \\
        12    &       & 1.58  & 16.71 &       & 0.60  & 6.08  &       & 0.51  & 5.36  &       & 2.42  & 5.64  & 0.25  &       & 0.00 \\
        13    &       & 1.48  & 17.93 &       & 0.41  & 5.29  &       & 0.29  & 4.47  &       & -1.77 & -5.51 & 0.47  &       & 0.00 \\
        14    &       & 1.39  & 19.31 &       & 0.32  & 5.22  &       & 0.26  & 4.62  &       & 1.66  & 6.16  & 0.41  &       & 0.00 \\
        15    &       & 1.48  & 20.45 &       & 0.35  & 5.51  &       & 0.21  & 4.34  &       & -1.38 & -6.21 & 0.58  &       & 0.00 \\
        16    &       & 1.40  & 21.80 &       & 0.34  & 5.84  &       & 0.27  & 4.82  &       & 1.87  & 6.28  & 0.47  &       & 0.00 \\
        17    &       & 1.27  & 23.25 &       & 0.39  & 4.82  &       & 0.25  & 3.46  &       & -2.55 & -7.01 & 0.51  &       & 0.00 \\
        18    &       & 1.44  & 24.15 &       & 0.34  & 5.16  &       & 0.28  & 4.72  &       & -1.61 & -4.91 & 0.48  &       & 0.00 \\
        19    &       & 1.58  & 25.13 &       & 0.33  & 6.59  &       & 0.28  & 6.08  &       & -1.72 & -5.17 & 0.42  &       & 0.00 \\
        20    &       & 1.88  & 26.11 &       & 0.58  & 7.12  &       & 0.50  & 6.53  &       & -2.75 & -4.93 & 0.39  &       & 0.00 \\
        
      \bottomrule
      
      \end{tabular}%
    } 
    \end{center}
    } 
\end{table}%

\clearpage

\begin{table}[ht!]
    
    \caption{\bf Market Factor Benchmark: Factor Investing by Boosted P-Trees}
    \label{tab:factors_b}
    
    {\footnotesize

    This table presents complementary results to Table \ref{tab:factors_a} for considering the market factor benchmark while growing P-Trees. We find all the $\alpha$'s are significant at 1\% confidence level.

    }
    
  {\footnotesize
    \begin{center}

        \begin{tabular}{l cccccc}
        \toprule
              & SR    & $\alpha_{CAPM}$ & $\alpha_{FF5}$ & $\alpha_{Q5}$ & $\alpha_{RP5}$ & $\alpha_{IP5}$ \\
              \hline

              \\
              \multicolumn{7}{c}{\underline{Panel A: 40 Years (1981-2020)}} \\
              \\
              
        P-Tree1    & 6.38  & 1.39***  & 1.37***  & 1.36***  & 1.30***  & 1.12*** \\
        P-Tree1-5  & 8.93  & 0.89***  & 0.87***  & 0.85***  & 0.83***  & 0.75*** \\
        P-Tree1-10 & 11.28 & 0.91***  & 0.90***  & 0.88***  & 0.86***  & 0.83*** \\
        P-Tree1-15 & 13.90 & 0.90***  & 0.89***  & 0.87***  & 0.86***  & 0.84*** \\
        P-Tree1-20 & 16.10 & 0.91***  & 0.91***  & 0.89***  & 0.88***  & 0.86*** \\
    
        \hline
        \\
        
              \multicolumn{7}{c}{\underline{Panel B1: 20 Years In-Sample (1981-2000)}} \\
              \\
        P-Tree1    & 7.13  & 1.85***  & 1.77***  & 1.71***  & 1.68***  & 1.58*** \\
        P-Tree1-5  & 12.74 & 1.54***  & 1.51***  & 1.48***  & 1.46***  & 1.41*** \\
        P-Tree1-10 & 19.22 & 1.50***  & 1.48***  & 1.47***  & 1.44***  & 1.41*** \\
        P-Tree1-15 & 28.43 & 1.42***  & 1.41***  & 1.39***  & 1.38***  & 1.38*** \\
        P-Tree1-20 & 37.94 & 1.34***  & 1.33***  & 1.32***  & 1.31***  & 1.32*** \\
    
              \\
              \multicolumn{7}{c}{\underline{Panel B2: 20 Years Out-of-Sample (2001-2020) }} \\
              \\
        P-Tree1    & 3.24  & 1.34***  & 1.31***  & 1.23***  & 1.17***  & 0.93*** \\
        P-Tree1-5  & 3.41  & 1.02***  & 1.00***  & 0.95***  & 0.89***  & 0.62*** \\
        P-Tree1-10 & 3.21  & 0.94***  & 0.92***  & 0.87***  & 0.82***  & 0.55*** \\
        P-Tree1-15 & 3.12  & 0.89***  & 0.89***  & 0.83***  & 0.81***  & 0.48*** \\
        P-Tree1-20 & 3.13  & 0.83***  & 0.82***  & 0.77***  & 0.75***  & 0.48*** \\
    
        \hline
              \\
              \multicolumn{7}{c}{\underline{Panel C1: 20 Years In-Sample (2001-2020)}} \\
              \\
        P-Tree1    & 5.83  & 1.51***  & 1.47***  & 1.50***  & 1.47***  & 1.69*** \\
        P-Tree1-5  & 9.34  & 1.32***  & 1.31***  & 1.30***  & 1.30***  & 1.32*** \\
        P-Tree1-10 & 14.36 & 1.13***  & 1.12***  & 1.11***  & 1.13***  & 1.09*** \\
        P-Tree1-15 & 20.49 & 1.04***  & 1.03***  & 1.03***  & 1.04***  & 1.00*** \\
        P-Tree1-20 & 26.16 & 1.05***  & 1.04***  & 1.04***  & 1.05***  & 1.05*** \\

              \\
              \multicolumn{7}{c}{\underline{Panel C2: 20 Years Out-of-Sample (1981-2000)}} \\
              \\
        P-Tree1    & 4.35  & 1.50***  & 1.42***  & 1.35***  & 1.58***  & 1.58*** \\
        P-Tree1-5  & 3.88  & 1.20***  & 1.07***  & 0.98***  & 1.25***  & 1.26*** \\
        P-Tree1-10 & 4.29  & 1.03***  & 0.93***  & 0.85***  & 1.11***  & 1.10*** \\
        P-Tree1-15 & 4.04  & 0.91***  & 0.82***  & 0.75***  & 0.99***  & 0.97*** \\
        P-Tree1-20 & 3.91  & 0.92***  & 0.83***  & 0.76***  & 1.00***  & 0.98*** \\
    
        \bottomrule
        \end{tabular}
        
    \end{center}
    }
\end{table}

\clearpage 
\section{Over-parameterized SDF via Random Split.}
\label{sec:random_split_SDF}

\cite{didisheim2024complexity} find that characteristics-managed factors generated by Random Fourier Feature (RFF) can be used to generate the SDF with excellent OOS performances, especially when the number of factors is much larger than the number of time length.
This section provides another specification of the SDF, namely the \textit{random split SDF}.
We continue using the P-Tree framework, which means we check if a leaf is splittable in each step, but the split criterion value is from a random number generator. There is no economic guidance on tree growth.
This section reports the OOS performances of the random split SDF and compares it with the random P-Forest SDF in Section \ref{sec:random_p_forest_complex}, where the only difference is whether there is a split criterion.

\begin{figure}[htbp]

    \caption{\bf{OOS Performance of Random Split SDF, \#Leaf = 10}}
    \label{fig: random_split_N10}
    {\footnotesize
    This figure reports the Sharpe ratio, and pricing error (HJD) of the realized OOS SDF portfolio. 
    Each P-Tree is split with random criteria and has 10 leaf basis portfolios. 
    The SDF is spanned by these characteristic-managed leaf basis portfolios of a large number of independent P-Trees.
    The total number of leaf basis portfolios of all P-Trees is denoted $P$.
    The horizontal axis shows model complexity $c = P/T$, with $c$ ranging from 0.1 to 100 and $T = 240$ months.    
    }
    
    \vspace{-0.2cm}

   \begin{center}

    \begin{subfigure}{0.45\textwidth}
        \begin{center} 
            \includegraphics[height=.95\textwidth]{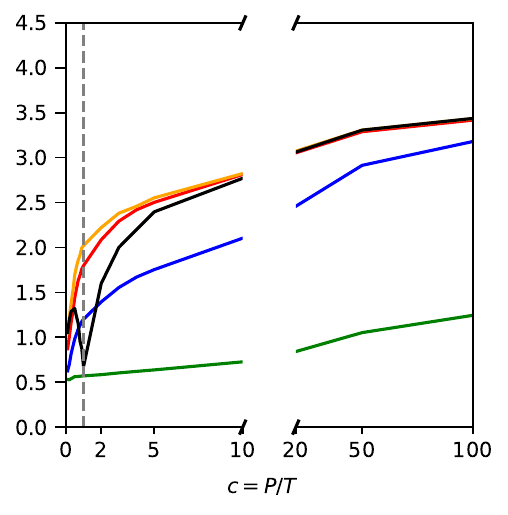}
            \caption{\footnotesize{Sharpe Ratio}}
        \end{center} 
    \end{subfigure}  
    \hfill
    \begin{subfigure}{0.45\textwidth}
        \begin{center} 
            \includegraphics[height=.95\textwidth]{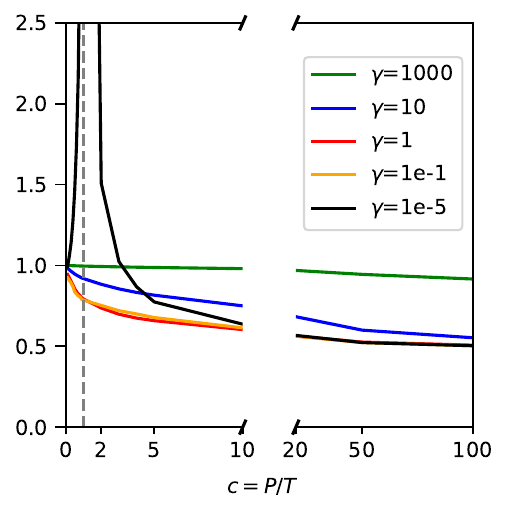}
            \caption{\footnotesize{Pricing Error}}
        \end{center}
    \end{subfigure}

    \end{center} 

\vspace{-0.8cm}   
\end{figure}

\paragraph{Random Split SDF.} 
The OOS performances of random split SDF are excellent. In the baseline result, we randomly generate 10 leaves in each P-Tree. We produce multiple P-Trees in parallel and stack all the leaf basis portfolios to construct an over-parameterized SDF. The number of leaf basis portfolios is $P$, and $c=P/T$ is the complexity measure. Following the train-test setting in our paper, we use 2001 to 2020 as the test sample, and the rolling window size is $T=240$. Following \cite{didisheim2024complexity}, we adopt a range of shrinkage levels $\gamma=\{10^3, 10, 0, 10^{-1}, 10^{-5}\}$, and we run 20 simulations to report the average performance. 
Figure \ref{fig: random_split_N10} illustrates the OOS performance of the baseline random split SDF. 
It shows two patterns as $c$ increases: 
(i) the Sharpe ratio exhibits double ascent for the lowest shrinkage case and permanent ascent for high shrinkage cases, and 
(ii) pricing error decreases for high shrinkage cases, with a spike around $c = 1$ for low shrinkage cases.
Interestingly, large regularized models achieve exceptionally high Sharpe ratios and low pricing errors, even though these basis portfolio are randomly generated.

\paragraph{Comparison.} Obviously, Figures \ref{fig: sr_split_rpf_10} and \ref{fig: random_split_N10} show similar patterns for the random P-Forest SDF and the random split SDF. 
The random P-Forest SDF has a Sharpe ratio about $4.0$ and pricing error less than $0.44$ for shrinkage level $\gamma=10^{-5}$ at $c=10$. However, the random split SDF has a Sharpe ratio of around $3.0$ and a pricing error about $0.60$.
The economics-guided split criteria help random P-Forest SDF outperform random split SDF. However, the inclusion of goal-oriented search is only intended to speed up the computation process and does not alter the performance bounds of SDF, if any.
With economic guidance, P-Tree can identify a relatively low-dimensional or sparse set of leaf basis portfolios that efficiently span the mean-variance efficient frontier. This property is intriguing in the over-parameterized environment and the classic parsimonious modeling world.

\paragraph{Role of \#Leaf.} 
The number of leaves (\# Leaf) is a critical parameter for a single P-Tree. We also investigate the impact of \# Leaf on random split SDF. Table \ref{tab:compare_NLeaf} reports the Sharpe ratio and pricing error for different \# Leaf with the sample $c=10$. We find that, as long as the shrinkage is not excessively large, i.e., $10^3$, the Sharpe ratio increases, and the pricing error decreases as \# Leaf increases from 5 to 100.
These findings are consistent for shrinkage levels $[10, 1, 10^{-1}, 10^{-5}]$. We believe that a larger \# Leaf results in more granular leaves, thus more diversification in the leaf basis portfolios, to benefit the OOS performance. Therefore, an increase in \# Leaf improves the OOS performance of the empirical SDF for the random split P-Trees.

We also attempted \# Leaf $ =2 $, with each P-Tree splitting only once, and the results can be found in Table \ref{tab:compare_NLeaf}. These empirical SDFs have Sharpe ratios less than 2 and pricing errors larger than 0.80, underperforming other specifications where \# Leaf $ = [5, 10, 20, 100] $. We believe these P-Trees restricted to one split do not involve enough randomness in the leaf basis portfolios.
There are only $61 \times 4$ distinct split candidates for such P-Trees, yielding only $488$ distinct portfolios. So, there are many identical portfolios, for example, when $c=10$ and the number of portfolios is $P=2,400$. Due to identical portfolios, we need to handle the collinearity issues, and the low penalty case ($\gamma=10^{-5}$) is ill-shaped. 

\begin{table}[htbp]
    \caption{\bf{Comparing OOS Performance of Random Split SDF: Role of \#Leaf}}
    \label{tab:compare_NLeaf}%
    {\footnotesize
    This table presents the OOS Sharpe ratio (Panel A) and pricing error (Panel B) of the random split SDF. The columns represent five levels of shrinkage.
    We have four specifications (RS, random split) of random split P-Trees in the rows.
    The number of leaves $N$ in each Tree is 5, 10, 20, and 100. The purpose of this table is to analyze the impact of $N$ on SDF in an over-parameterized environment. For a fair comparison, we report these results of complexity $c=P/T=10$ for all specifications.

    \begin{center}
        
        \begin{tabular}{lccccc}

        \toprule
        \\
           & $\gamma$=1000 & $\gamma$=10  & $\gamma$=1   & $\gamma$=1e-1 & $\gamma$=1e-5 \\
        \\
        \hline
        \\
        \multicolumn{6}{c}{Panel A: Sharpe Ratio} \\
        \\
        RS+2     & 0.46  & 1.17  & 1.60  & 1.68  & 0.99 \\
        RS+5     & 0.62  & 1.80  & 2.60  & 2.71  & 2.59 \\
        RS+10    & 0.73  & 2.13  & 2.87  & 2.91  & 2.86 \\
        RS+20    & 0.79  & 2.30  & 2.97  & 2.99  & 2.96 \\
        RS+100   & 0.86  & 2.51  & 3.03  & 3.02  & 3.01 \\
        \hline
        \\
        \multicolumn{6}{c}{Panel B: Pricing Error} \\
        \\
        RS+2     & 0.99  & 0.90  & 0.83  & 0.87  & 3.16  \\
        RS+5     & 0.98  & 0.80  & 0.64  & 0.64  & 0.70 \\
        RS+10    & 0.98  & 0.75  & 0.60  & 0.60  & 0.62 \\
        RS+20    & 0.98  & 0.72  & 0.58  & 0.58  & 0.59 \\
        RS+100   & 0.97  & 0.69  & 0.57  & 0.57  & 0.58 \\
        \bottomrule

        \end{tabular}%
    
    \end{center}

    }
\vspace{-0.5cm}   
\end{table}%

\paragraph{Role of the grid of threshold.} The choice of threshold grid is a crucial tuning parameter for P-Trees. Consistent with conventions in empirical asset pricing, quintile splitting (Grid 5) is used as the default setting. Additionally, alternative specifications are explored for robustness, including Grid 3 (two cutpoints and tertile portfolios) and Grid 10 (nine cutpoints and decile portfolios).

The impact of threshold grids on the OOS performance of random split P-Trees is shown in Table \ref{tab:factors_a_grid_1}. As the grid denser, the Sharpe ratio increases, and the pricing error decreases. This improvement is attributed to the enhanced complexity and randomness of the generated basis portfolios, leading to better performance with appropriate regularization. These findings align with the observations in \cite{didisheim2024complexity}, indicating that larger models' OOS performance improves with increased complexity.

\begin{table}[h!]
  
  \caption{\bf Threshold Grid and Random Split P-Tree}
  \label{tab:factors_a_grid_1}%

  {\footnotesize

    This table presents the results of the threshold grid specifications for random split P-Trees. We consider three configurations: Grid 3, Grid 5, and Grid 10, with Grid 5—corresponding to quintile asset sorting—serving as the default setting for P-Trees.
    We report the out-of-sample (OOS) Sharpe ratio of the Stochastic Discount Factor (SDF) and the pricing errors estimated by random split P-Trees. The training period spans from 1981 to 2000, while the test period covers 2001 to 2020.
  
  \begin{center}

    \begin{tabular}{l cccccc}
        
        \toprule

        \\
        Grid & & $\gamma$=1000 & $\gamma$=10  & $\gamma$=1   & $\gamma$=1e-1 & $\gamma$=1e-5 \\
        
        \\
        \multicolumn{7}{c}{Sharpe Ratio} \\
        \\

        3  & & 0.69 & 2.03 & 2.82 & 2.91 & 2.87 \\
        5  & & 0.73  & 2.13  & 2.87  & 2.91  & 2.86 \\
        10 & & 0.77 & 2.24 & 2.96 & 2.97 & 2.92 \\
        
        \\
        \multicolumn{7}{c}{Pricing Error} \\
        \\
        
        3  & & 0.98 & 0.76 & 0.60 & 0.60 & 0.62 \\
        5  & & 0.98  & 0.75  & 0.60  & 0.60  & 0.62 \\
        10 & & 0.98 & 0.73 & 0.58 & 0.59 & 0.61 \\
        
        \bottomrule
        
        \end{tabular}%
    
    \end{center}
    
    }
\vspace{-0.5cm} 
\end{table}%

\end{document}